\pdfoutput=1  % uncomment for arXiv, I guess
\documentclass[11pt]{article}
\usepackage{authblk}  % ad hoc
 \usepackage[]{acl} % Remove the "review" option to generate the final version.

\usepackage{times}
\usepackage{latexsym}

\usepackage[T1]{fontenc}

\usepackage[utf8]{inputenc}

\usepackage{microtype}

\title{Do Prompt-Based Models Really Understand\\ the Meaning of Their Prompts?}
\author[1,2]{Albert Webson}
\author[1]{Ellie Pavlick}
\affil[ ]{\{albert\_webson, ellie\_pavlick\}@brown.edu}
\affil[1]{Department of Computer Science, Brown University}
\affil[2]{Department of Philosophy, Brown University}

\usepackage{graphicx}
\usepackage{booktabs}
\usepackage{array}  % for table resizing
\usepackage{enumerate}
\usepackage{amssymb}
\usepackage{amsmath}

\usepackage{subcaption}
\usepackage{makecell}

\renewcommand{\sec}[1]{\S\ref{sec:#1}}
\newcommand{\rf}[1]{\autoref{#1}}

\graphicspath{{./vector-v2/}{./appendix/}{./vector-v1/}{./raster/}}

\definecolor{brown}{RGB}{103, 04, 5}
\definecolor{lunar}{RGB}{21, 48, 122}
\definecolor{Gray}{gray}{0.9}
\definecolor{mint}{rgb}{0.24, 0.71, 0.54}
\newcommand{\prompt}{\texttt}
\renewcommand{\bold}{\textbf}

\newcommand{\prem}{\{premise\} }
\newcommand{\hypo}{\{hypothesis\}}

\newcommand{\pron}{\{pronoun\}}
\newcommand{\refs}{\{referent\}}

\usepackage{contour}
\usepackage[normalem]{ulem}

\contourlength{0.8pt}
\newcommand{\myuline}[1]{%
  \uline{\phantom{#1}}%
  \llap{\contour{white}{#1}}%
}

\usepackage{pifont}  % for check and cross marks
\newcommand{\cmark}{\ding{51}}%

\newcommand{\fnt}[1]{\footnotetext{#1}}
\newcommand{\fn}{\footnotemark\ }
\begin{document}
\maketitle

% OpenReview TLDR: GPT-3 and T0 often perform just as well with intentionally irrelevant and pathologically misleading prompts as they do with instructive prompts, thus questioning if models really understand the meaning of their prompts.

\begin{abstract}
Recently, a boom of papers has shown extraordinary progress in zero-shot and few-shot learning with various prompt-based models. 
It is commonly argued that prompts help models to learn faster in the same way that humans learn faster when provided with task instructions expressed in natural language. 
In this study, we experiment with over 30 prompt templates manually written for natural language inference (NLI). We find that models learn just as fast with many prompts that are intentionally irrelevant or even pathologically misleading as they do with instructively “good” prompts. 
Further, such patterns hold even for models as large as 175 billion parameters \citep{gpt} as well as the recently proposed instruction-tuned models which are trained on hundreds of prompts \citep{t0}. 
That is, instruction-tuned models often produce good predictions with irrelevant and misleading prompts even at zero shots. 
% In sum, notwithstanding prompt-based models’ impressive improvement, we find evidence that such improvement is not derived from models understanding task instructions in ways analogous to humans' use of task instructions.
%In sum, despite prompt-based models' dramatic improvement in zero-shot and few-shot learning, we find limited evidence that models' improvement is derived from models understanding task instructions in ways analogous to humans' use of task instructions.
%In sum, notwithstanding prompt-based models’ impressive improvement, we find evidence of serious limitations that question the degree to which language models really understand the meaning of prompts in the way humans do.
In sum, notwithstanding prompt-based models’ impressive improvement, we find evidence of serious limitations that question the degree to which such improvement is derived from models understanding task instructions in ways analogous to humans' use of task instructions.
\end{abstract} % I really disagree with drawing a strong positive or negative conclusion here. The topic is too new to warrant a critical paper, everyone is still figuring things out. It will be better received, and sound more intelligent, if it reads as coming from a researcher who is curious and eager to do more work in the area, not as someone who is fed up with the work being done. :)

\section{Introduction} \label{sec:intro}
Suppose a human is given two sentences: “No weapons of mass destruction found in Iraq yet.” and “Weapons of mass destruction found in Iraq.”
They are then asked to respond 0 or 1 and receive a reward if they are correct. In this setup, they would likely need a large number of trials and errors before figuring out what they are really being rewarded to do. This setup is akin to the pretrain-and-fine-tune setup which has dominated NLP in recent years, in which models are asked to classify a sentence representation (e.g., a CLS token) into some arbitrary dimensions of a one-hot vector. %since at least \citet{word2vec}.
%\footnote{Although researchers see the those labels as natural language like “entail”, “contradict”, or “neutral”, recall that models only see them as an arbitrary dimension of a one-hot vector.} 
In contrast, suppose a human is given a prompt such as: \myuline{Given that “}no weapons of mass destruction found in Iraq yet.\myuline{”, is it definitely correct that “}weapons of mass destruction found in Iraq.\myuline{”?}\fn Then it would be no surprise that they are able to perform the task more accurately and without needing many examples to figure out what the task is.
\fnt{This prompt is adapted from MultiNLI \citep[p. 3]{williams-etal-2018-broad}'s instructions to crowdsourced workers, while the example is the first one in RTE's training set.} %We italicize prompts to avoid ambiguity of nested quotes.}

Similarly, reformatting NLP tasks with prompts such as the underlined text above has dramatically improved zero-shot and few-shot performance over traditional fine-tuned models \citep{schick2,le-scao-rush-2021-many,t0,flan}. Such results naturally give rise to the hypothesis that the extra prompt text included within each input example serves as semantically meaningful task instructions which help models to learn faster, in the way task instructions help humans to learn faster. This hypothesis is implicitly assumed by many and explicitly argued by \citet{mishra2021natural}, \citet{schick1}, and \citet{gpt}.

While last years saw a gold rush of papers (summarized in \sec{literature}) that proposed automatic methods for optimizing prompts, \citet{logan2021cutting} compare a representative sample of these newly proposed methods and report that \citet{schick2}'s manually written prompts still on average outperform the automatically searched prompts across a range of SuperGLUE tasks \citep{sglue}. Such findings suggest that expert-crafted prompts are among the best, if not \textit{the} best, which reinforces the above hypothesis that models benefit from meaningful instructions. 

In this paper, we test this hypothesis by evaluating various language models on NLI in zero-shot and few-shot settings using more than 30 manually written templates and 13 sets of LM target words for a total of over 390 prompts. 
% TODO maybe make a bullet list of the below?
We find that in most cases models learn identically as fast when given irrelevant or misleading templates as they do when given instructively good templates.
Further, models ranging from 235 million to 175 billion parameters all exhibit this behavior, as do the instruction-tuned models, which are trained on hundreds of manually written prompts. 
While we confirm \citet{t0}'s finding that instruction tuning substantially improves the performance and robustness of prompts, we also find that instruction-tuned models can be, in some sense, too robust and less sensitive to the semantics of the prompts, as compared to their non-instruction-tuned equivalents. 
Finally, models are much more sensitive to the choice of the LM target words as opposed to the meaning of the instruction templates. % which is the opposite of what we expect from humans % Additionally, models are overly sensitive to the choice of the LM targets: The mapping \{“yes” $\rightarrow$ entailment, “no” $\rightarrow$ non-entailment\} substantially outperforms all other word-to-label mappings, even when “yes”/“no” is not syntactically or semantically well-formed with the overall prompt (\sec{verbalizer}).
In sum, despite prompt-based models' dramatic improvement in zero-shot and few-shot learning, we find limited evidence that models' improvement is derived from models understanding task instructions in ways analogous to humans' use of task instructions.

%\begin{table*}[t]
%\vspace{-4ex}
%\resizebox{\textwidth}{!}{%
%\begin{tabular}{@{}lll@{}}
%\toprule
%Category & Description & Examples \\ 
%\midrule
%instructive &\makecell[l]{How we would describe the NLI task\\ to a human who has never seen the task before.} &\makecell[l]{\prem Are we justified in saying that “\hypo”?\\ Given \prem Should we assume that “\hypo” is true?} \vspace{2ex} \\ 
%\makecell[l]{misleading-\\ moderate} &\makecell[l]{Instruct the models to perform a task related\\ or tangential to NLI such that, if the model\\ were to perform the task as explicitly instructed,\\ it would perform poorly on NLI in general.\footnotemark} &\makecell[l]{\prem Can that be paraphrased as: “\hypo”?\\ \prem Are there lots of similar words in “\hypo”?} \vspace{2ex} \\
%\makecell[l]{misleading-\\ extreme} &\makecell[l]{Instruct the models to perform a task unrelated\\ to NLI.} &\makecell[l]{\prem is the sentiment positive? \hypo\\ \prem is this a sports news? \hypo} \vspace{2ex} \\ 
%irrelevant &\makecell[l]{Concatenate the premise, a sentence unrelated\\ to any NLP task, and the hypothesis.} &\makecell[l]{\prem If bonito flakes boil more than a few seconds\\ the stock becomes too strong. "\hypo"?} \vspace{2ex} \\ 
%null &\makecell[l]{Concatenate the premise and the hypothesis\\ without any additional text.} &\makecell[l]{\{premise\} \{hypothesis\} \\ \hypo \prem} \\ 
%\bottomrule
%\end{tabular}}
%\caption{Prompt templates used in this paper. See \rf{sec:all-templates} for the full list.}
%\label{tab:templates}
%\end{table*}

\section{Related Work} \label{sec:literature}
\subsection{Prompt-Based Models} \label{sec:prompt-explainer} 
At the time of writing, the terms “prompt tuning” and  “prompting” can refer to any one or combination of three approaches described below: 

\bold{Discrete Prompts} reformat each example with some template text. For example, in a sentiment analysis task, the template can be \prompt{\{sent\} In summary, the restaurant is [prediction]}, where the predicted mask word is then converted to a class prediction by a predefined mapping, e.g., \{“great” $\rightarrow$ positive, “terrible” $\rightarrow$ negative\}. The prompts can be manually written \citep{schick1,bragg2021flex} or automatically generated \citep{gao2021making,shin-etal-2020-autoprompt}. This approach typically tunes all parameters of the model, but its few-shot performance can exceed that of very large models (e.g., GPT-3 175B) despite using a 3 orders of magnitude smaller LM \citep{schick2,tam2021improving}. 

\bold{Priming} (a.k.a. in-context learning) prepends $k$ priming examples to the evaluation example, where each example is optionally wrapped in a template such as \prompt{Question: \{sent$_1$\} True or false? \{label$_1$\} … Question: \{sent$_k$\} True or false? \{label$_k$\} Question: \{eval\_sent\} True or false? [prediction]}. Notably, although models see labeled examples, their parameters do not receive gradient updates based on those examples. Although this approach is intriguing, \citet{gpt} report that it only performs well on the largest GPT-3 model, the API of which is costly and difficult to use for academic research (see \rf{sec:non-profit} for details). 
%we don't compare to chain of thought, not instruction, require examples.
\nocite{MetaICL} %shows instructions are optional for priming
% the API of which is prohibitively expensive for most academic research and severely limited by various quotas,
%\footnote{In addition to the high cost of its API (priming by token), cost quota, context length quota, and FT quota. Appendix details difficulty.}
%  "the largest proprietary gpt-3 model, limiting research/reproducibility."

\bold{Continuous Prompts} prepend examples with special tokens, optionally initialized with word embeddings; but during learning, those tokens can be updated arbitrarily such that the final embeddings often \textit{do not} correspond to any real word in the vocabulary (e.g., \citealp{T5LMA,li-liang-2021-prefix,qin-eisner-2021-learning}). This approach often efficiently tunes a much smaller set of model parameters, but these methods have not yet reported success in few-shot settings. Moreover, foregoing prompts as expressed in natural language makes it much harder to study their semantics, and it is not clear if continuous prompts serve as task-specific instructions or simply more efficient model parameters (see \citealp{cont-prompt-analysis} for a detailed analysis).
%waywardness show continuous prompts also not meaningful}

\subsection{Analyses of Prompts}
In this paper, we focus on discrete prompts because we can manually write and control their wording and semantics. We measure the effect of prompt semantics by the model's $k$-shot performance where $k = \{0,4,8,16,32,64,128,256\}$. This setup resembles that of \citet{le-scao-rush-2021-many}, but their study focuses on comparing \citet{schick2}'s existing small set of prompts against traditional fine-tuning over the training trajectories of entire training sets, whereas our study focuses on the few-shot learning trajectories among a much more diverse set of prompts designed to test specific hypotheses about the effect of prompt semantics on few-shot learning speed.

At a high-level, our findings contradict \citet{mishra2021natural}'s claim that models benefit from elaborate instructions adapted from crowdsourcing annotation guides. 
But note that they define “instructions” more broadly as including priming examples, and they find that “GPT-3 benefits the most from positive examples, mildly from definition, and deteriorates with negative examples.” (p. 18). 
In other words, if we ablate priming and narrow “instructions” to just the description of a task, we in fact have the same finding that instructions are only modestly beneficial over no instructions (cf. our irrelevant templates). %but we further show that good instructions have no consistent benefit over bad instructions, thus raising questions of whether models' use of prompts can be fairly described as “understanding”. % TODO shrink discussion on Mishra
In a similar vein, concurrent work by \citet{lampinen2022} finds that other components of a prompt such as explanations of priming examples are helpful, but models are indifferent to whether the instructions in fact describe their tasks. 

Finally, a growing body of concurrent work also questions the degree to which models need meaningful instructions \citep{khashabi2021prompt,prasad2022grips}. One particularly noteworthy finding is that \citet{min2022rethinking} show that models learn just as well with incorrect labels as opposed to correct labels in priming, concluding that prompts are helping models to learn the distribution of the input text and space of possible labels (as opposed to specifying instructions of the task).

\section{Overall Setup} \label{sec:setup} % General Setup? 
%\paragraph{Implementation}
We implement a manual discrete prompt model%\footnote{Publicly available on GitHub along with all hyperparameters, interactive figures, and statistical test results. Anonymized for submission but included in supplementary materials.} 
\footnote{All code, interactive figures, and statistical test results are available at \url{https://github.com/awebson/prompt_semantics}}
which in essence is the same as that of \citet{schick2}, except their implementation includes several augmentations such as self-labeling and ensembling of multiple prompts for competitive results. In order to focus on measuring the effect of prompts themselves, our implementation does not include those augmentations. Following \citet{t0} and \citet{flan}, we evaluate by a rank classification of the target words. %(see \rf{sec:rank-eval} for details).

\paragraph{Baseline Model}
\label{sec:model}
In preliminary experiments, we fine-tuned and prompt-tuned BERT, DistilBERT, RoBERTa, ALBERT, and T5 (\citealp{devlin2018bert,sanh2019distilbert,liu2019roberta,lan2019albert,t5}; all implemented via \citealp{wolf-etal-2020-transformers}). Confirming prior work \citep{schick2,tam2021improving}, we find that ALBERT consistently yields the best performance, so we use it as our baseline model.

\begin{figure}[t]
    \vspace{-2ex}
    \centering
    \includegraphics[width=\linewidth]{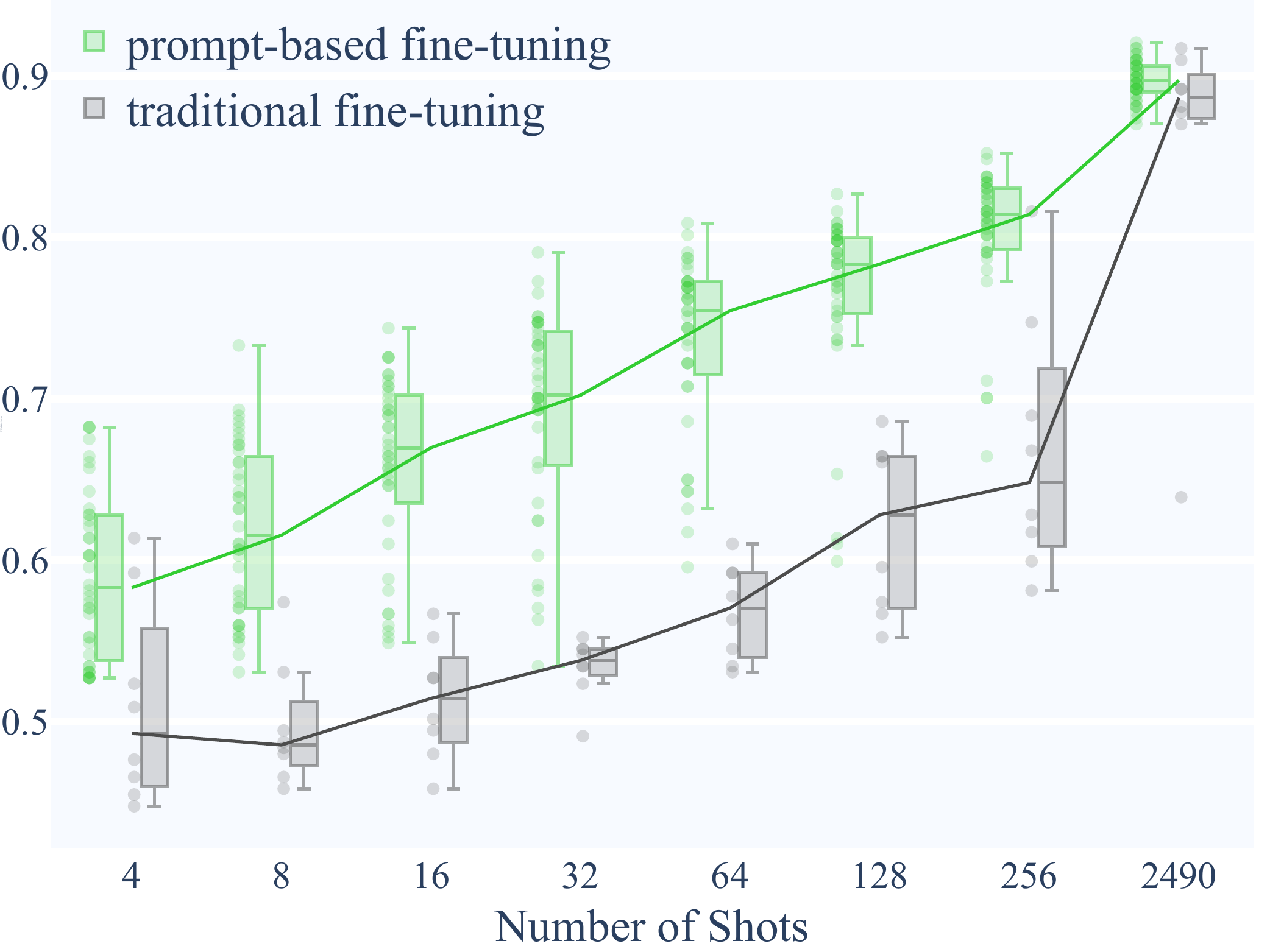}
    % Baseline model (ALBERT) on RTE validation. 
    \caption{How to read these figures: Each dot is the performance of one prompt under one random seed (which controls the sets of few-shot examples) of our baseline model (ALBERT) on RTE validation set. Boxes span from the first quartile to the third quartile, while lines inside boxes mark the medians. Later figures omit the points except outliers in order to improve legibility. See the interactive figures in our GitHub repository or \rf{sec:ind-temp} for the results of individual prompts.} %See \rf{sec:all-RTE} for the corresponding tables with numerical results.}
    \label{fig:ft-pt-check}
    \vspace{-2ex}
\end{figure}

To verify that our implementation is comparable with prior work, \autoref{fig:ft-pt-check} reports the RTE validation accuracy of our baseline model. At 32 shots, our implementation yields a median accuracy of 70.22\% (mean = 69.29\%, std. dev. = 6.3\%), which is comparable to the 69.8\% reported by \citet{schick2}.
Further, \autoref{fig:ft-pt-check} confirms \citet{le-scao-rush-2021-many}'s finding that, while both fine-tuning and prompt-tuning converge to similar results when fully trained on the entire set ($n = 2490$ for RTE), prompt-tuning yields the largest improvement in the few-shot setting. 
Going forward, we focus on studying the few-shot learning trajectory between 4 and 256 examples. 
% which is reasonable as we would expect a human would also be able to eventually figure out the underlying task based on the loss and reward even when the prompts are arbitrary.
%Variance across seeds is moderate but acceptable. PT vary much less than FT. 

\paragraph{Instruction-Tuned Model}
We additionally experiment with T0, a recently proposed instruction-tuned model which is trained on over 60 datasets %\footnote{Importantly, T0 always holds out all NLI prompts and all NLI datasets in its training, which makes it a fair comparison to other models in this paper.} 
formatted with hundreds of manually written prompts \citep{t0}. We experiment with both sizes of T0  (3B and 11B), as well as their non-instruction-tuned version, T5 LM-Adapted \citep{T5LMA}, as a baseline. 

\paragraph{Very Large Model} \label{sec:setup-gpt}
Lastly, we experiment with the largest GPT-3 (175B) via priming (a.k.a. in-context learning). Although fine-tuning is technically available, it is extremely limited by OpenAI's various quotas. %and that no baseline comparison is publicly available. 
% one of which is that, despite what is reported in \citet{gpt}, its context length only permits up to 16 shots for RTE, so we report 16-shot accuracy for all of our large ($\geq$ 11B) models.
See \rf{sec:non-profit} for details on how we circumvent challenges in reproducing \citet{gpt}'s results.

\paragraph{Data}
NLI is a task where a model is asked to classify whether one piece of text (the ``premise'') entails another (the ``hypothesis'').
We focus on NLI because all T0 variants holds out all NLI prompts and all NLI datasets in its training, which makes it a fair comparison to other models in this paper. %Although FLAN \citep{flan} has other held-out tasks, the model is not publicly available. 
% We focus on NLI because, first, it is the only task held out by all variants of T0.\footnote{While FLAN has other held-out tasks, the model is not publicly available.} 
% compared to the usual suite of NLP classification tasks such as topic classification and question answering,\footnote{Consider other NLP tasks such as topic classifications. As long as the target words (e.g., “politics” or “sports”) are provided, humans likely do not really need detail instructions to figure out the task is to associate the input text with target words of similar topics.}
% NLI is in theory more sensitive to differences in task instructions. For example, depending on if an instruction asks for strictly logical entailment or pragmatic inference, humans can give different predictions on the same premise and hypothesis. Thus, we conjecture that NLI's sensitivity to nuanced differences in task instructions can magnify measurements of to what extent are prompt-based models sensitive to the meaning of prompts.

We use Recognizing Textual Entailment (RTE, \citealp{dagan}, inter alios), a series of expert-annotated NLI datasets. Specifically, we use the SuperGLUE collection of RTE (i.e., RTE1, 2, 3, and 5; all converted to binary classification) and report their validation accuracy for comparability with prior work on prompts. 

We also experiment with Adversarial NLI (ANLI, \citealp{anli}), Heuristic Analysis for NLI Systems (HANS, \citealp{mccoy}), and Winograd Schema Challenge (WSC, \citealp{winograd}), reported in Appendices \ref{sec:all-ANLI}, \ref{sec:hans-preliminary}, and \ref{sec:wsc-preliminary}, respectively. We find no qualitative difference between their and the main RTE results except that ANLI requires much larger number of shots before obtaining any above-random accuracy, as it is designed to be a highly challenging set.

\paragraph{Random Seeds \& Example Sampling} \label{sec:random-seeds} All experiments are run over the same set of 4 random seeds. Within a given seed, all models see the same set of examples. For instance, under seed 1, the 4-shot models see examples 550--553, the 8-shot models see examples 550--557, and so on. Across different seeds, a different starting example index is drawn. The exact training example indices are also recorded in our GitHub repository for reproducibility.

\paragraph{Statistical Tests} \label{sec:stats} We use both ANOVA and its nonparametric equivalent, the Kruskal–Wallis test. After finding a significant difference among multiple categories of templates, we report pairwise significance with the independent two-sample $t$-test and the Wilcoxon rank-sum test. % Significance is always computed using models at the same number of shots.
We set $\alpha = 0.05$ and apply the Bonferroni correction to account for multiple comparisons. %(i.e., at each number of shots). 
For all results reported in this paper, both $t$-test and Wilcoxon agree.

% Throughout this paper, when we do not find a statistically significant difference, we say there is “no practical difference”, implying that although it is possible to use less standard statistical tests to reject the null hypothesis, their effect size is so small that there is no practical difference between the two conditions (e.g., two categories of prompts) such that we should expect them to make a difference on a different dataset or a different model.  

\section{Effect of Templates} \label{sec:template} \label{sec:template-results}

%\paragraph{Template Categories} 
%\label{sec:template-sem}
Our research question is whether models understand prompts as meaningful task instructions analogous to how humans would. For intuition, suppose an experimenter provides a human annotator with an informative instruction of a reasonably easy task.
If the annotator understands the instruction, we expect them to perform better than when the experimenter provides intentionally misleading instructions, makes irrelevant chitchat, or says nothing at all. 
Accordingly, we write various prompt templates that correspond to these different scenarios and evaluate models' performance with these templates in zero-shot and few-shot settings. 
%We test the hypothesis that models understand prompts as meaningful task instructions by comparing good prompts to different types of bad prompts and observing the effects on their few-shot learning speed.  

\renewcommand{\prem}{\{prem\} }
\renewcommand{\hypo}{\{hypo\}}
\begin{table}[t]
%\vspace{-2ex}
\resizebox{.51\textwidth}{!}{%
\begin{tabular}{@{}ll@{}}
\toprule
Category & Examples \\ 
\midrule
instructive &\makecell[l]{\prem Are we justified in saying that “\hypo”?\\ Suppose \prem Can we infer that “\hypo”?} \vspace{2ex} \\ 
\makecell[l]{misleading-\\ moderate} &\makecell[l]{\prem Can that be paraphrased as: “\hypo”?\\ \prem Are there lots of similar words in “\hypo”?} \vspace{2ex} \\
\makecell[l]{misleading-\\ extreme} &\makecell[l]{\prem is the sentiment positive? \hypo\\ \prem is this a sports news? \hypo} \vspace{2ex} \\ 
irrelevant &\makecell[l]{\prem If bonito flakes boil more than a few seconds\\ the stock becomes too strong. "\hypo"?} \vspace{2ex} \\ 
null &\makecell[l]{\{premise\} \{hypothesis\} \\ \{hypothesis\} \{premise\}} \\ 
\bottomrule
\end{tabular}}
\caption{Example templates for NLI.}
\label{tab:templates}
\vspace{-2ex}
\end{table}
\renewcommand{\prem}{\{premise\} }
\renewcommand{\hypo}{\{hypothesis\}}

\subsection{Method}
%\paragraph{Templates} 
We write 5 categories of templates (\rf{tab:templates}), with at least 5 templates for each category (10 for instructive): %10 misleading split evenly, and at least 5 templates for each other category. 
\begin{itemize}
    \item Instructive: how we would describe the NLI task to a human who has never seen this task before.
    \item Misleading-Moderate: instruct the models to perform a task related or tangential to NLI such that, if the model were to perform the task as explicitly instructed, it would perform poorly on NLI in general.\footnote{An author manually labeled the 30 training examples seen by models under random seed 1 (example nos. 550--580), among which we find 17 pairs of entailment, 5 or 8 pairs (depending on how strictly one judges their acceptability) of summarizations, and only one pair of paraphrase.}
    \item Misleading-Extreme: instruct the models to perform a task unrelated to NLI.
    \item Irrelevant: concatenate the premise, a sentence unrelated to any NLP task, and the hypothesis.
    \item Null: concatenate the premise and the hypothesis without any additional text.
\end{itemize}
See \rf{tab:templates} for examples and \rf{sec:all-templates} for the full list.
We use “prompt” to mean a unique combination of a template and a predefined LM target word for each class label. 
For example, \{“yes” $\rightarrow$ entailment, “no” $\rightarrow$ non-entailment\} are the default targets for the template \prompt{\prem Should we assume that \hypo? [prediction]}. 
In this section, to control for the effect of target words, a template's performance is always reported with “yes”/“no” as its target words, which consistently perform best.
In \rf{sec:verbalizer}, we control for the templates and study the effect of different target words.
%Except in ablation studies (\rf{sec:qmarks}), 
We further control for punctuation, declarative vs. interrogative templates, %\footnote{Declarative templates require inserting masks in the middle of a sentence, e.g., \prompt{\prem It [mask] be true that \hypo} where \{“must” $\rightarrow$ entailment, “might” $\rightarrow$ non-entailment\}. This is only natural for encoder-only mask LMs. To make it a fair comparison with encoder-decoder models, we only use interrogative templates in our main experiments.} 
and the order of concatenation (always \prompt{\prem \textit{some template text} \hypo [prediction]}). 

After preliminary experiments, to avoid cherry picking, all prompts reported in this paper were written prior to evaluation, i.e., we do not allow retroactively editing prompts for performance manipulations, except for an ablation study that explicitly studies the effect of punctuation (\rf{sec:qmarks}). %retroactively excluding or including templates from the categories defined above.

\begin{figure}[t]
    \vspace{-4ex}
    \centering
    \includegraphics[width=\linewidth]{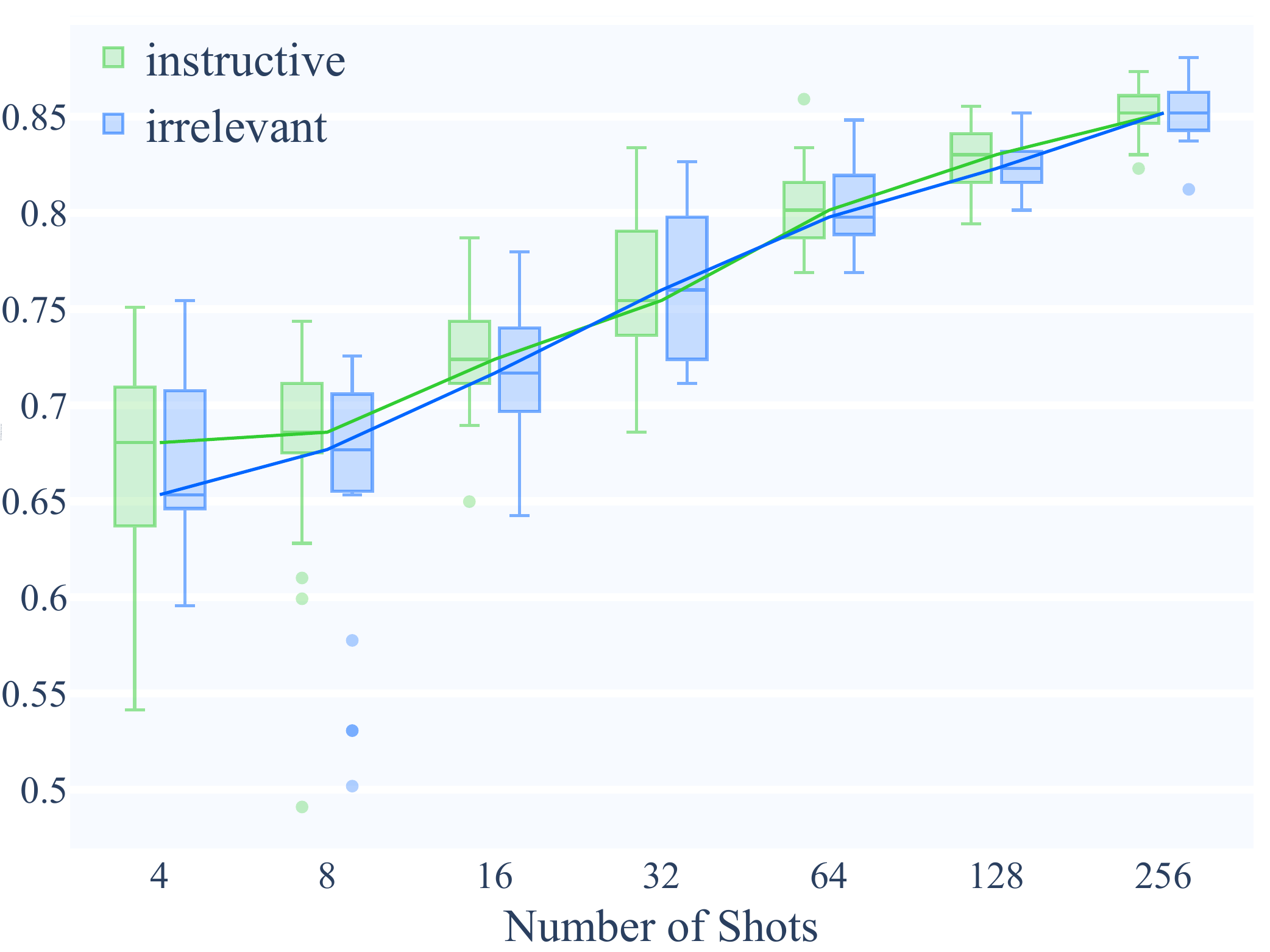}
    \caption{T0 (3B) on RTE. There is no practical difference between the performance of the models trained with instructive templates vs. those trained with irrelevant templates at any number of shots.}
    \label{fig:irrelevant-T0-3B}
\end{figure}

\begin{figure}[t]
    \centering
    \includegraphics[width=\linewidth]{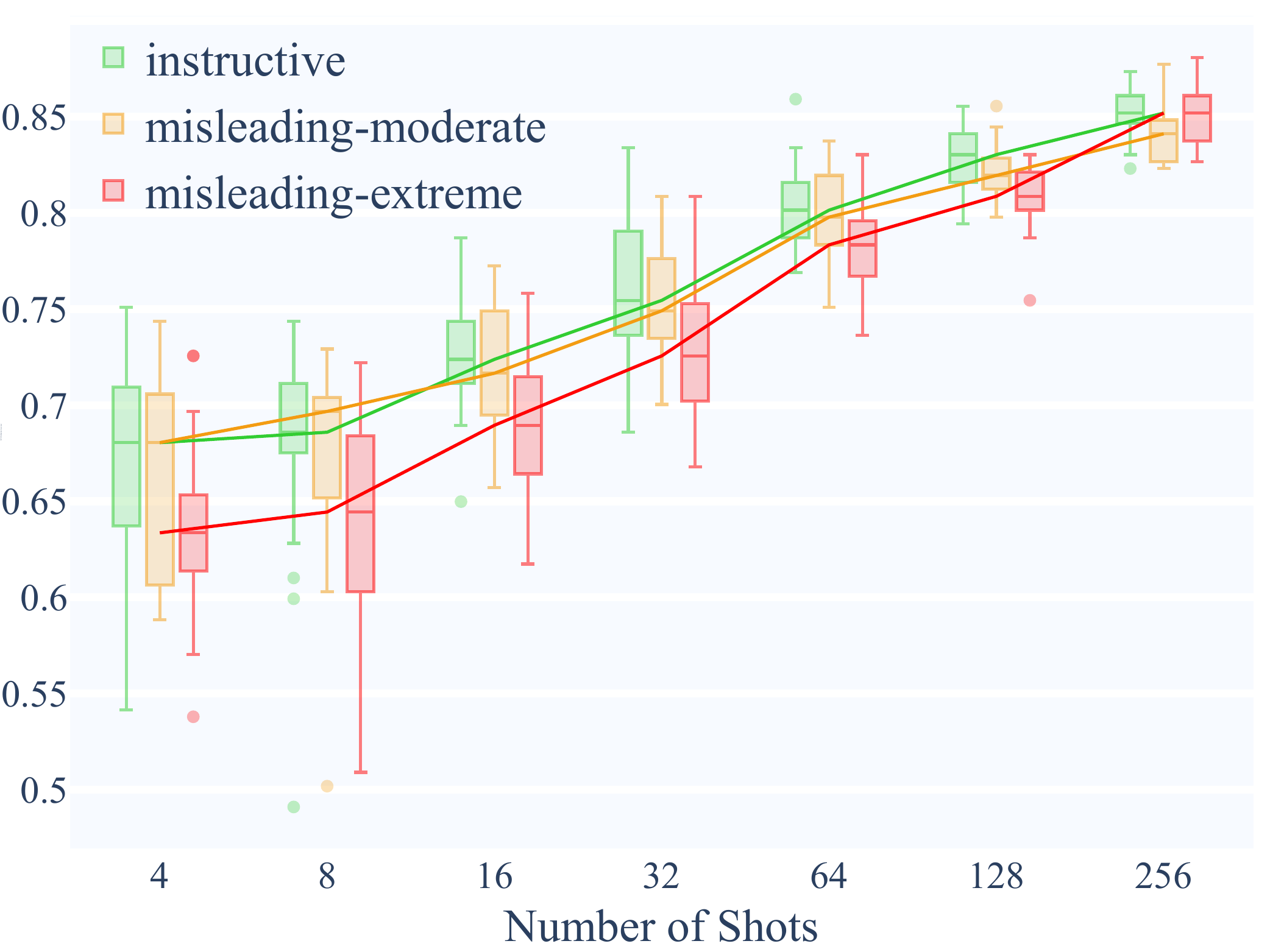}
    \caption{T0 (3B) on RTE. There is no practical difference between models trained with instructive and misleading-moderate templates at any number of shots. But models trained with misleading-far templates are statistically significantly worse from 8 to 128 shots.}
    \label{fig:misleading-T0-3B}
    \vspace{-2ex}
\end{figure}

\begin{figure}[t!]
    \vspace{-4ex}
    \centering
    \includegraphics[width=\linewidth]{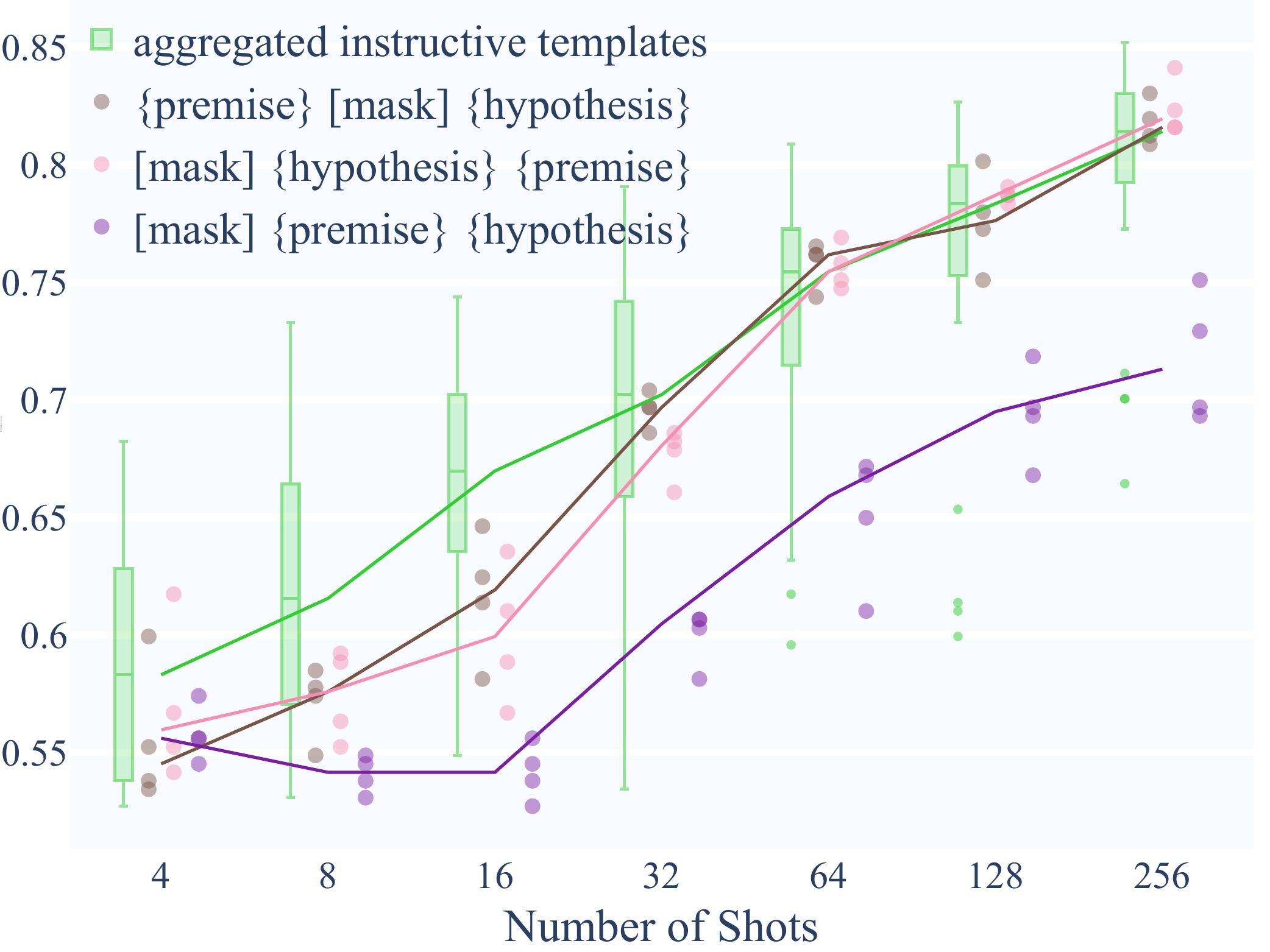}
    \caption{ALBERT on RTE. After 32 shots, models trained with 2 null templates learn just as fast as the instructive templates, but models trained with other null templates (e.g., purple) are much worse.}
    \label{fig:pt-null}
    \vspace{-2ex}
\end{figure}

\begin{figure*}[t]
    \centering
    \vspace{-6ex}
    \includegraphics[width=.85\linewidth]{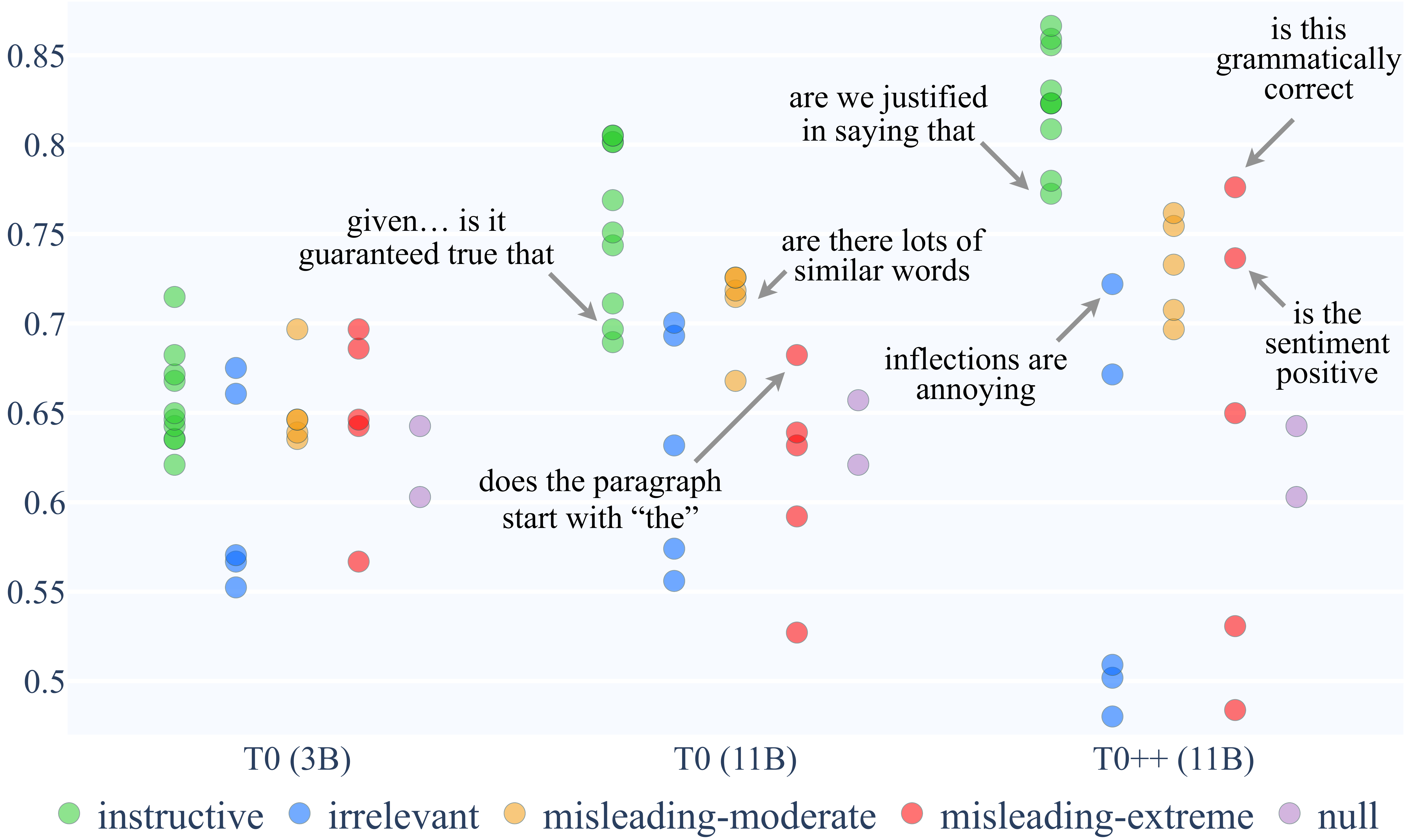}
    \caption{Zero-shot accuracy of instruction-tuned models on RTE. Each prompt's performance is a single point (unlike the few-shot figures where each prompt is approximated by multiple points with multiple samplings of few-shot examples.) Arrows highlight some prompts with their excerpts. See \rf{sec:all-zero-shot} for the full results.} %See \rf{sec:all-templates} for the full list of prompts.
    \label{fig:zero-shot}
    \vspace{-1ex}
\end{figure*}

\begin{figure}[t]
    \centering
    % \vspace{-2ex}
    \includegraphics[width=\linewidth]{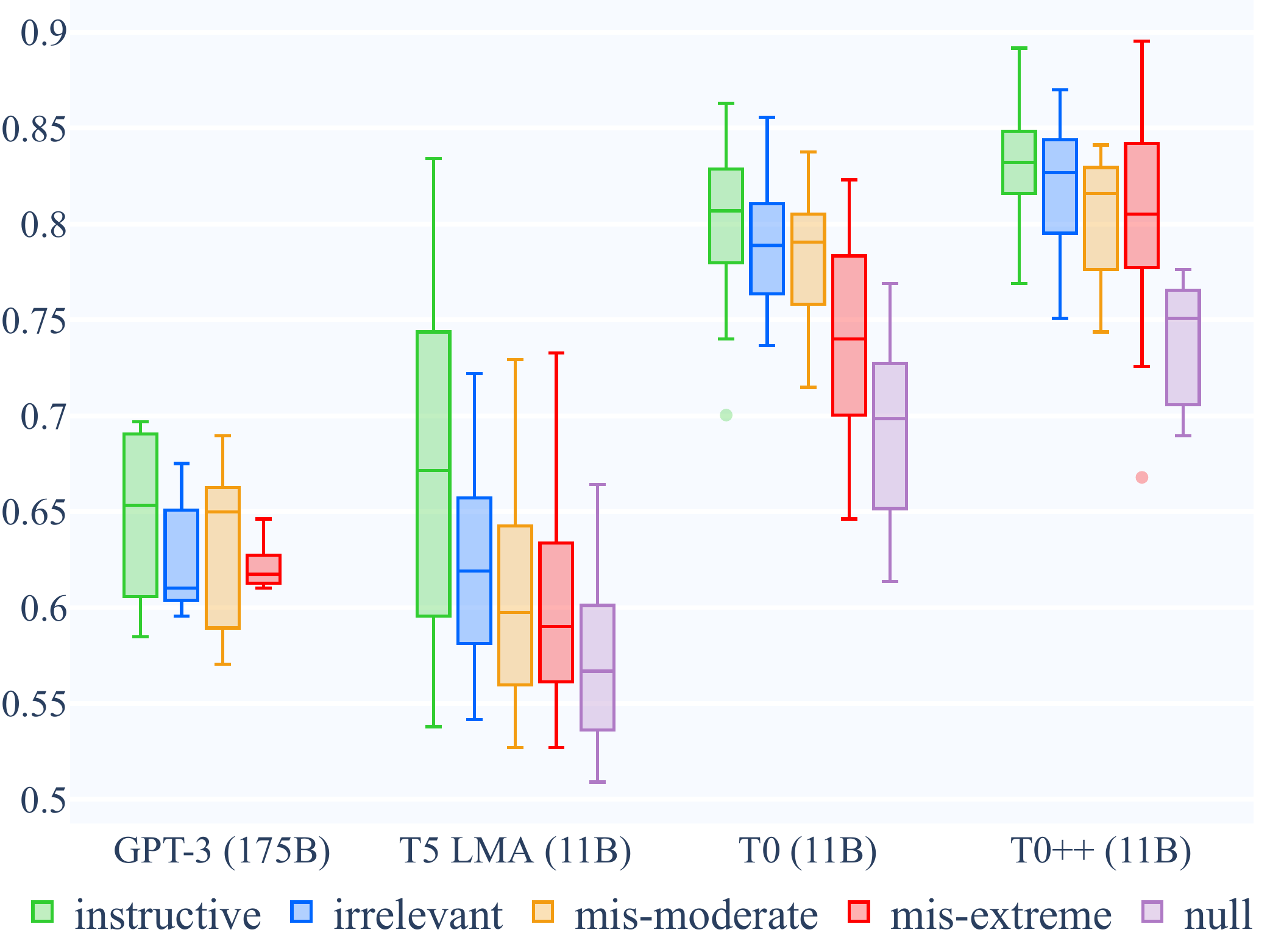}
    \caption{16-shot accuracy of four large models on RTE. For GPT-3, there is no practical difference between any template categories except null (not plotted because they are below 0.5). For T5, there is no practical difference between instructive and irrelevant. For T0, there is no practical difference between instructive and irrelevant nor between instructive and misleading-moderate. For T0++, there is no practical difference between instructive and irrelevant nor between instructive and misleading-extreme.} 
    % GPT-3 uses priming, with more effective 16-shots examples seen.} 
    % Only 1 "seed" for GPT due to high experiment cost and that, although Brown et al. reported 32-shot performance, their commercial API only allows 2048 tokens which prevents priming beyond 16 shots for RTE.
    % Possibly promote this to be page 1 figure 1. I know you said to "not give away results in intro", but I'm afraid that's a little too high-minded when papers must compete for attention…?
    \label{fig:big-models}
    \vspace{-2ex}
\end{figure}

\subsection{Result}
\paragraph{Irrelevant Templates} 
%Unlike the misleading templates, which at least ask the model to perform \textit{some} NLP task, irrelevant templates only provide utterly irrelevant chitchat, often not even asking any question. 
We find that models trained with irrelevant templates learn just as fast as those trained with instructive templates, with no practical difference\footnote{We acknowledge that a lack of a statistically significant difference does not entail “no difference”. While it is true that we find no statistically significant difference with the independent two-sample $t$-test and the Wilcoxon rank-sum test whenever we say “no practical difference”, note that our argument, here and throught the paper, hinges on the very small effect sizes, not the significance tests, i.e., the two categories of prompts perform too similarly in absolute terms.} %we cannot expect the prompt category to make a difference on a different dataset or a different model.}
at any number of shots (\rf{fig:irrelevant-T0-3B}). This is true for all models and all datasets we experimented, including the largest GPT-3 (\rf{fig:big-models}). %as well as the instruction-tuned T0

\paragraph{Misleading Templates}
\label{sec:misleading}
%While the success of irrelevant prompts suggests models may be insensitive to the meaning of the prompts, the results on misleading templates suggest otherwise. 
There is no consistent relation between the performance of models trained with templates that are moderately misleading (e.g. \prompt{\prem Can that be paraphrased as "\hypo"?}) vs. templates that are extremely misleading (e.g., \prompt{\prem Is this a sports news? \hypo}). 
T0 (both 3B and 11B) perform better given misleading-moderate (\rf{fig:misleading-T0-3B}), 
ALBERT and T5 3B perform better given misleading-extreme (Appendices \ref{sec:baseline-figs} and \ref{sec:t5-3B-rte}), 
whereas T5 11B and GPT-3 perform comparably on both sets (\rf{fig:big-models}; also see \rf{tab:summary} for a summary of statistical significances.)
Despite a lack of pattern between the two misleading categories, however, it is consistent that each model exhibits significantly better performance on instructive templates compared to at least one category of misleading templates. % revised from "are able to differentiate" 

\paragraph{Null Templates} \label{sec:dumb-matter}
%One interpretation of the above results is that models simply are ignoring the template words as some kind of noise. We will fully discuss this in \rf{sec:alternatives}, but this interpretation faces difficulty in reconciling with the results on null templates, which simply concatenate the premise and the hypothesis with no template words (thus no noise). 
Models trained with null templates perform far worse than all other categories of templates (see \rf{sec:all-RTE} for all null results). Here, we focus on ALBERT (an encoder-only masked language model), %\footnote{\todo{} technically possible with rank eval, but worse performance in practice.} 
which allows more permutation of concatenation orders by placing mask in the middle of sentences. We see that, although null templates are much worse in aggregate, some subset of them (e.g., \prompt{\prem [mask] \hypo}) are still able to learn nearly as fast as the average instructive template after 32 shots (\rf{fig:pt-null}). %Additionally, punctuation can also have an outsized effect (which we control for in the main experiments; see \rf{sec:qmarks} for an ablation study). 

\paragraph{Zero-Shot}

So far, we have focused on few-shot results. At zero shots, all models (including GPT-3 175B) perform only marginally above random, except the instruction-tuned T0. Thus, for our analysis of zero shot performance, we focus on T0. 
\rf{fig:zero-shot} shows that there is no practical difference between the performance of T0 3B given instructive templates and either category of misleading templates. 
T0 11B performs better, although it also shows no practical difference between misleading-moderate and instructive templates. 
Lastly, T0++ (trained on more datasets than other T0 variants), is the only model in this paper that shows statistically significantly different performance across all categories of prompts. 
However, there remains the caveat that it still performs arguably too well in absolute terms with pathological prompts, which we discuss in the next section. % some still outperform instructive ones

% even though statistically lower, still seems too high in absolute terms cf. what we expect a human does at zero shots
%few-shot still bad, so still not "learning from instructions".

%Notably, T0 is capable of acheving good performance with misleading templates even at zero-shot. That is, without any gradient update nor priming, out-of-the-box T0 perform NLI well as \prompt{are there similar words} and \prompt{does that have the same meaning}, etc. with no statsitical significance from the instructive templates (\rf{fig:zero-shot}). All other models, including GPT-3, perform marginally above random with all prompts at zero-shot. {\inprogress In other words, we confirm \citep{t0}'s finding that instruction tuning makes models robust to prompts, but perhaps too robust. But T0++ passes all the checks.

%comparing t0 3b to t5 3b, "too tobust"
%comparing T0 11B to T5 11B, checked one less checks
%limited improvement; and only in zero-shot, few-shot still bad
%}

\begin{table*}[]
\centering
% \vspace{-4ex}
\resizebox{0.9\textwidth}{!}{%
\begin{tabular}{@{}lllcccc@{}}
\toprule
& size & \#shots & inst. \textgreater\ mis-moderate & inst. \textgreater\ mis-extreme & inst. \textgreater\ irrelevant & inst. \textgreater\ null \\ \midrule
T0 & 3B & 0 &  &  &  &\cmark \\
T0 & 11B & 0 &  &\cmark  &\cmark  &\cmark \\
T0++ & 11B & 0 &\cmark  &\cmark  &\cmark  &\cmark \\ 
\midrule
ALBERT & 235M & 4 - 256 &\cmark & & &\cmark  \\
T5 LMA & 770M & 4 - 256 &  &  &  &  \\
T5 LMA & 3B & 4 - 256 &\cmark  &  &  &\cmark  \\
T0 & 3B & 4 - 256 &  &\cmark  &  &\cmark  \\
T5 LMA & 11B & 16  &\cmark  &\cmark  &  &\cmark  \\
T0 & 11B & 16 &  &\cmark  &  &\cmark \\
T0++ & 11B & 16 &\cmark  &  &  &\cmark \\
GPT-3 & 175B & 16 &  &  &  &\cmark \\ 
\bottomrule
\end{tabular}%
}
\caption{Checkmarks indicate where two categories of templates lead to statistically significantly different performance, as measured by an independent two-sample $t$-test and a Wilcoxon rank-sum test; both tests always agree in this table. A lack of checkmark indicates where model performance fails to differentiate the two categories, i.e., models do not understand the differences between the prompt categories. We consider significant differences (checkmarks) between categories of prompts to be necessary—but not sufficient—for language understanding.} 
%A lack of checkmark indicates where models fail to differentiate the two categories, which entails that models do not understand them. But note that a presence of checkmark does not entail that models positively understand them, as understanding consists of more criteria that is beyond the scope of our experiments.
% this is a necessary but insufficient condition for language understanding. We only claim that, by failing to satisfy it, existing models are far from understanding. We do not claim that future models which satisfy this condition automatically achieve understanding, nor do we claim that understanding is a binary variable.
\label{tab:summary}
%  \vspace{-2ex}
\end{table*}

% -----
%In both RTE and ANLI, a premise that entails a hypothesis rarely has the hypothesis also entailing the premise. However, with null templates, the prompt does not provide models with any information about which sentence is the premise and which is the hypothesis and, when [mask] is not inserted in the middle,\fn does not even provide information about how the input paragraph should be segmented into a premise and a hypothesis. 
%Despite these pathological conditions that would make NLI difficult for humans, 
%\fnt{Our tokenizer does not insert special [SEP] tokens between premises and hypotheses.}
% Once again suggesting that models are sensitive to factors that are largely invariant for human semantics. 
%Yet, how does a null template know if the first or the second sentence is the hypothesis or the premise?\footnote{Because prompt-based models use a LM objective, our tokenizer does not insert special [SEP] tokens.} And why are they so sensitive to the order of concatenation? It might be plausible to conjecture that the pretraining corpus and maybe English speakers in general tend to state the antecedent first and then the consequent, and that's why \prompt{\prem [yes/no] \hypo} (\autoref{fig:pt-null} lemon) performs almost as well as the best instructive template, but then why is \prompt{[yes/no] \hypo\ \prem} (orange) a close second? And why are the other 4 order of concatenation significantly worse (\autoref{sec:all-null})? 
%if model figure out the longer is premise, then it shouldn't matter. If not figure out, then why does matter?

\subsection{Discussion} \label{sec:summary} % talk to the reader
Recall that a common assumption in the literature is that prompts require experts to clearly and correctly describe the task at hand (\S\ref{sec:intro}). 
In contrast, \rf{tab:summary} summarizes that, with the exception of T0++ at zero shots, all models perform essentially as well with some pathological prompts as they do with proper prompts. Notably, despite being much larger than its competitors, GPT-3 shows the same patterns of behaviors, % fares worse, 
suggesting that mere scaling does not address this issue. Meanwhile, the evidence from instruction tuning is mixed. Although \citet{t0} are right that instruction tuning yields substantial improvement in performance as well as robustness as measured by variance, T0 is somewhat too robust and less sensitive to the semantics of the prompts in terms of distinguishing proper instructions from pathological ones, compared to T5 of the same size in the few-shot setting (\rf{fig:big-models}).

In the zero-shot setting, we do see that that the largest model instruction-tuned with the most datasets (T0++) improves a model's sensitivity to prompt semantics. This is a positive result, but it comes with the caveat that there still exist numerous examples of pathological prompts that perform just as well as the proper ones do. 
To be charitable to randomness in neural models, we hold this study to a higher standard by comparing means and medians among categories with statistical tests. 
Nevertheless, for our research question, existence proofs alone are still alarming. % and there exists several even for T0++. 
For example, without any gradient update nor priming, it is striking that out-of-the-box T0++ scores a high accuracy of 78\% with the extremely misleading \prompt{\prem Is that grammatically correct? \hypo}, the same accuracy as it achieves with a proper instruction \prompt{\prem Are we justified in saying "\hypo"?} 
If models were truly classifying whether the text is grammatical, it would have only scored 52.7\% because RTE is written by experts and all examples are grammatical.
Even templates that underperform the instructive ones seem to be too good. For example, it is difficult to imagine a human scoring 72\% \textit{zero-shot} with the prompt \prompt{\prem Inflections are annoying and thank god that Middle English got rid of most of them. \hypo} 
for a nuanced task like NLI. % Recall the opening example in \rf{sec:intro}; it is not at all obvious to a human how they are supposed to classify a pair of sentences when there is no task instruction

% And in the few-shot setting, it is not clear why instruction-tuned models revert back to being unable to differentiate good from bad prompts, arguably worse so than the non-instruction-tuned T5 despite achieving higher absolute performance (\rf{fig:big-models}). 

% ---
%After all, it is unlikely that there exists any misleading, irrelevant, or null prompt that makes humans learn as fast as proper instructions 
%And why does all of these differences go away for few-shot? {\inprogress challenging if models really learn from instructions. }
%Considered the evidence holistically, we hesitate to claim that instruction tuning solves everything, despite its impressive improvement.

%In contrast, for all model and dataset combination we experimented, there always exists several irrelevant or misleading template that enable the models to learn \textit{faster} than many instructive ones. 

%\todo{maybe discuss more caveats on positive evidence of instruction tuning: (1) Instruction tuning on hundreds of irrelevant prompts (task identifiers) works worse than instruction tuning on instructive prompts (FLAN Appendix B), although note that instruction tuning's training is fully supervised not few-shot like in our study. (2) maybe size + inst tuning combined will improve. Still a matter of degree. Just 2\% of difference? Say something about robustness != understanding?}

\section{Effect of Target Words} \label{sec:verbalizer}

\subsection{Method}
In this experiment, we study the effect of different LM target words given a fixed template. We write 4 categories of targets, with at least 3 pairs of target words for each category (except the singleton yes-no category): 
\begin{enumerate}
    \item Yes-no: Model is expected to predict the word ``yes'' for entailment and ``no'' for non-entailment.
    \item Yes-no-like: Semantically equivalent to yes-no but using superficially different words, e.g., ``true''/``false'', “positive”/“negative”.
    \item Arbitrary: Model is expected to predict arbitrary words that have no semantic relation to the entailment task, e.g., ``cat'' for entailment, “dog” for non-entailment.
    \item Reversed: Model is expected to predict the opposite of the (intuitive) yes-no and yes-no-like labels, e.g., ``no'' for entailment, ``yes'' for non-entailment.
\end{enumerate}
See Appendix \ref{sec:all-target-words} for the full list.
Within the arbitrary category, in addition to the common anglophone first names as \citet{le-scao-rush-2021-many} use, we also include word pairs with high semantic similarity, low similarity, and pairs which are highly frequent in the English language, but we find no consistent difference among these various subcategories of the arbitrary category. 
%\fnt{With declarative templates, another category is their template-specific targets. They are excluded from experiments in this section because combining declarative templates with other target categories yield ungrammatical prompts.} 
%We permute all 27 templates from the previous section and retrain them with 9 sets of target words for a total of 234 prompts. % using more than 30 different manually written templates and 10 sets of LM target words for a total of over 300 prompts. 

\subsection{Result}
\label{sec:verb-result}
\begin{figure}[t]
    \vspace{-4ex}
    \includegraphics[width=\linewidth]{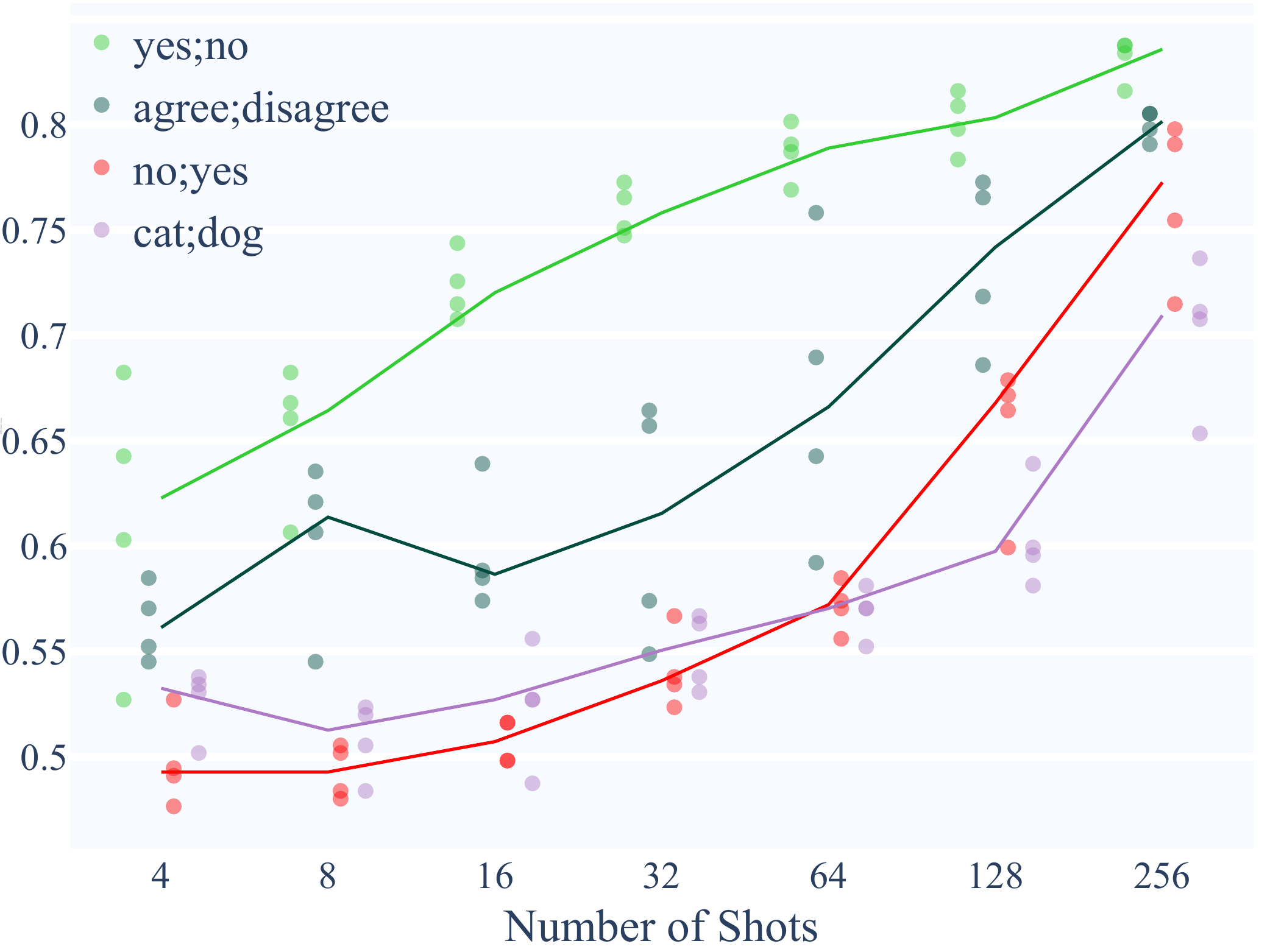}    
    \caption{The best-performing instructive template for ALBERT on RTE, \prompt{\{prem\} Are we justified in saying that "\{hypo\}"?} with select LM targets from each category.} %See \rf{sec:ceteris-paribus} for results with other templates.}
    \label{fig:justified-verb-A2}
\end{figure}

For both ALBERT and T0, we find that models trained with yes-no targets learn a good deal faster than those trained with yes-no-like targets and dramatically faster than those with arbitrary and reversed targets. For example, \autoref{fig:justified-verb-A2} shows the top-performing instructive template trained with different target words. %The large effect sizes are particularly noteworthy, e.g., 
At 32 shots, the difference between the median accuracies of “yes”/“no” vs. “no”/“yes” is $22.2\%$, far larger than the effect size of varying categories of templates in \rf{sec:template}.  % A2: 76.53, 75.09, 53.79, 53.43; 75.78 - 53.61 = 22.17
Aggregating over all combination of templates and targets, \rf{fig:targets-T0} confirms that the choice of target words matter much more than the meaning of the templates. 
%In fact, in most cases, the effect of target words far outweighs the effect of templates (\rf{fig:targets-A2}) %confirms the same trends when aggregating over all templates.  %Aggregating over all top-performing templates from each category (\autoref{fig:verb-agg}), the difference between median yes-no verbalizer (purple) and median arbitrary verbalizer (blue) are significant from 3- to 250-shot, and the effect size is often as large as 15\%. 

\begin{figure}[t]
    \vspace{-2ex}
    \includegraphics[width=\linewidth]{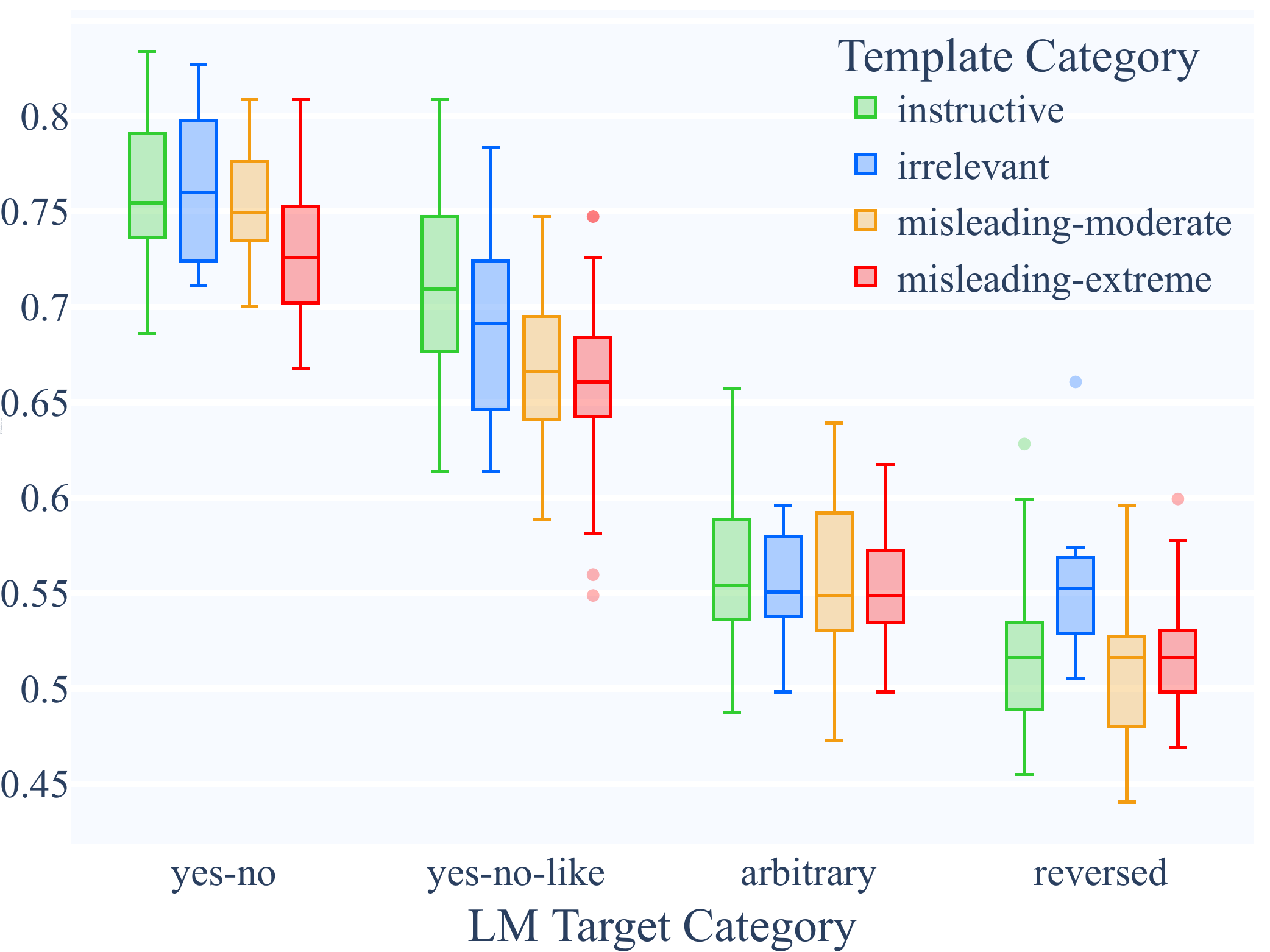}    
    \caption{T0 (3B)'s 32-shot accuracy with of all template-target combinations on RTE. In general, the choice of target words (x-axis groups) matters much more than the choice of templates (colors).}
    \label{fig:targets-T0}
    \vspace{-2ex}
\end{figure}

\begin{figure}[t]
    \vspace{-4ex}
    \includegraphics[width=\linewidth]{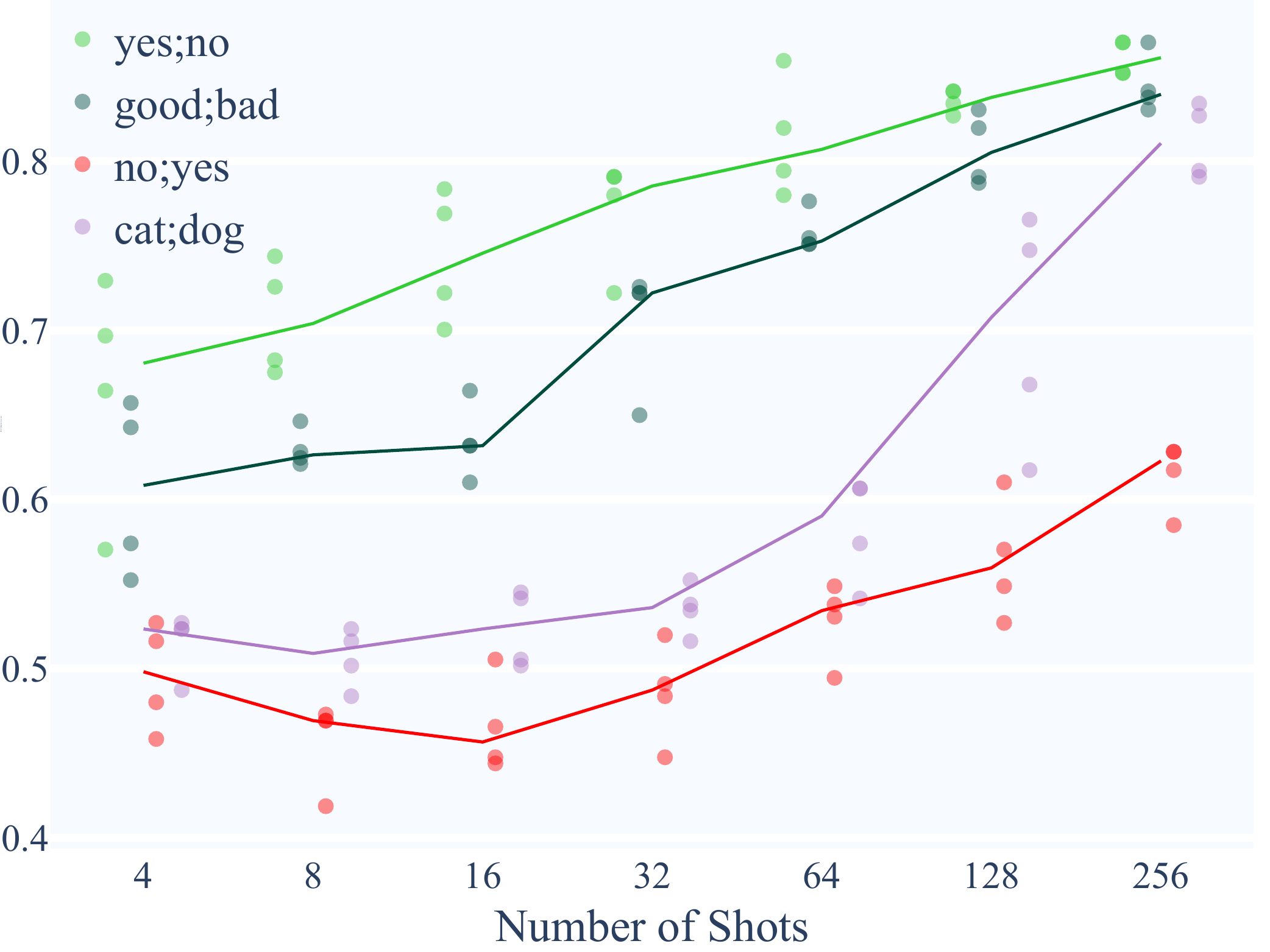}    
    \caption{The best-performing instructive template for T0 (3B) on RTE, \prompt{\{prem\} Based on the previous passage, is it true that "\{hypo\}"?} with select LM targets from each category.} %See \rf{sec:ceteris-paribus} for results with other templates.}
    \label{fig:justified-verb-T0}
    \vspace{-1ex}
\end{figure}

\subsection{Discussion}
\begin{figure}[t]
    \vspace{-2ex}
    \includegraphics[width=\linewidth]{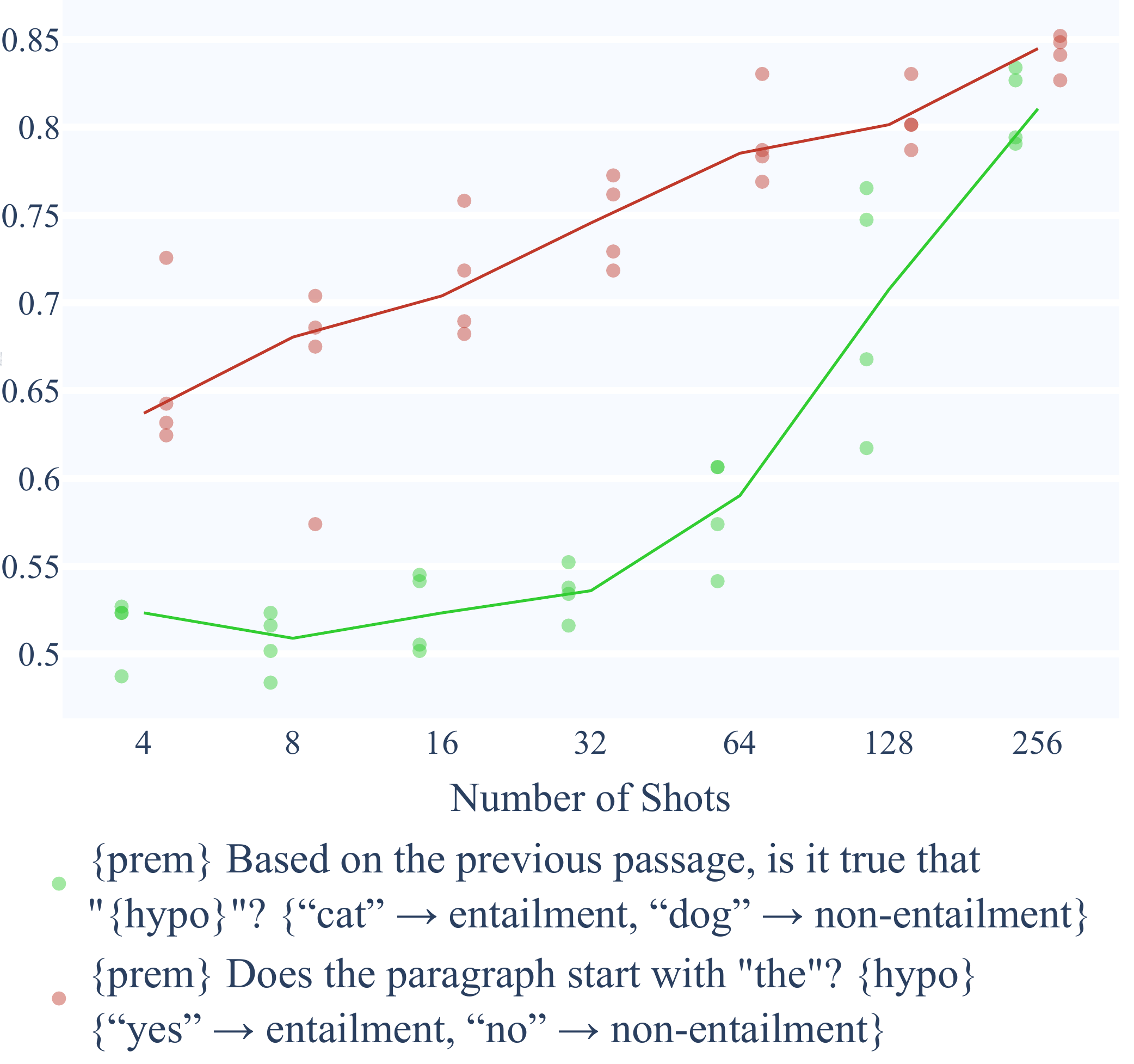}    
    \caption{T0 (3B) on RTE. Misleading templates + yes-no targets (red) learn substantially faster than instructive templates + arbitrary targets (green), which is the opposite of what we expect from humans.}
    \label{fig:verb-override}
    \vspace{-2ex}
\end{figure}

% E: I'd say state the positive results more directly, less qualified. otherwise, it comes across as biased. like we are trying to sweep the positive results under the rug.
The fact that models consistently learn slower with arbitrary and reversed target words is a positive result: this type of performance differential is consistent with what we expect for models that are correctly sensitive to the semantics of the words.
However, there are several important negative results in these experiments as well. First, the effect of the target words overrides the semantics of the overall prompt. Consider two kinds of template-target combinations:
\begin{enumerate}
     \item An irrelevant or misleading template + yes-no targets, e.g., \prompt{\prem\ Does the paragraph start with "the"? [yes/no] \hypo}
     \item An instructive template + arbitrary targets, e.g., \prompt{\prem Based on the previous passage, is it true that "{\hypo}"? [cat/dog]}  % good;bad also work
\end{enumerate}
\autoref{fig:verb-override} shows that combinations such as (1) often dramatically outperform (2). % which is the opposite of what we expect because (1) is a pathological condition under which we would expect that a human would be confused and would still need a large number of trials and errors to figure out what is the actual task. In contrast,
However, (2) simply requires figuring out a mapping: “Reply ‘cat’ if entailed and reply ‘dog’ if not entailed”.
For humans, this can be learned in a few shots, e.g., \citet{ferrigno2017universal} showed that adults can reach 60\% accuracy in 18 trials\fn for an arbitrary map of \{more numerous $\rightarrow$ star shape, less numerous $\rightarrow$ diamond shape\} without receiving any language instructions. In contrast, models under many arbitrary LM targets struggle to reach 60\% median accuracy even by 64 shots with instructive templates (\rf{fig:verb-override} green; \rf{fig:justified-verb-A2} red, purple). 
\fnt{And this comparison is heavily charitable to the models because “18 trials” means that humans see 18 examples for 18 times in total, whereas “20-shot” means that models can see the same 20 examples over and over again for many epochs.} 

Further, even given intuitive yes-no-like targets such as “agree”/“disagree” and “good”/“bad”, models learn much slower compared to when given “yes”/“no”. As \autoref{fig:justified-verb-A2} (green vs. dark green) and \rf{fig:targets-T0} (first vs. second x-axis group) show, there exists a large performance gap between yes-no and yes-no-like targets which is not closed until 256 shots. %whereas for humans, the difference between answering “yes”/“no” vs. answering “true”/“false” should be trivial and likely would not require 128 or more examples to close any gap 
Moreover, when we try to help the models by appending target hints such as “True or false?” to the templates, performance often drops instead,
echoing \citet{t0} and \citet{flan}'s findings that including answer choices in input sequence make models perform worse for certain tasks.
% whereas for humans, including answer choices should only increase or be neutral performance
% “reply ‘cat’ if entailed and reply ‘dog’ if not entailed”

%\comment{need some signposting here} As for the effect of the target words, if models understand their meaning, we would expect that:
%\begin{align}
%    \text{yes-no} &= \text{yes-no-like} \\
%    \text{yes-no} &> \text{arbitrary} \\
%    \text{yes-no} &> \text{reversed}
%%    \text{yes-no + irrelevant} &> \text{misleading + yes-no-like}
%%    not \text{any template + yes-no} &> \text{any template + yes-no-like}
%\end{align}
%
%Although models satisfy (5) and (6), they all fail to satisfy (4), i.e., asking the models to answer “yes”/“no” vs. “true”/“false” makes a substantial difference, often much greater than the effect of the templates (\rf{fig:verb-override} and \rf{tab:16}). Moreover, such target word bias is not only not fixed by appending templates with target word hints, e.g., “reply ‘true’ if entailed and reply ‘false’ if not entailed”, they in fact consistently hurt performance, even with simple hints like “yes or no?" when “yes”/“no” targets are expected.

% section titles are for people who skim the paper
% Alternatives for friendly !> indifferent
\section{General Discussion} %and Future Directions} 
\label{sec:discussion}
\subsection{Summary and Interpretation}
% some positive results from target words, 0-shot large models, but we still think. 
% Existence of success pathlogical prompts is problematic. 
Our main research question is whether models understand prompts as meaningful task instructions analogous to how humans would.
Again, suppose an experimenter provides a human annotator with an informative instruction of a reasonably easy task.
If the annotator understands the instruction, we expect them to perform better than when the experimenter provides misleading instructions, irrelevant instructions, or no instructions at all.
\rf{sec:template-results} shows that the performance of most models is insensitive to the difference between instructive and irrelevant templates, moderately sensitive between instructive and misleading templates, and highly sensitive between instructive and null templates. 
Comparing to the effect of the templates, however, \rf{sec:verbalizer} shows that models are much more sensitive to the semantics of the target words: they learn far slower with arbitrary or reversed target words as desired. However, they are overly sensitive to semantically equivalent yes-no-like words (i.e., performing much worse with “agree”/“disagree” than with “yes”/“no”), and the choice of target words override the semantics of the templates (e.g., performing much better given a irrelevant template with “yes”/“no” targets than with an instructive template with arbitrary targets such as “cat”/“dog”).

Our main argument throughout the paper shares the same logic as a recent line of studies \citep{sinha2021unnatural,oconnor2021context,pham2021order,gupta2021bert} which argue that the fact that LMs achieve good performance under ideal conditions is insufficient to establish language understanding because they also succeed under pathological conditions (e.g., sentences with shuffled word order) where humans fail catastrophically.\footnote{See \citet{ravishankar2022word}, \citet{papadimitriou2022classifying}, and \citet{kulmizev2021schr} for a nuanced ongoing debate on the extent models know vs. use syntactic coding properties on what kinds of examples. But even considering these new evidences, we think \citet{sinha2021unnatural} are at least correct that, as they find that human experts perform far worse on shuffled NLI inferences than RoBERTa does, models must be processing linguistic inferences quite differently from how humans do, regardless of whether models know word order information.}
In other words, the fact that models are so good at inferring the gold labels from pathological inputs casts major doubts on whether models make inferences in any way that resembles how humans make inferences. 
% In this paper, pathological instructions, and pathlogical target words. On instructions, few-shot always bad, zero-shot successs only largest, but still existence proof. Humans fail catstrophically.
% For pathlogical target words, models have positive results, but shows outsize effect, higlights how much instruction templates don't matter.
For our results, the fact that models are so good at learning from pathological instructions likewise casts major doubts on whether models understand prompts as instructions in any way that resembles how humans understand instructions.

\subsection{Alternative Interpretations and Future Directions}
\label{sec:alternatives}
As with any extrinsic evaluation, accuracy cannot directly measure understanding. %and it is worth asking whether the observed negative results might be attributable to something else. %other than a lack of understanding of the prompts. 
% In addition to a lack of understanding, we discuss the extent to which the observed negative results may be attributed to other confounding factors.
For example, a human could perfectly understand an instruction but still, e.g., have the same accuracy with instructive vs. irrelevant templates because the task itself is too hard (a lack of competence) or because they for some reason ignore the instructions (a lack of compliance). We discuss these two possibilities below. %  We consider both alternatives as explanations for the behaviors observed in our experiments, but ultimately conclude models' behavior should still be mostly attributed to a lack of understanding of their given prompts.

\paragraph{Lack of Competence} % in NLI per se, not in NLI instructions
This is primarily a concern for non-instruction-tuned models at zero shots, where all models perform only slightly above random, and thus a lack of statistical significance among template categories is ambiguous as to whether models lack understanding of NLI instructions vs. if models lack the competence in NLI per se.
This is why our study largely focuses on the few-shot setting, where a lack of competence is less of a concern, as models do competently achieve good accuracies that are only moderately below the state-of-the-art non-few-shot models. 
% also, humans sometimes misinterpret instructions at zero shots;  crowdsourcing & cogsci experiments almost always provide familiarization examples. So few shots are "better" evaluations.

Another counterargument is that maybe no models ever actually reason about if a premise entails a hypothesis. Maybe they just always exploit spurious or heuristic features %\footnote{FWIW, in preliminary experiments, our models are competitive on HANS.} Although you could say ANLI also measures against heuristics, and all models are still pretty bad at ANLI in absolute terms.
and, if only they were competent in properly reasoning about entailment relations, then the meaning of NLI instructions would matter. 
This argument is possible, although, first, it hinges on to what extent NLI (or any other behavioral evaluation) can measure language understanding, which is a complex debate beyond the scope of this paper. 
Second, in preliminary experiments (\rf{sec:hans-preliminary}), our models actually zero-shot transfer reasonably well to HANS \citep{mccoy}, a dataset designed to diagnoses models use of NLI heuristics. \nocite{utama-etal-2021-avoiding} %\fnt{We few-shot train models on RTE and zero-shot evaluate them on HANS, as it is commonly done with diagnostic datasets. Their HANS performance closely tracks their RTE performance.} % Contra Utama et al. 2021 Avoiding Inference Heuristics in Few-shot Prompt-based Finetuning
% Lampinen shows capable of reasoning beyond heuristics?
Thus, it is unlikely that models are entirely incompetent in reasoning about entailment relations and solely rely on heuristics. 
% Further, the argument “models never actually reason about entailments” also does not explain why entailment instructions and irrelevant templates outperform misleading and null templates.
Regardless, further differentiating competence in understanding task instructions vs. competence in tasks per se is an important direction for future work.

\paragraph{Lack of Compliance}
Another interpretation is that irrelevant prompts perform the same as the instructive ones because models simply ignore the prompts altogether. %This explanation is also suggested by prior work on generative tasks which found that models often generate text unrelated to, and thus possibly ignore, their prompts \citep{efrat2020turking,mishra2021natural}. 
However, a lack of compliance alone cannot explain our results. If models truly ignore the prompts, we should not see any systematic differences between any categories of prompts. Instead, we do see consistent patterns that instructive and irrelevant templates make models learn significantly faster than misleading and null templates do (\rf{tab:summary}).
% (2) instructive prompts suppress the effect of punctuation that plays a large role in irrelevant prompts. 
% (3) yes-no targets make models learn faster than with arbitrary and reversed targets.

%\paragraph{task identifier} For humans, both irrelevant and null templates provide zero useful information. Yet, for models, irrelevant templates consistently outperform null templates. One possible explanation is that encoder-decoder and decoder-only models are designed to use a task identifier (which can be semantically meaningless tokens such as  \texttt{mnli: }) appended to each data example. Models may be using the extra tokens from the irrelevant templates as some kind of task identifier, which are not available in the null templates. However, it is not the case that all tokens serve as equally good task identifiers, as shown by the performance gap among the irrelevant, misleading, and instructive templates. The fact that such gap exists but their inequalities do not match that of humans (criteria 1 - 3) is precisely why we argue models do not understand the meaning of prompts as humans do (\sec{alternatives}).
%Therefore, models must be affected by/comply with the prompt somehow, but they are sensitive to prompts in ways different from how humans would follow the prompts. In other words, the question is not \prompt{whether} they comply with the instructions, but \prompt{how} how they follow the instructions, i.e., a question of understanding. 

A more nuanced counterargument is that although models do not ignore their prompts entirely, perhaps it “takes less effort” for models to use the spurious or heuristic features for predictions as opposed to the more complex syntactic or semantic features \citep{lovering2021predicting,warstadt-etal-2020-learning} required to properly comply with the instructions. However, spurious features alone likewise cannot explain the observed performance gaps. Recall that, within each random seed, all models see exactly the same training examples (with the same spurious features). 
Thus, to the extent that models perform differently with some prompts compared to others, it may be due to some complex interactions between the (spurious or semantic) features in prompts and the spurious features in data examples. One possible example of this interaction is that punctuation has a large effect for irrelevant templates, but instructive templates seem to be able to suppress such effect (\rf{sec:qmarks}). Investigating the nature of this interaction is a promising direction for future work, and it suggests a way in which the semantics of the prompt might matter, e.g., by affecting the models' inductive biases, even if models do not interpret or use the instructions in the same way as humans would.

\section{Conclusion} %Takeaways
In this study, we train several prompt-based models with over 30 manually written templates and 13 sets of LM targets 
for NLI. We find that models often learn equally fast with misleading and irrelevant templates as they do with instructive ones,
and that the choice of the target words overrides the meaning of the overall prompts. 
This is true for all models and datasets with which we experimented in the few-shot setting. Despite the mixed evidence in the zero-shot setting with instruction-tuned models, overall, these results contradict a hypothesis commonly assumed in the literature that prompts serve as semantically meaningful task instructions and that writing high-performing prompts requires domain expertise.
Although we find that existing models are far from fully understanding the meaning of their prompts, we agree that learning from instructions is an important research direction, and we propose several future directions of investigating models' understanding of the meaning of prompts.

\section*{Ethical Considerations}
% appear to be intepretable, invite lay-person
% they don't do what they do
% especially in sensitive works, where transparency and accountability is important
The fact that even the largest LMs \textit{appear} to follow yet do not actually follow users' instructions has important implications, especially considering the increasing commercial use of LMs. 
% For example, an employer could prompt a model: \prompt{Based on the candidate's resume, and please don't be racist, should we hire this person?}
% A model may perform well with this prompt, but it may perform equally well with the prompt \prompt{Based on the candidate's resume, and please be grammatically correct, should we hire this person?}
While traditional fine-tuned models also pose challenges in interpretability, with prompt-based models, an illusion of instruction following can be more pernicious than having no instructions at all. The intuitive interface that prompts provide might make them more accessible to lay users, and can mislead users to think that their instructions \textit{are} being understood and followed. %, upon seeing high performance with “expert-crafted” prompts, 
%reducing the users' tendency to analyze the biases or other ethical concerns of model predictions; %(see, for example, OpenAI's broad claim for its commerical API in Appendix \ref{sec:gpt-api}) 
%and it is extra complicated by our finding that, again, models do not simply ignore the prompts, but that the semantics of the prompts affect model decisions in unpredictable ways.
%While we do not expect every LM user to invest in fine-grained studies of exactly how their models reach their decisions, 
%Users should be made aware that just because they prompt models with ethical instructions and observe high performance does not entail that the models are in fact acting ethically. 
%While we are inclining to agree with the hypothesis that newer models may have acquired a cognitively sophisticated method of reasoning with prompts \citep{lampinen2022,palm}, independent of whether they are human-like.
%In applications of LMs where ethics is critical, 
Our results suggest that cautions are needed even more than they were with traditional fine-tuned models.

%Sewon Min, Andrew Lampinen. Still, even with chain-of-thought explanations of priming examples, the fact that models do not  

%Some interpret overt robustness to prompt wording and high accuracy as a good thing (Logan and Schick 2), and they are right for certain applications

%As large models increasingly unable to be fine-tuned, zero-shot instructions (optionally with priming) are popular (cite all recent large models?) The use of instructions also are critical. 
%
%especially with GPT-3, they sell an API, widely for commercial use, and has a dedicated section and even customer support on prompt engineering, perpetuating the myth of expert knowledge and understanding, when heuristic association with performance (Appendix A).

%claims of models being unbiased, passing evaluations of, while perniciously reasoning in other associationism. 

\section*{Acknowledgments}
We are grateful to Colin Raffel, Victor Sanh, Sasha Rush, Stephen Bach, Roman Feiman, Teven Le Scao, Ian Tenney, Dan Garrette, Jason Wei, Satoshi Sekine, Mike Tien-Chien Chiang, Xavier Fontaine, Pierre Colombo, Ryan Teehan, Debajyoti Datta, William Rudman, Ruochen Zhang, Daniel Cohen, George Zerveas, Eric Rosen, Kaiyu Zheng, Nihal Nayak, Roma Patel, Charles Lovering, Tian Yun, Jack Merullo, and Aaron Traylor for comments and discussions on early drafts of this paper. Special thanks to Victor, Colin, and Teven for technical clarifications and code review. 

Furthermore, Albert is indebted to Colin and Sasha for their patience on the many iterations of the zero-shot \autoref{fig:zero-shot} as well as invaluable mentorship throughout the T0 project.

\bibliography{custom}

\clearpage
% \vspace{2ex}
\tableofcontents

\clearpage
\appendix

\begin{figure}[t]
    \centering
    \includegraphics[width=\linewidth]{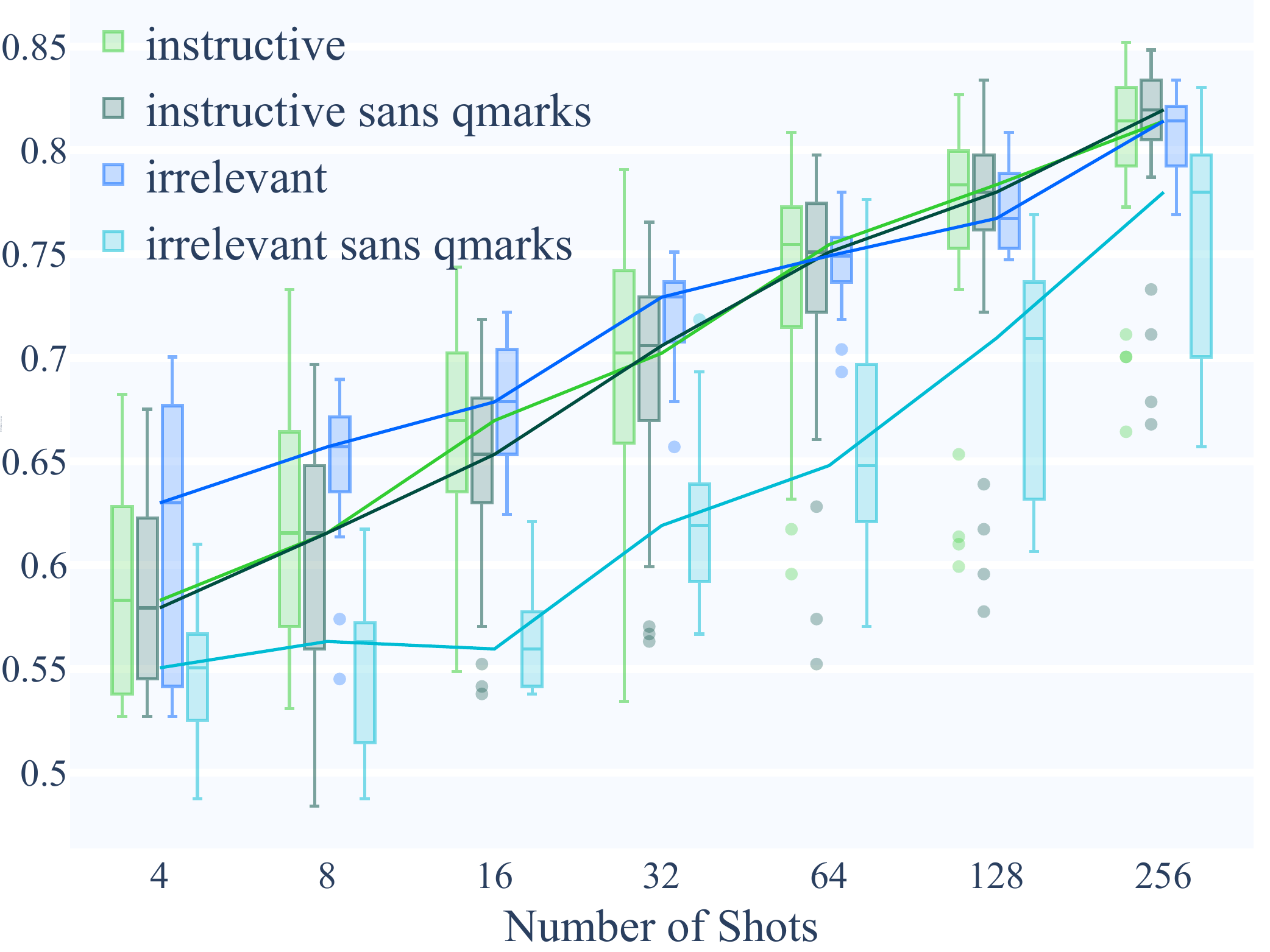}
    \caption{ALBERT on RTE. Note that (1) irrelevant templates slightly outperform the instructive templates, albeit without statistical significance. (2) Irrelevant templates are far worse without quotation and question marks. (3) But there is no significant difference between instructive templates with or without qmarks.}
    \label{fig:qmarks-A2}
\end{figure}

\begin{figure}[t]
    \centering
    \includegraphics[width=\linewidth]{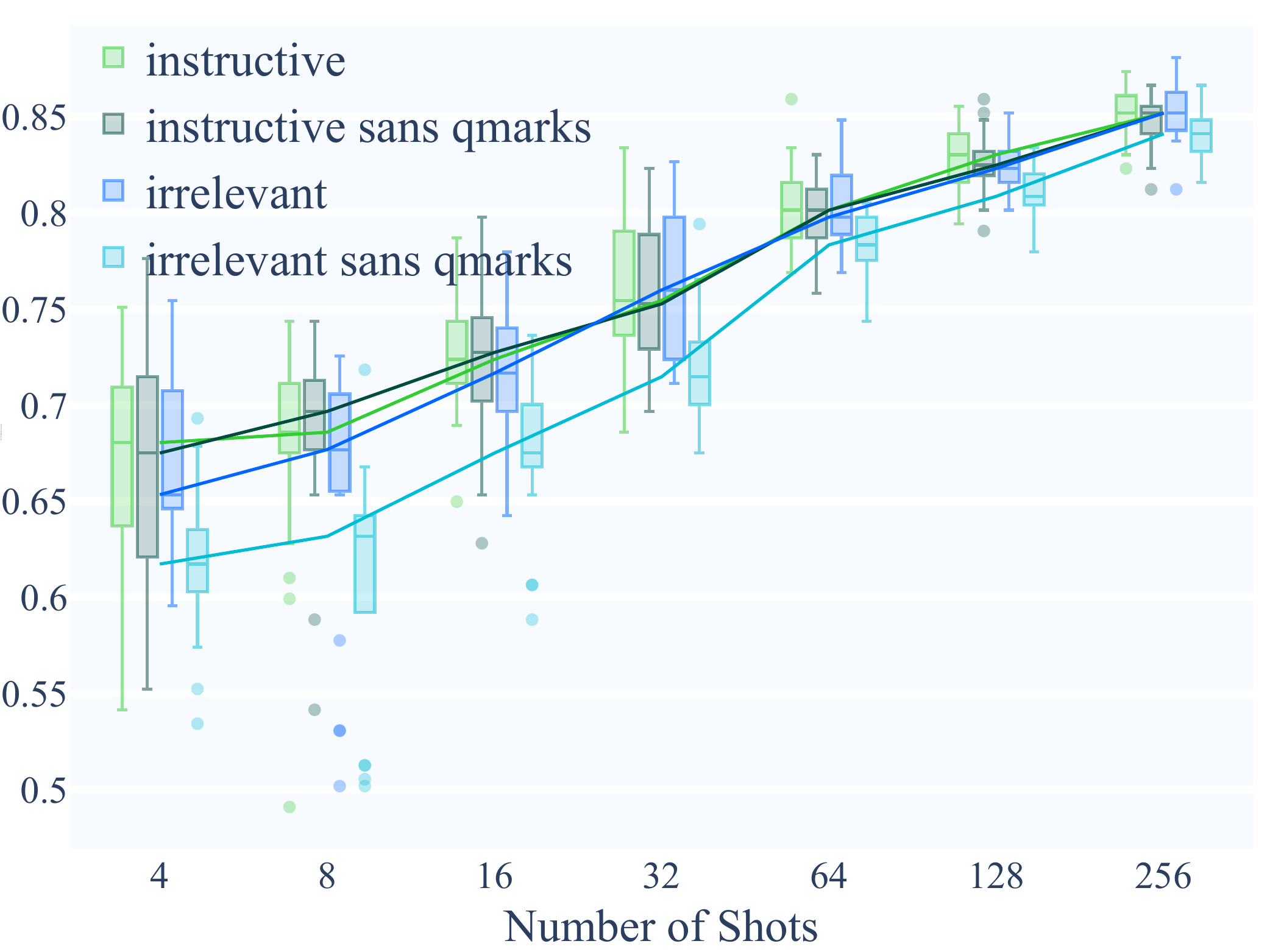}
    \caption{T0 (3B) on RTE. Like ALBERT, irrelevant sans qmarks are significantly worse than irrelevant at each and every shot, but there is no significant difference between instructive with or without qmarks.}
    \label{fig:qmarks-T0-3B}
\end{figure}

\begin{figure}[t]
    \centering
    \includegraphics[width=\linewidth]{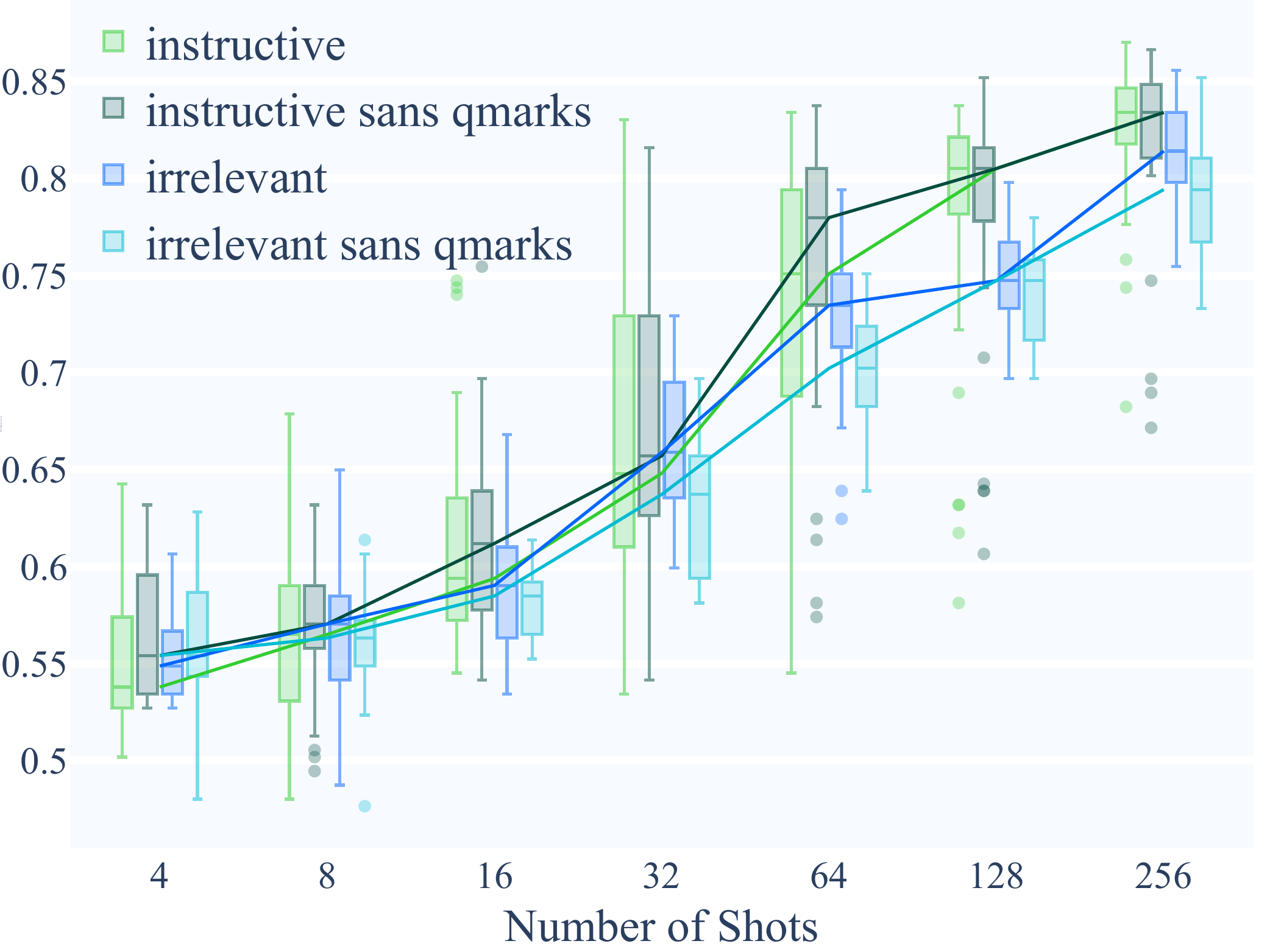}
    \caption{T5 LM-Adapted (3B). Unlike the other models, there is no statistical significance between irrelevant with or without qmarks. However, instructive sans qmarks statistically significantly outperform instructive at 32 and 64 shots.}
    \label{fig:qmarks-T5-3B}
\end{figure}

\section{Effect of Punctuation}
\label{sec:qmarks}
%Unlike the misleading templates, which at least ask the model to perform \textit{some} NLP task, irrelevant templates only provide utterly irrelevant chitchat, often not even asking any question. 

For irrelevant templates, we find a large effect from the use of quotation and question marks in templates. It is natural to write such punctuation in instructive templates as they help humans to parse an NLI hypothesis as an embedded clause within an instruction sentence (e.g., \prompt{Given \prem Should we assume that "\hypo" is true?}). 
For control, we also use quotation and question marks (“qmarks” hereafter) in irrelevant templates where they would not have made sense naturally, e.g., \prompt{\prem Single-family zoning is bad for American cities. "\hypo"?} As an ablation, when we remove these qmarks from irrelevant templates, the performance of ALBERT and T0 drops substantially (Figures \ref{fig:qmarks-A2} and \ref{fig:qmarks-T0-3B}). In contrast, for T5, qmarks make no difference for irrelevant templates; yet, removing qmarks from instructive templates—where qmarks are natural—boosted performance instead for T5 (\rf{fig:qmarks-T5-3B}), but not for T0 nor ALBERT.

Additionally, as a coincidence, most misleading templates contain both quotation and question marks, while most misleading-far templates contain only question marks (\rf{sec:all-templates}). But as noted in \rf{sec:misleading}, there is no consistent pattern between those two misleading categories. In other words, punctuations alone cannot explain everything. As discussed in \rf{sec:alternatives}, the full explanation is likely a combined interactions between the spurious features and the semantics of the templates. 

Lastly, note that \citet{schick2} and many subsequent papers' prompts for NLI (e.g., \prompt{"\hypo" ? | [mask]. "\{premise\}"}) are basically null templates with some variation in punctuation between the hypothesis and the premise. We find that models learn poorly with the vanilla \prompt{\hypo\ [mask] \{premise\}}, but they learn as fast as the instructive templates with Schick \& Schütze's punctuated version. %(see Appendix \ref{sec:preliminary} for detail results).
That being said, note again that punctuation alone cannot explain the performance gap, since models trained with \prompt{[mask] \hypo\ \prem} (\rf{fig:pt-null}, pink) perform second to best, %even though no [mask] or any template punctuation segment their premises and hypotheses. 
yet swapping their premises and hypotheses (\rf{fig:pt-null}, purple) makes it the worst performing among all null templates. % we fix prem, template, hypo throughout this paper

\section{Details and Lessons from Experimenting with GPT-3's API}
\label{sec:non-profit}

\subsection{Choice of Model} \label{sec:gpt-api}
We use the \texttt{davinci} model provided by OpenAI LP's API, which corresponds to\footnote{OpenAI never actually discloses which one of their commercially named \texttt{ada, babbage, curie, davinci} “engines” correspond to models of which size. However, \citet{eval-harness} estimate that they correspond to 350M, 1.3B, 6.7B, and 175B respectively.} the 175 billion parameter model reported in \citet{gpt}. Concurrent to our work, OpenAI released a new product called the “Instruct Series”, 
%which was speculated by \citet{t0} and \citet{flan} as an instruction-tuned version of GPT-3. 
%While it would be interesting to study an instruction-tuned 175B model,%\footnote{Especially since \citet{flan}'s 137B FLAN is not publicly available.} 
but we decided to not experiment with the Instruct Series because no academic paper or technical documentation of any kind is available with the Instruct Series at the time of writing aside from the following claim on their website:\footnote{\url{http://beta.openai.com/docs/engines/instruct-series-beta}}

\begin{quotation}
The Instruct models share our base GPT-3 models’ ability to understand and generate natural language, but they’re better at understanding and following your instructions. You simply tell the model what you want it to do, and it will do its best to fulfill your instructions. This is an important step forward in our goal of building safe models that are aligned with human interests.
\end{quotation}

Crucially, the Instruct Series is inappropriate for reproducible research because it is unknown what datasets and prompts these models are trained on, and whether any task categories are systematically held out as done by \citet{t0} and \citet{flan}. If it is trained on any prompt or dataset of NLI, it would not be zero-shot, making it an unfair comparison to other models in our experiments. Second, it is still in beta and its training, held-out, and prompt mixtures could change. At least two Instruct Series models were made available in sequence during our writing, and it is not clear if we experiment on an older version, whether it will still be available and reproducible in the future. 

\subsection{Priming vs. Fine-Tuning}
As mentioned in \rf{sec:setup-gpt}, we use priming (a.k.a. in-context learning) in lieu of fine-tuning because, at the time of writing, OpenAI's fine-tuning API is limited to 10 runs per month. To train 30 prompts at only two number of shots would take 6 months, assuming we get hyperparameters right at first try. Further, each training run is limited to a maximum of 5 epochs, which often entails an insufficient number steps for few-shot training. We were unable to fine-tune GPT to any reasonable accuracy with our allowed 10 tries in the first month. Finally, the fine-tuning API is limited to GPT variants up to 6.7B, not the 175B model we plan to experiment with.

With priming, we are able to reproduce \citet{gpt}'s zero-shot performance on RTE but only with their exact prompt reported in their Figure G.31, all other (even instructive) prompts perform at random at zero shots, suggesting that their reported prompt is highly cherry-picked. We are unable to reproduce their reported few-shot result because they report it at 32 shots, but their API only permits a context length up to 2049 tokens, which is insufficient for RTE. We find that 16 shots are the highest one can reach within the token limit.\footnote{Depending on the length of the prompt template, 2 or 3 examples still exceed the token limit, in which case we remove one priming example, keeping the other 15 priming examples and the to-be-predicted example unmodified.}

Like the gradient updated models, we document the exact examples we use for few-shot priming in our GitHub repository. Unlike the gradient updated models, which are trained on the same $k$ examples, priming models use different sets of $k$ priming examples for each inference example \citep[p. 20]{gpt}. This means that GPT's performance reflects the fact that, overall, it has seen far more than $k$ examples, making it not directly comparable to the few shots of the gradient updated models. This is not ideal, but our GPT few-shot performance already underperforms what \citet{gpt} report, so we choose to not further restrict it to have the same fixed priming examples for all inference examples, which could run into a lack of competence issue (\S\ref{sec:alternatives}) that make its results unusable for our research question. 

Lastly, unlike the gradient updated models, we do not run multiple seeds with our GPT experiments because, first, they are expensive. As the API bills by token, using $k$ shots of priming example effectively multiplies the total cost by $k$. Second, OpenAI imposes a monthly quota for each lab, so running multiple seeds will take several more months to complete.

\subsection{Other Tips for Working with GPT-3}

Using the \texttt{logprobs} argument in their API, we obtain the top 99 predicted target word and their log probabilities.\footnote{Although sometimes the API returns less than the number of \texttt{logprobs} the user specifies, in which case we contacted OpenAI's customer support who provided us refund by store credit. At the time of publishing, OpenAI now restricts \texttt{logprobs} to a maximum of 5.} 
Following \citet{t0} and \citet{flan}, we evaluate by a rank classification of the target words, i.e., if the gold target word is “yes”, we consider it as correct as long as the probability of “yes” is higher than that of “no”, regardless of whether “yes” is the top-1 prediction generated by the model. 

Alarmingly, we find that these top-99 predictions are semantically inconsistent ranked, e.g., for one data example and its top-99 word predictions, it is often the case that, e.g., P(yes) > P(no) but P(Yes) < P(No). Thus, the choice of the target words' surface form makes a substantial difference in the overall performance. (Not to mention the problem of choosing between yes/no, true/false, correct/incorrect, etc. as studied in \rf{sec:verbalizer}.)
OpenAI recommends having no trailing space in the input and let the model predict the first token with a leading space as in “\textvisiblespace Yes”. We find that although stripping the leading space sometimes leads to higher performance for some prompts, overall not applying stripping or other token normalization performs the best.
%the token with a leading space and  its first letter capitalized (but not all capitalized, which is a different token) performs the best, so we follow OpenAI's recommendation and %use “\textvisiblespace Yes” and “\textvisiblespace No” as target words of rank classification 

Another point researchers should pay attention to is the use of what OpenAI calls a “separator” inserted between priming examples. In preliminary experiments, we initially use newline characters as appeared in \citet{gpt}'s Appendix G. We later discover that OpenAI recommends using \verb|###| or \verb|\n###\n| as separators. We use the latter and find consistent performance improvement over just using newline characters, and we use it throughout in our main experiments.

%Finally, for researchers interested in reproducing GPT-3's results on other datasets, we refer readers to \citet{t0}'s Appendix D3 and \citet{flan}'s Appendix A2 for additional evaluation tricks used for other datasets.

%Cost, compute

%---

%We find three major obstacles in working with OpenAI's GPT-3 API: lack of documentation, restrictive quota, and high cost. They may have merits (e.g., quota and approval before launching a business, beyond scope of this paper), but our discussion focuses on academic research. A problem of expectation. Is this a commercial service or an academic standard to which papers in the field are expected to be compared to?

%Of course, not recorded exact indices of priming examples. Could have sampled multiple times. As they did in cherry picking their prompts. The retrieval paper found the exact priming examples matter a lot. So do us.
%
%\subsection{Lack of Documentation}
%%Although impression that GPT-3 is a model which new papers are expect to compare to, really it is a commercial service. 
%We use the Base series “Curie” engine.
%
%no hyperparameters. Sampling. 

% tokenize “neither” as 2 tokens despite paper reporting, 
%whatever is most available tokenized as a single token

%API bug rank eval != response-chosen token. Contact customer support for store credits and rerun experiments.

%\subsection{Cost}
%Scaled by dataset size, priming examples, and multiple inference for rank eval.

\section{Hyperparameters}
For encoder-only models, we follow \citet{schick2} and \citet{le-scao-rush-2021-many}'s recommendations and use a learning rate of $1e^{-5}$. For T5 and T0 models, we follow \citet{t5} and \citet{t0}'s recommendations and use a learning rate of $1e^{-4}$. We run several preliminary experiments with learning rates ($3e^{-4}, 1e^{-4}, 5e^{-5}, 1e^{-5} $) deviating from their recommendations and they perform worse, although our search is not exhaustive due to the high cost of running multiple prompts with multiple random seeds. 

Note that T5 and T0 are trained with the Adafactor optimizer \citep{adafactor} in Mesh TensorFlow. Our implementation is in PyTorch, and we find that fine-tuning T5 with PyTorch's implementation of Adafactor yields substantially worse results than the usual choice of the AdamW optimizer. We corresponded with \citet{t5}, who advised us that it might be due to the fact that PyTorch does not have the same learning rate scheduler implementation as TensorFlow's Adafactor does. They recommended us to simply use AdamW, which is what we did. This is somewhat unfortunate because Adafactor is much more memory efficient, which would have drastically reduced the compute resources required and thus enable more comprehensive experiments of the 11B models, which are currently limited to 0 shots and 16 shots only. 

Although most models seem to obtain the highest validation accuracy at very early epochs, we train all models to 30 epochs (20 epochs for 11B models) to be safe and select the checkpoint with the highest validation accuracy. 

All models use a batch size of 4 with 4 gradient accumulation steps for an effective batch size of 16.

Note that because we use a rank classification of single-token target words, decoding sampling methods (e.g., beam search, top-$k$, top-$p$) are unnecessary.

We follow \citet{t5} and add EOS tokens for input sequences, which yields higher few-shot performance compared to not adding EOS as done by \citet{t0}. However, we omit EOS in the zero-shot setting, which exactly reproduces the results reported by \citet{t0}. See T0's GitHub repository readme\footnote{\url{https://github.com/bigscience-workshop/t-zero/tree/master/examples}} for more information.

%We use random seeds $1,2,3,4$, which are especially important for controlling the exact few-shot examples used for training (\rf{sec:random-seeds}). For RTE, these seeds should yield a \texttt{starting\_example\_index} of $550,231,974,966$ respectively.

\section{Compute Used}

Each ALBERT 235M model is trained on a single Nvidia RTX3090. Their main experiments took approximately 192 GPU hours. % 24 hrs * 4 gpus % bumped to 48 for Sec 5 
%A single run (with one prompt, one number of shots, and one random seed, on RTE) takes approximately X hours. In total, the main experiments took Y GPU? hours.

Each T5 LMA 770M model is trained on a single A6000. Their main experiments took approximately 48 GPU hours. % 12 hrs * 4 gpus

The 3B models are each trained by partitioning their layers over four RTX3090s. T5 and T0's main experiments took approximately 2,304 GPU hours in total. % 48 hrs * 16 gpus * 2 models % bumped to 72 hours for Sec 5

The 11B models are each trained on eight V100s (each with 32GB of memory). T5, T0, and T0++'s main experiments took approximately 1,728 GPU hours in total. (Due to their large GPU memory requirement, we were only able to complete one number of shots.) % 72 hrs * 8 gpus * 3 models

%\clearpage
%\newpage
\begin{figure}[h]
    \section{Additional Figures Discussed in the Main Text} \label{sec:baseline-figs}
%    \vspace{-4ex}
    \centering
    \includegraphics[width=\linewidth]{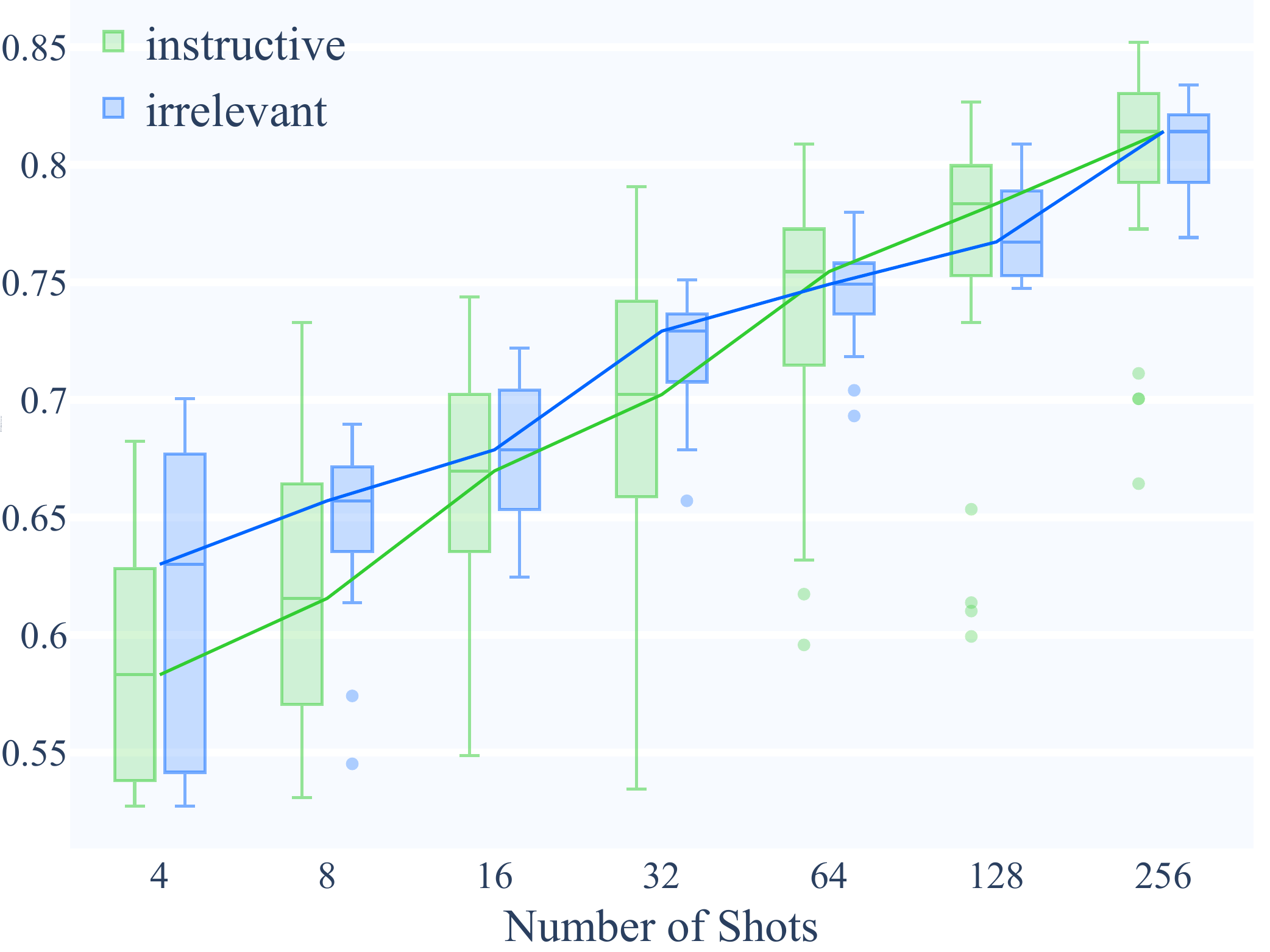}
    \caption{ALBERT on RTE. Models trained with irrelevant templates actually slightly outperform the instructive templates, albeit without statistical significance at any number of shots.}
%    ALBERT (235M) on RTE. Models trained with misleading-far templates learn as fast as those trained with instructive templates with no statistical significance at any number of shots. In contrast, models trained with misleading templates are significantly worse than the instructive ones from 16 to 32 shots.
    \label{fig:irrelevant-A2}
\end{figure}

\begin{figure}[h]
%    \vspace{-4ex}
    \centering
    \includegraphics[width=\linewidth]{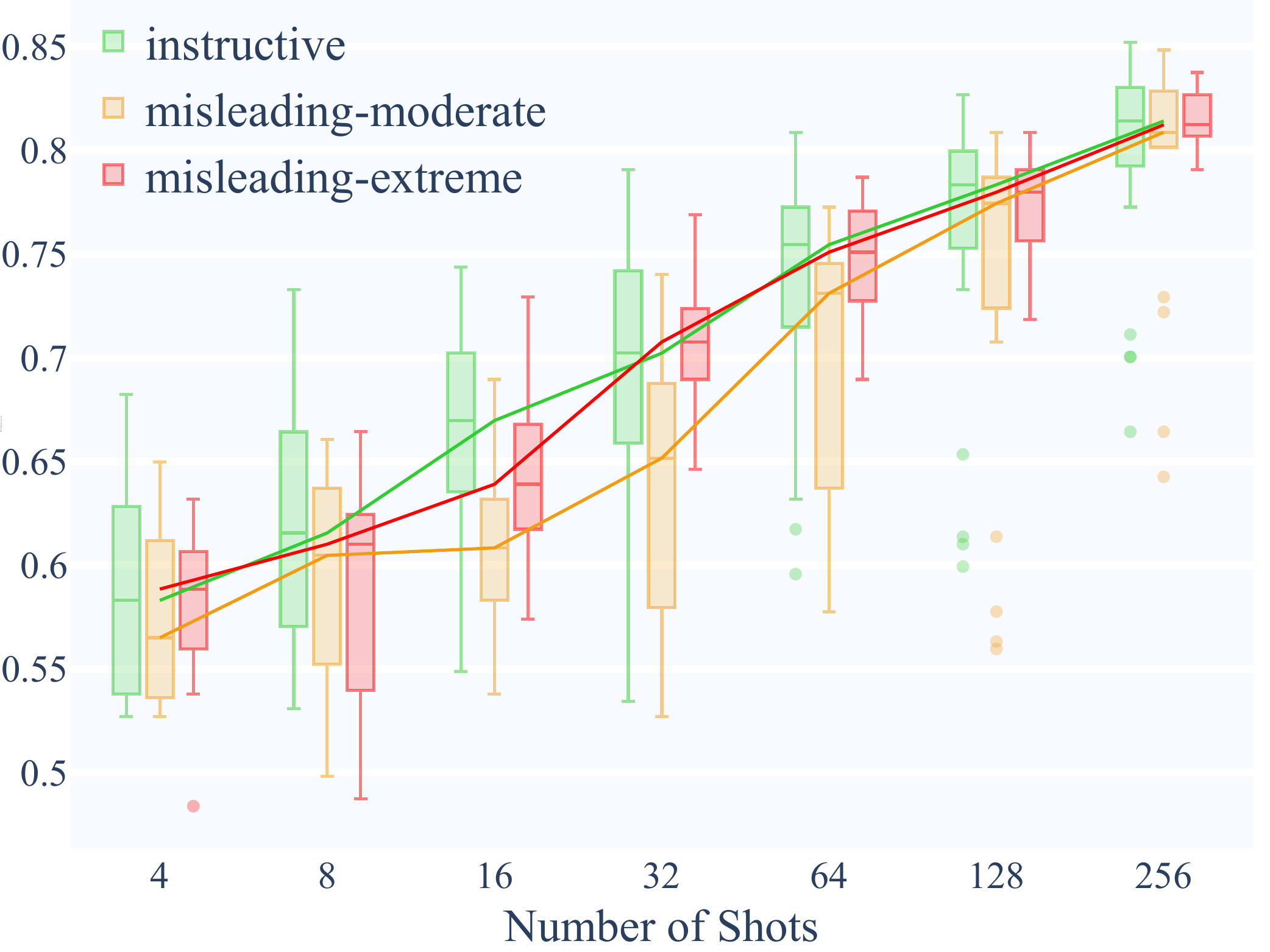}
    \caption{ALBERT on RTE. There is no statistical significance between misleading-extreme and instructive at any number of shots. In contrast, models trained with misleading-moderate templates are significantly worse than the instructive ones from 16 to 64 shots.}
    \label{fig:misleading-A2}
    \vspace{-2ex}
\end{figure}

\newpage
\begin{figure}[t!]
    \vspace{-95ex}
    \includegraphics[width=\linewidth]{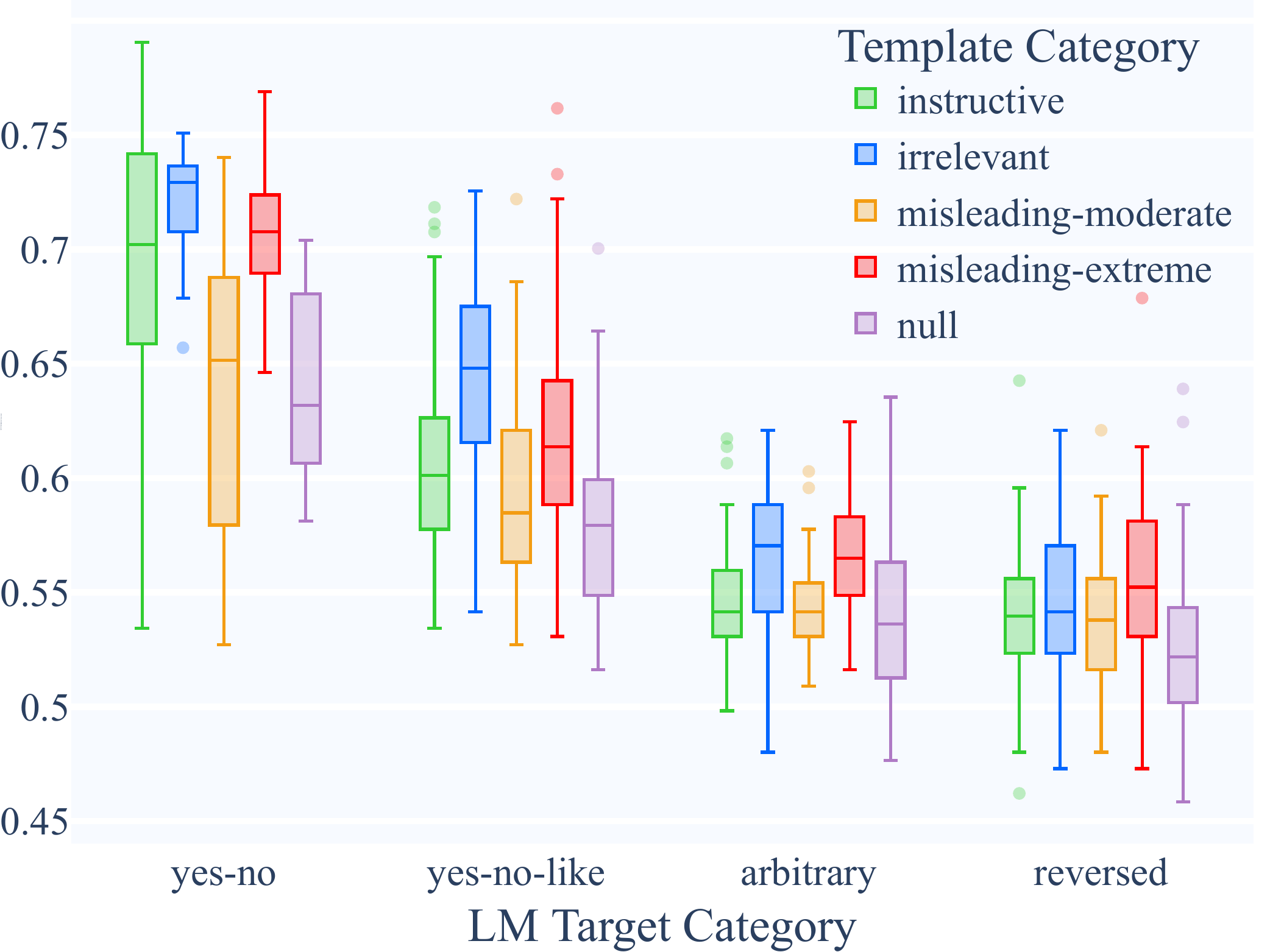}    
    \caption{Median accuracies of all template-target combinations at 32 shots. In general, the choice of target words (x-axis groups) matters much more than the choice of templates (colors).} % which is counterintuitive because humans would care much more about the templates (the task instructions) than the targets (which words they need to respond with)
    \label{fig:targets-A2}
\end{figure}

\renewcommand{\arraystretch}{1.2}

\begin{table*}[]
\section{All Prompts} \label{sec:all-templates}
\subsection{Main Experiment Templates} 
\resizebox{\textwidth}{!}{%
\begin{tabular}{lll} 
\toprule
category & template & adapted from \\ 
\midrule
instructive & \makecell[l]{\prem Using only the above description and what you know about the world, \\ "\hypo" is definitely correct. Yes or no?} & \citet[p. 3]{williams-etal-2018-broad} \\ 
instructive & \prem$\backslash$nquestion: \hypo Yes or no?$\backslash$nanswer: & \citet[p. 59]{gpt} \\ 
instructive & \prem Are we justified in saying that "\hypo"? \\ 
instructive & Given \prem Should we assume that "\hypo" is true? \\ 
instructive & \prem Based on the previous passage, is it true that "\hypo"? \\ 
instructive & Given \prem Is it guaranteed true that "\hypo"? \\ 
instructive & Suppose \prem Can we infer that "\hypo"? \\ 
instructive & Given that \prem Does it follow that "\hypo"? \\ 
instructive & \prem Question: Does this imply that "\hypo"? \\ 
instructive & Given that \prem Therefore, it must be true that "\hypo"? \\ 
%instructive sans qmarks & \makecell[l]{\prem Using only the above description and what you know about the world, \hypo is definitely correct. Yes or no} \\ 
%instructive sans qmarks & \prem$\backslash$nquestion: \hypo Yes or no$\backslash$nanswer: \\ 
%instructive sans qmarks & \prem Are we justified in saying that \hypo \\ 
%instructive sans qmarks & Given \prem Should we assume that \hypo is true \\ 
%instructive sans qmarks & \prem Based on the previous passage, is it true that \hypo \\ 
%instructive sans qmarks & Given \prem Is it guaranteed true that \hypo \\ 
%instructive sans qmarks & Suppose \prem Can we infer that \hypo \\ 
%instructive sans qmarks & Given that \prem Does it follow that \hypo \\ 
%instructive sans qmarks & \prem Question: Does this imply that \hypo \\ 
%instructive sans qmarks & Given that \prem Therefore, it must be true that \hypo \\ 
 \\ 
misleading-moderate & \prem Do most of the above words appear in the following passage? \hypo \\ 
misleading-moderate & \prem Are there lots of similar words in "\hypo"? \\ 
misleading-moderate & \prem Does that have the same meaning as "\hypo"? \\ 
misleading-moderate & \prem Can that be paraphrased as: "\hypo"? \\ 
misleading-moderate & \prem Can that be summarized as "\hypo"? \\ 
\\
misleading-extreme & \prem Does the paragraph start with "the"? \hypo \\ 
misleading-extreme & \prem Is this grammatically correct? \hypo \\ 
misleading-extreme & \prem Is the sentiment positive? \hypo \\ 
misleading-extreme & \prem Is this a sports news? \hypo \\ 
misleading-extreme & \prem Is this French? \hypo \\ 
 \\ 
%irrelevant sans qmarks & \prem Single-family zoning is bad for American cities. \hypo \\ 
%irrelevant sans qmarks & \makecell[l]{\prem When Bolyai sent Gauss his discovery of non-Euclidean geometry, Gauss replied that he arrived at the same results 30 years ago. \hypo} \\ 
%irrelevant sans qmarks & \prem If bonito flakes boil more than a few seconds, the stock becomes too strong. \hypo \\ 
%irrelevant sans qmarks & \makecell[l]{\prem Inflections are annoying and thank god that Middle English got rid of most of them. \hypo} \\ 
%irrelevant sans qmarks & \makecell[l]{\prem Is the pious loved by the gods because it is pious? Or is it pious because it is loved by the gods? \hypo} \\ 
irrelevant & \prem Single-family zoning is bad for American cities. "\hypo"? \vspace{1ex} \\ 
irrelevant & \makecell[l]{\prem Inflections are annoying and thank god that\\ Middle English got rid of most of them. "\hypo"?} \vspace{1ex} \\ 
irrelevant & \makecell[l]{\prem When Bolyai sent Gauss his discovery of non-Euclidean geometry,\\ Gauss replied that he arrived at the same results 30 years ago. "\hypo"?} & \citet[p. 141]{greenberg} \vspace{1ex} \\ 
irrelevant & \makecell[l]{\prem If bonito flakes boil more than a few seconds,\\ the stock becomes too strong? "\hypo"?} & \citet[p. 148]{tsuji} \vspace{1ex} \\ 
irrelevant & \makecell[l]{\prem Is the pious loved by the gods because it is pious?\\ Or is it pious because it is loved by the gods? "\hypo"?} & \citet[10a]{euthyphro}\\ 
\\ 
null & \prem \hypo \\ 
null & \hypo \prem \\ 
null (MLM only) & \prem \{mask\} \hypo \\ 
null (MLM only) & \hypo \{mask\} \prem \\ 
null (MLM only) & \{mask\} \prem \hypo \\ 
null (MLM only) & \{mask\} \hypo \prem \\ 
\bottomrule
\end{tabular}
}
\caption{All prompts used in the main text of the paper. All templates use “yes”/“no” as target words for the entailment and non-entailment classes, respectively. For ternary NLI datasets, we use “unclear” for the neutral class, which performs best after preliminary experiments with other ternary words: “maybe”, “sometimes”, “perhaps”, “possibly”, and “neither”.
Keen readers may notice that some of the instructive templates (e.g., \prompt{should we assume}) do not instruct a strict entailment task. We intentionally wrote a mixture of instructions that asks for strictly logical entailment and pragmatic inference, intending to measure if models can distinguish between the two on datasets such as HANS \citep{mccoy} that magnify different predictions caused by pragmatic effects. Of course, this research question became moot as we found that models cannot even distinguish among much more pathological prompts.}  %(see \rf{tab:verbalizers}).}
%\caption{30-shot accuracy of all templates used in the main paper. To avoid cherry picking, all of them were written prior to analysis and no retroactive exclusion or inclusion of templates from the categories defined above is allowed. We intentionally aimed to write more minimal pairs, while more diverse prompts will be reported in a future version of this paper. For experiments in \rf{sec:verbalizer}, models are trained by permutating these templates with all sets of LM targets (see Appendix \ref{sec:ceteris-paribus}) in addition to their default targets.}
\label{tab:all-templates}
\end{table*}

\begin{table*}[]
\subsection{Ablation Experiment Templates}
\centering
\resizebox{\textwidth}{!}{%
\begin{tabular}{lll} 
\toprule
category & template \\ 
\midrule
instructive sans qmarks & \makecell[l]{\prem Using only the above description and what you know about the world, \hypo is definitely correct. Yes or no} \\ 
instructive sans qmarks & \prem$\backslash$nquestion: \hypo Yes or no$\backslash$nanswer: \\ 
instructive sans qmarks & \prem Are we justified in saying that \hypo \\ 
instructive sans qmarks & Given \prem Should we assume that \hypo is true \\ 
instructive sans qmarks & \prem Based on the previous passage, is it true that \hypo \\ 
instructive sans qmarks & Given \prem Is it guaranteed true that \hypo \\ 
instructive sans qmarks & Suppose \prem Can we infer that \hypo \\ 
instructive sans qmarks & Given that \prem Does it follow that \hypo \\ 
instructive sans qmarks & \prem Question: Does this imply that \hypo \\ 
instructive sans qmarks & Given that \prem Therefore, it must be true that \hypo \\ 
 \\ 
irrelevant sans qmarks & \prem Single-family zoning is bad for American cities. \hypo \\ 
irrelevant sans qmarks & \makecell[l]{\prem Inflections are annoying and thank god that Middle English got rid of most of them. \hypo} \\ 
irrelevant sans qmarks & \makecell[l]{\prem When Bolyai sent Gauss his discovery of non-Euclidean geometry,\\ Gauss replied that he arrived at the same results 30 years ago. \hypo} \\ 
irrelevant sans qmarks & \makecell[l]{\prem If bonito flakes boil more than a few seconds, the stock becomes too strong. \hypo} \\ 
irrelevant sans qmarks & \makecell[l]{\prem Is the pious loved by the gods because it is pious. Or is it pious because it is loved by the gods. \hypo} \\
%\\ 
%null & \prem \hypo \\ 
\bottomrule
\end{tabular}
}
\caption{Used in the study of the effect of question and quotation marks in \rf{sec:qmarks}.}
\label{tab:all-templates}
\end{table*}

\renewcommand{\arraystretch}{1}

\begin{table*}[h]
    \subsection{All Target Words} \label{sec:all-target-words} 
    \vspace{1ex}
    \centering
    \resizebox{11em}{!}{%
    \begin{tabular}{@{}ll@{}}
    \toprule
    Category & Target Words       \\ \midrule
    yes-no & yes;no  \\
    \\
    yes-no-like & true;false  \\
    yes-no-like & positive;negative \\
    yes-no-like & right;wrong \\
    yes-no-like & correct;incorrect \\
    yes-no-like & agree;disagree   \\
    yes-no-like & good;bad  \\
    \\
    reversed & no;yes  \\
    reversed & false;true  \\
    reversed & negative;positive  \\
    \\
    arbitrary & B;C  \\
    arbitrary & cat;dog   \\
    arbitrary & she;he   \\ \bottomrule
    \end{tabular}
    }
    \caption{LM targets used in \rf{sec:verbalizer}. Again, for ternary NLI datasets, we use “unclear” for the neutral class, which performs best after preliminary experiments with other ternary words: “maybe”, “sometimes”, “perhaps”, “possibly”, and “neither”. Within the arbitrary category, in addition to the common anglophone first names as \citet{le-scao-rush-2021-many} use, we also tried word pairs with high semantic similarity (“cat”/“dog”), low similarity (“cake”/“piano”, “write”/“sleep”), and pairs which are highly frequent in the English language (“she”/“he”, “the”/“a”) in preliminary experiments, but we find no consistent difference among these various subcategories of the arbitrary category. } % note has to be single-token
\end{table*}

\clearpage
\begin{figure*}
    \vspace{-4ex}
    \section{Aggregated Results} \label{sec:all-RTE} 
    \subsection{ALBERT on RTE}
    \centering
    \includegraphics[width=.95\linewidth]{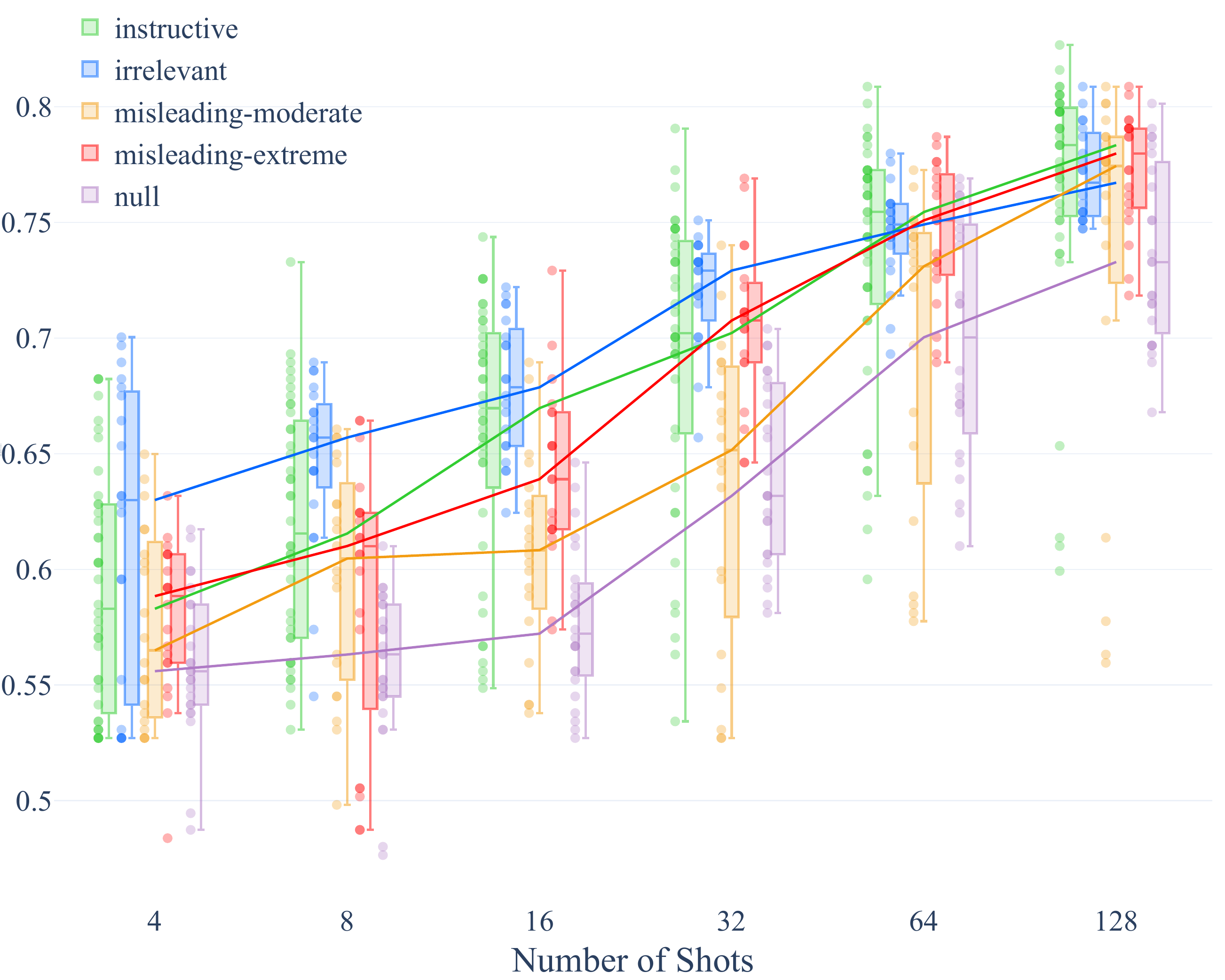}    
%    \caption{in aggregate of all template categories}
    \vspace{-4ex}
\end{figure*}

\begin{table*}[h]
\centering
\resizebox{0.55\textwidth}{!}{%
\begin{tabular}{rlrrrr}
\toprule
 num. shots &   template category &  median &  q3 - q1 &   mean &  std. dev. \\
\midrule
          4 &         instructive &  0.5830 &   0.0885 & 0.5907 &     0.0517 \\
          4 &          irrelevant &  0.6300 &   0.1291 & 0.6170 &     0.0645 \\
          4 &  misleading-extreme &  0.5884 &   0.0469 & 0.5787 &     0.0342 \\
          4 & misleading-moderate &  0.5650 &   0.0722 & 0.5753 &     0.0418 \\
          4 &                null &  0.5560 &   0.0433 & 0.5599 &     0.0324 \\
          8 &         instructive &  0.6155 &   0.0920 & 0.6186 &     0.0524 \\
          8 &          irrelevant &  0.6570 &   0.0307 & 0.6471 &     0.0374 \\
          8 &  misleading-extreme &  0.6101 &   0.0677 & 0.5899 &     0.0595 \\
          8 & misleading-moderate &  0.6047 &   0.0767 & 0.5969 &     0.0490 \\
          8 &                null &  0.5632 &   0.0397 & 0.5586 &     0.0326 \\
         16 &         instructive &  0.6697 &   0.0605 & 0.6594 &     0.0558 \\
         16 &          irrelevant &  0.6787 &   0.0488 & 0.6787 &     0.0294 \\
         16 &  misleading-extreme &  0.6390 &   0.0506 & 0.6413 &     0.0384 \\
         16 & misleading-moderate &  0.6083 &   0.0443 & 0.6072 &     0.0427 \\
         16 &                null &  0.5722 &   0.0379 & 0.5767 &     0.0327 \\
         32 &         instructive &  0.7022 &   0.0813 & 0.6929 &     0.0638 \\
         32 &          irrelevant &  0.7292 &   0.0235 & 0.7206 &     0.0236 \\
         32 &  misleading-extreme &  0.7076 &   0.0334 & 0.7056 &     0.0340 \\
         32 & misleading-moderate &  0.6516 &   0.0992 & 0.6350 &     0.0666 \\
         32 &                null &  0.6318 &   0.0731 & 0.6414 &     0.0392 \\
         64 &         instructive &  0.7545 &   0.0542 & 0.7353 &     0.0548 \\
         64 &          irrelevant &  0.7491 &   0.0198 & 0.7455 &     0.0218 \\
         64 &  misleading-extreme &  0.7509 &   0.0416 & 0.7451 &     0.0299 \\
         64 & misleading-moderate &  0.7310 &   0.0993 & 0.6953 &     0.0688 \\
         64 &                null &  0.7004 &   0.0848 & 0.6998 &     0.0516 \\
        128 &         instructive &  0.7834 &   0.0451 & 0.7661 &     0.0551 \\
        128 &          irrelevant &  0.7671 &   0.0343 & 0.7704 &     0.0200 \\
        128 &  misleading-extreme &  0.7798 &   0.0334 & 0.7729 &     0.0255 \\
        128 & misleading-moderate &  0.7744 &   0.0550 & 0.7354 &     0.0842 \\
        128 &                null &  0.7329 &   0.0695 & 0.7369 &     0.0389 \\
\bottomrule
\end{tabular}%
}
\end{table*}

\clearpage
\begin{figure*}
    \vspace{-4ex}
    \subsection{ALBERT on ANLI R1} \label{sec:all-ANLI}
    \centering
    \includegraphics[width=.95\linewidth]{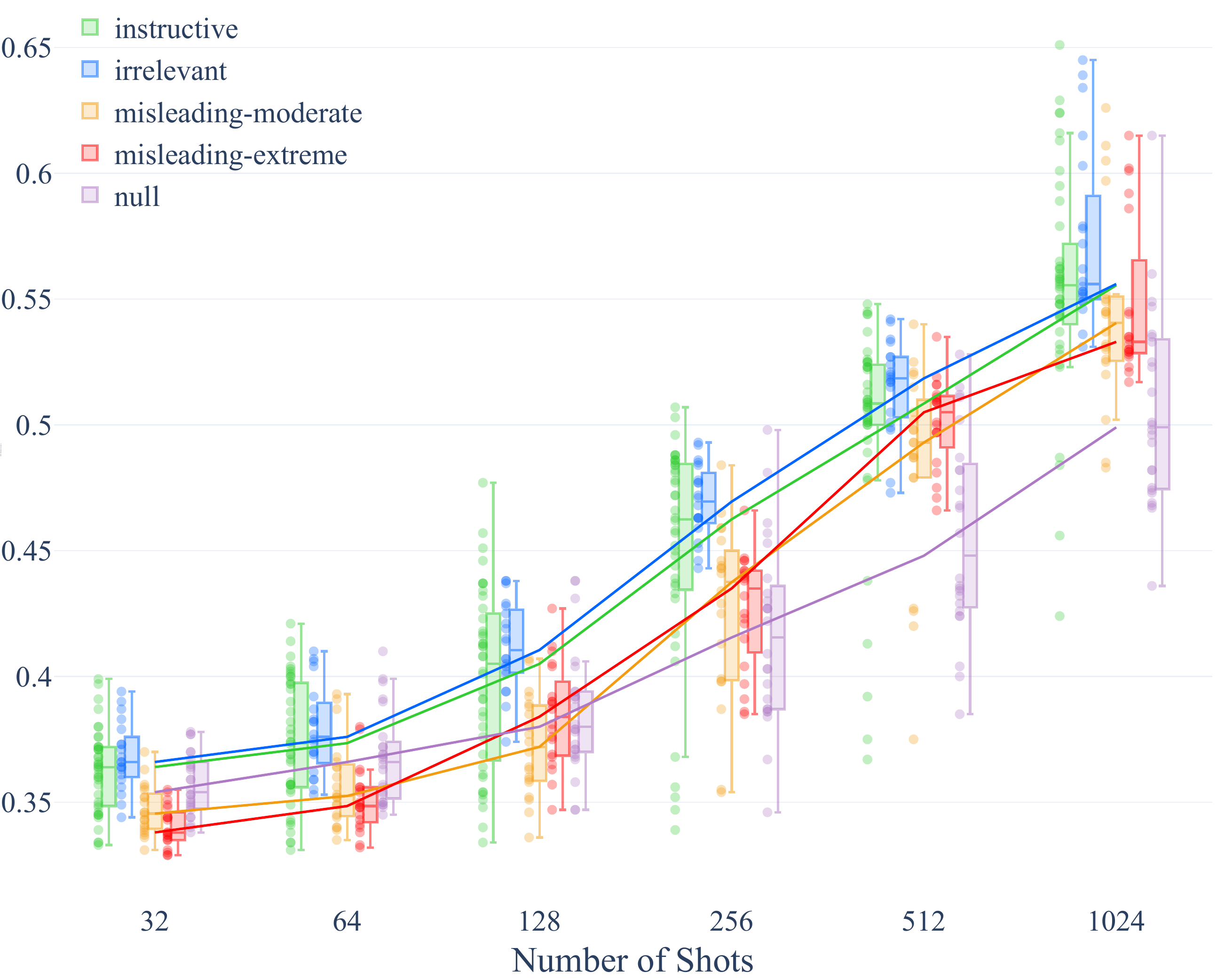}    
%    \caption{in aggregate of all template categories}
    \vspace{-4ex}
\end{figure*}
\begin{table*}[h]
\centering
\resizebox{0.55\textwidth}{!}{%
\begin{tabular}{rlrrrr}
\toprule
 num. shots &   template category &  median &  q3 - q1 &   mean &  std. dev. \\
\midrule
         32 &         instructive &  0.3640 &   0.0232 & 0.3625 &     0.0166 \\
         32 &          irrelevant &  0.3660 &   0.0140 & 0.3681 &     0.0134 \\
         32 &  misleading-extreme &  0.3380 &   0.0100 & 0.3404 &     0.0081 \\
         32 & misleading-moderate &  0.3455 &   0.0130 & 0.3470 &     0.0098 \\
         32 &                null &  0.3540 &   0.0177 & 0.3567 &     0.0122 \\
         64 &         instructive &  0.3735 &   0.0408 & 0.3738 &     0.0251 \\
         64 &          irrelevant &  0.3760 &   0.0210 & 0.3788 &     0.0178 \\
         64 &  misleading-extreme &  0.3485 &   0.0135 & 0.3510 &     0.0129 \\
         64 & misleading-moderate &  0.3525 &   0.0197 & 0.3574 &     0.0171 \\
         64 &                null &  0.3660 &   0.0208 & 0.3675 &     0.0184 \\
        128 &         instructive &  0.4050 &   0.0562 & 0.3992 &     0.0356 \\
        128 &          irrelevant &  0.4105 &   0.0240 & 0.4120 &     0.0176 \\
        128 &  misleading-extreme &  0.3840 &   0.0262 & 0.3843 &     0.0204 \\
        128 & misleading-moderate &  0.3720 &   0.0295 & 0.3725 &     0.0199 \\
        128 &                null &  0.3800 &   0.0235 & 0.3857 &     0.0247 \\
        256 &         instructive &  0.4625 &   0.0490 & 0.4504 &     0.0450 \\
        256 &          irrelevant &  0.4695 &   0.0175 & 0.4694 &     0.0147 \\
        256 &  misleading-extreme &  0.4350 &   0.0297 & 0.4263 &     0.0231 \\
        256 & misleading-moderate &  0.4375 &   0.0492 & 0.4265 &     0.0353 \\
        256 &                null &  0.4155 &   0.0475 & 0.4167 &     0.0365 \\
        512 &         instructive &  0.5085 &   0.0235 & 0.4992 &     0.0434 \\
        512 &          irrelevant &  0.5185 &   0.0230 & 0.5154 &     0.0186 \\
        512 &  misleading-extreme &  0.5050 &   0.0172 & 0.5008 &     0.0177 \\
        512 & misleading-moderate &  0.4930 &   0.0285 & 0.4839 &     0.0413 \\
        512 &                null &  0.4480 &   0.0550 & 0.4564 &     0.0399 \\
       1024 &         instructive &  0.5555 &   0.0270 & 0.5557 &     0.0449 \\
       1024 &          irrelevant &  0.5560 &   0.0345 & 0.5729 &     0.0351 \\
       1024 &  misleading-extreme &  0.5330 &   0.0265 & 0.5477 &     0.0316 \\
       1024 & misleading-moderate &  0.5405 &   0.0247 & 0.5447 &     0.0388 \\
       1024 &                null &  0.4990 &   0.0588 & 0.5062 &     0.0392 \\
\bottomrule
\end{tabular}%
}
\end{table*}

\clearpage
\begin{figure*}
    \vspace{-4ex}
    \subsection{T5 770M on RTE}
    \centering
    \includegraphics[width=.95\linewidth]{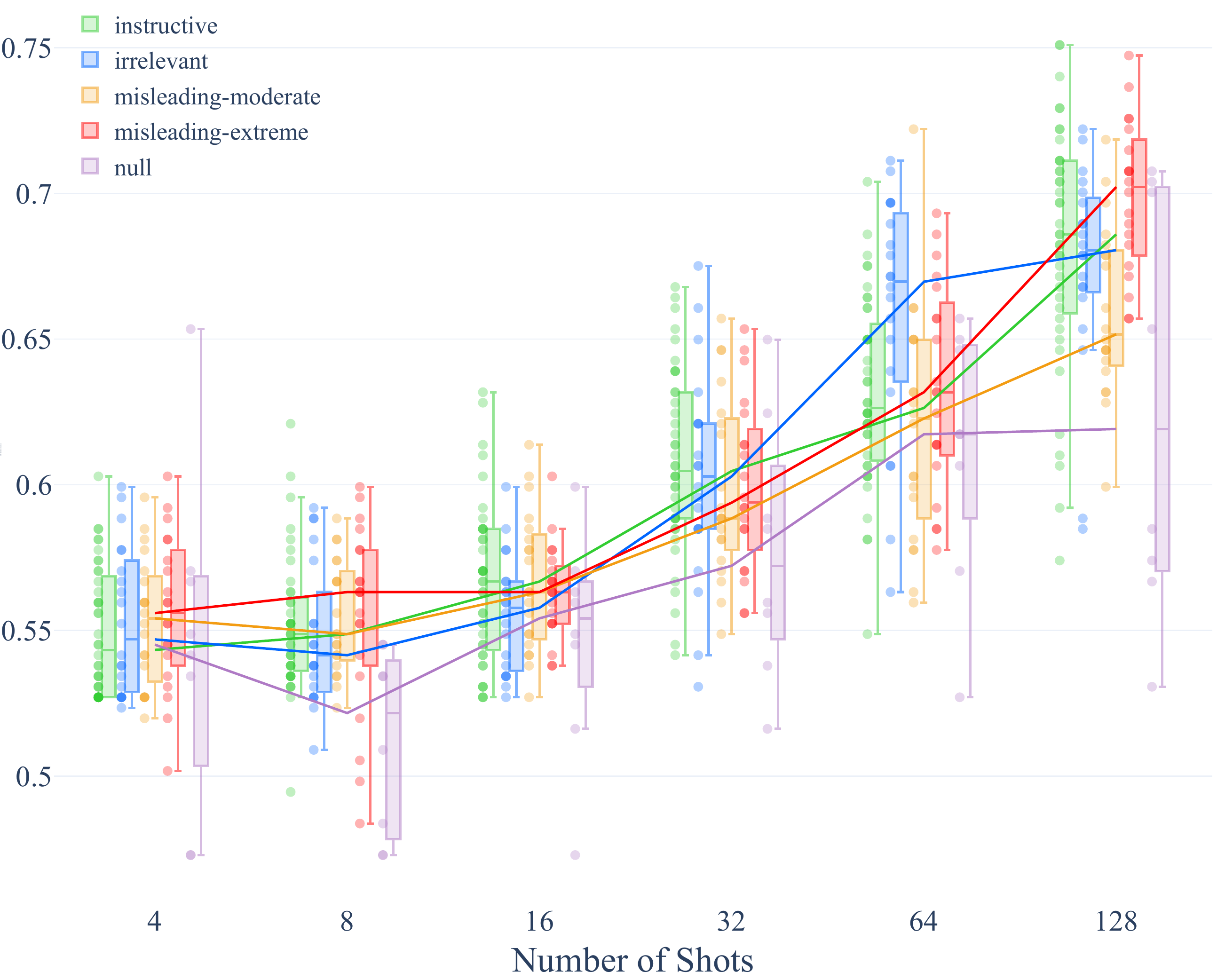}    
%    \caption{in aggregate of all template categories}
%    \vspace{-4ex}
\end{figure*}

\begin{table*}[h]
\centering
\resizebox{0.55\textwidth}{!}{%
\begin{tabular}{rlrrrr}
\toprule
 num. shots &   template category &  median &  q3 - q1 &   mean &  std. dev. \\
\midrule
          4 &         instructive &  0.5433 &   0.0406 & 0.5493 &     0.0219 \\
          4 &          irrelevant &  0.5469 &   0.0424 & 0.5532 &     0.0252 \\
          4 &  misleading-extreme &  0.5560 &   0.0361 & 0.5561 &     0.0263 \\
          4 & misleading-moderate &  0.5542 &   0.0325 & 0.5531 &     0.0220 \\
          4 &                null &  0.5451 &   0.0487 & 0.5451 &     0.0578 \\
          8 &         instructive &  0.5487 &   0.0235 & 0.5516 &     0.0232 \\
          8 &          irrelevant &  0.5415 &   0.0280 & 0.5480 &     0.0244 \\
          8 &  misleading-extreme &  0.5632 &   0.0379 & 0.5545 &     0.0322 \\
          8 & misleading-moderate &  0.5487 &   0.0280 & 0.5543 &     0.0192 \\
          8 &                null &  0.5217 &   0.0560 & 0.5122 &     0.0317 \\
         16 &         instructive &  0.5668 &   0.0406 & 0.5662 &     0.0277 \\
         16 &          irrelevant &  0.5578 &   0.0298 & 0.5558 &     0.0199 \\
         16 &  misleading-extreme &  0.5632 &   0.0190 & 0.5634 &     0.0160 \\
         16 & misleading-moderate &  0.5632 &   0.0343 & 0.5666 &     0.0239 \\
         16 &                null &  0.5542 &   0.0271 & 0.5469 &     0.0381 \\
         32 &         instructive &  0.6047 &   0.0433 & 0.6078 &     0.0317 \\
         32 &          irrelevant &  0.6029 &   0.0361 & 0.6025 &     0.0366 \\
         32 &  misleading-extreme &  0.5939 &   0.0352 & 0.5996 &     0.0292 \\
         32 & misleading-moderate &  0.5884 &   0.0424 & 0.5986 &     0.0311 \\
         32 &                null &  0.5722 &   0.0460 & 0.5772 &     0.0443 \\
         64 &         instructive &  0.6264 &   0.0433 & 0.6318 &     0.0324 \\
         64 &          irrelevant &  0.6697 &   0.0542 & 0.6585 &     0.0421 \\
         64 &  misleading-extreme &  0.6318 &   0.0478 & 0.6336 &     0.0355 \\
         64 & misleading-moderate &  0.6227 &   0.0578 & 0.6195 &     0.0400 \\
         64 &                null &  0.6173 &   0.0496 & 0.6115 &     0.0442 \\
        128 &         instructive &  0.6859 &   0.0514 & 0.6820 &     0.0421 \\
        128 &          irrelevant &  0.6805 &   0.0307 & 0.6749 &     0.0362 \\
        128 &  misleading-extreme &  0.7022 &   0.0361 & 0.6987 &     0.0260 \\
        128 & misleading-moderate &  0.6516 &   0.0379 & 0.6597 &     0.0295 \\
        128 &                null &  0.6191 &   0.1291 & 0.6277 &     0.0717 \\
\bottomrule
\end{tabular}%
}
\end{table*}

\clearpage
\begin{figure*}
    \vspace{-4ex}
    \subsection{T5 3B on RTE} \label{sec:t5-3B-rte}
    \centering
    \includegraphics[width=.95\linewidth]{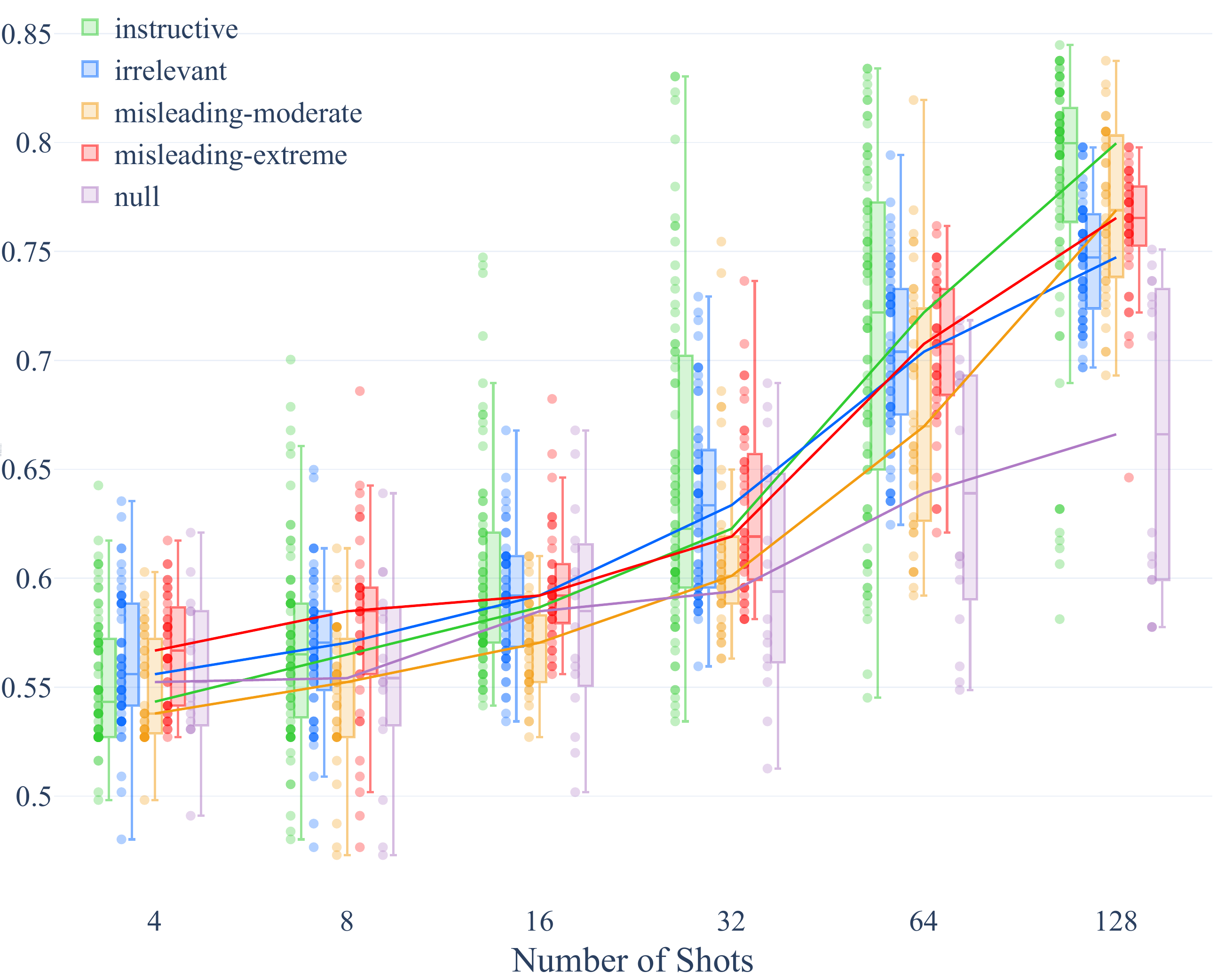}    
%    \vspace{-4ex}
\end{figure*}

\begin{table*}[h]
\centering
\resizebox{0.55\textwidth}{!}{%
\begin{tabular}{rlrrrr}
\toprule
 num. shots &   template category &  median &  q3 - q1 &   mean &  std. dev. \\
\midrule
          4 &         instructive &  0.5433 &   0.0442 & 0.5524 &     0.0297 \\
          4 &          irrelevant &  0.5560 &   0.0469 & 0.5611 &     0.0308 \\
          4 &  misleading-extreme &  0.5668 &   0.0442 & 0.5671 &     0.0251 \\
          4 & misleading-moderate &  0.5379 &   0.0415 & 0.5497 &     0.0247 \\
          4 &                null &  0.5523 &   0.0514 & 0.5575 &     0.0334 \\
          8 &         instructive &  0.5650 &   0.0514 & 0.5680 &     0.0427 \\
          8 &          irrelevant &  0.5704 &   0.0343 & 0.5676 &     0.0332 \\
          8 &  misleading-extreme &  0.5848 &   0.0397 & 0.5773 &     0.0431 \\
          8 & misleading-moderate &  0.5523 &   0.0442 & 0.5485 &     0.0309 \\
          8 &                null &  0.5542 &   0.0523 & 0.5553 &     0.0459 \\
         16 &         instructive &  0.5866 &   0.0505 & 0.6005 &     0.0467 \\
         16 &          irrelevant &  0.5921 &   0.0406 & 0.5907 &     0.0279 \\
         16 &  misleading-extreme &  0.5921 &   0.0262 & 0.5953 &     0.0271 \\
         16 & misleading-moderate &  0.5704 &   0.0298 & 0.5693 &     0.0212 \\
         16 &                null &  0.5848 &   0.0614 & 0.5833 &     0.0481 \\
         32 &         instructive &  0.6227 &   0.1056 & 0.6463 &     0.0757 \\
         32 &          irrelevant &  0.6336 &   0.0623 & 0.6349 &     0.0416 \\
         32 &  misleading-extreme &  0.6191 &   0.0542 & 0.6315 &     0.0393 \\
         32 & misleading-moderate &  0.6011 &   0.0298 & 0.6134 &     0.0440 \\
         32 &                null &  0.5939 &   0.0848 & 0.6031 &     0.0548 \\
         64 &         instructive &  0.7220 &   0.1227 & 0.7113 &     0.0784 \\
         64 &          irrelevant &  0.7040 &   0.0578 & 0.7032 &     0.0408 \\
         64 &  misleading-extreme &  0.7076 &   0.0478 & 0.7039 &     0.0352 \\
         64 & misleading-moderate &  0.6697 &   0.0957 & 0.6792 &     0.0569 \\
         64 &                null &  0.6390 &   0.0984 & 0.6397 &     0.0618 \\
        128 &         instructive &  0.7996 &   0.0496 & 0.7769 &     0.0627 \\
        128 &          irrelevant &  0.7473 &   0.0415 & 0.7468 &     0.0271 \\
        128 &  misleading-extreme &  0.7653 &   0.0262 & 0.7604 &     0.0295 \\
        128 & misleading-moderate &  0.7690 &   0.0632 & 0.7685 &     0.0373 \\
        128 &                null &  0.6661 &   0.1318 & 0.6640 &     0.0716 \\
\bottomrule
\end{tabular}%
}
\end{table*}

\clearpage
\begin{figure*}
    \vspace{-4ex}
    \subsection{T0 3B on RTE}
    \centering
    \includegraphics[width=.95\linewidth]{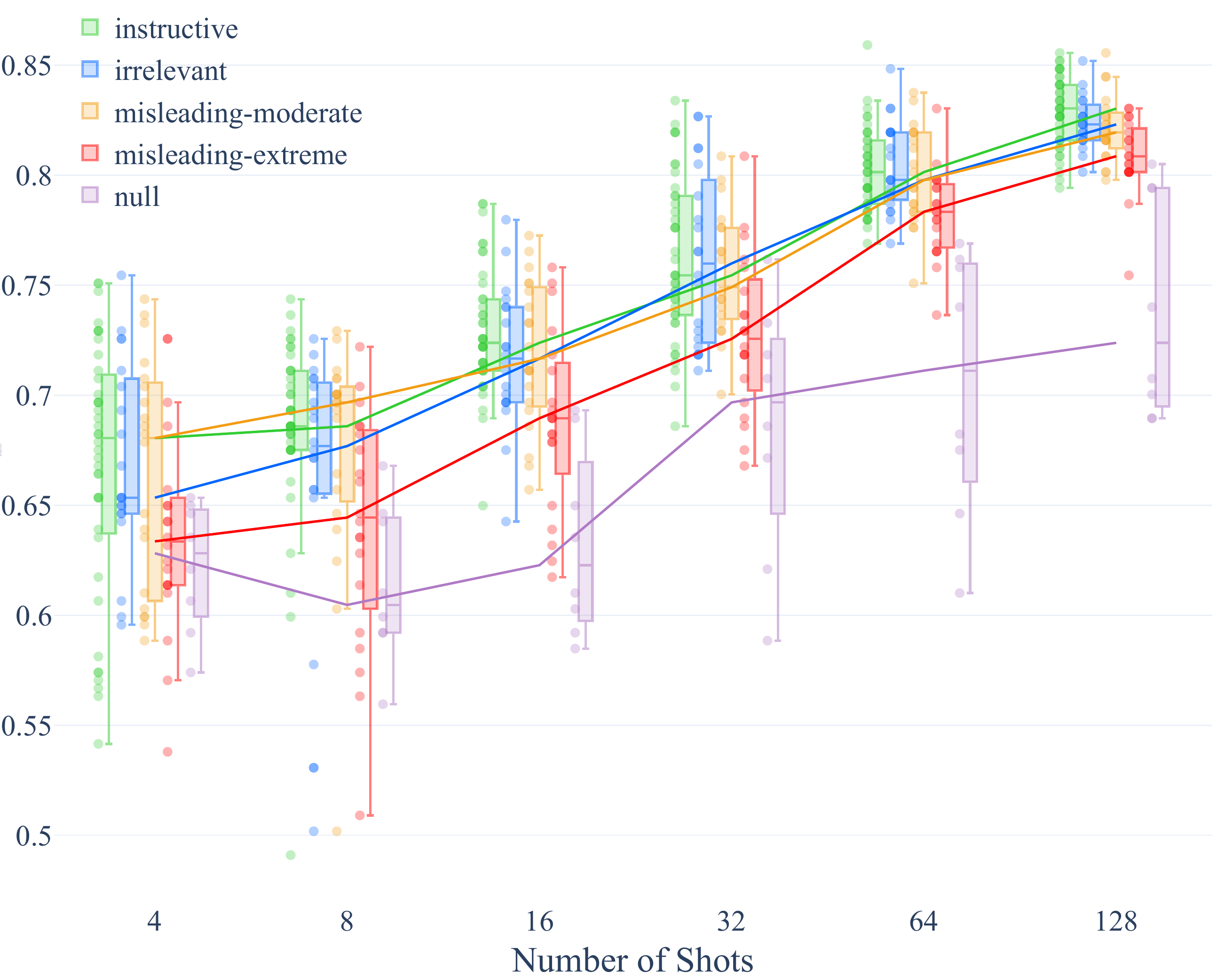}    
%    \vspace{-4ex}
\end{figure*}

\begin{table*}[h]
\centering
\resizebox{0.55\textwidth}{!}{%
\begin{tabular}{rlrrrr}
\toprule
 num. shots &   template category &  median &  q3 - q1 &   mean &  std. dev. \\
\midrule
          4 &         instructive &  0.6805 &   0.0704 & 0.6677 &     0.0580 \\
          4 &          irrelevant &  0.6534 &   0.0596 & 0.6695 &     0.0450 \\
          4 &  misleading-extreme &  0.6336 &   0.0379 & 0.6368 &     0.0469 \\
          4 & misleading-moderate &  0.6805 &   0.0966 & 0.6644 &     0.0525 \\
          4 &                null &  0.6282 &   0.0442 & 0.6223 &     0.0292 \\
          8 &         instructive &  0.6859 &   0.0361 & 0.6850 &     0.0438 \\
          8 &          irrelevant &  0.6769 &   0.0487 & 0.6579 &     0.0674 \\
          8 &  misleading-extreme &  0.6444 &   0.0749 & 0.6401 &     0.0543 \\
          8 & misleading-moderate &  0.6968 &   0.0478 & 0.6747 &     0.0530 \\
          8 &                null &  0.6047 &   0.0514 & 0.6137 &     0.0357 \\
         16 &         instructive &  0.7238 &   0.0325 & 0.7290 &     0.0284 \\
         16 &          irrelevant &  0.7166 &   0.0433 & 0.7171 &     0.0315 \\
         16 &  misleading-extreme &  0.6895 &   0.0415 & 0.6879 &     0.0410 \\
         16 & misleading-moderate &  0.7166 &   0.0523 & 0.7191 &     0.0337 \\
         16 &                null &  0.6227 &   0.0596 & 0.6322 &     0.0423 \\
         32 &         instructive &  0.7545 &   0.0542 & 0.7627 &     0.0369 \\
         32 &          irrelevant &  0.7599 &   0.0695 & 0.7621 &     0.0397 \\
         32 &  misleading-extreme &  0.7256 &   0.0451 & 0.7278 &     0.0361 \\
         32 & misleading-moderate &  0.7491 &   0.0406 & 0.7551 &     0.0279 \\
         32 &                null &  0.6968 &   0.0632 & 0.6859 &     0.0578 \\
         64 &         instructive &  0.8014 &   0.0289 & 0.8027 &     0.0190 \\
         64 &          irrelevant &  0.7978 &   0.0298 & 0.8040 &     0.0204 \\
         64 &  misleading-extreme &  0.7834 &   0.0271 & 0.7827 &     0.0201 \\
         64 & misleading-moderate &  0.7978 &   0.0361 & 0.8000 &     0.0225 \\
         64 &                null &  0.7112 &   0.0912 & 0.7053 &     0.0600 \\
        128 &         instructive &  0.8303 &   0.0253 & 0.8292 &     0.0161 \\
        128 &          irrelevant &  0.8231 &   0.0153 & 0.8244 &     0.0118 \\
        128 &  misleading-extreme &  0.8087 &   0.0190 & 0.8088 &     0.0174 \\
        128 & misleading-moderate &  0.8195 &   0.0135 & 0.8215 &     0.0152 \\
        128 &                null &  0.7238 &   0.0966 & 0.7401 &     0.0505 \\
\bottomrule
\end{tabular}%
}
\end{table*}

\clearpage
\begin{figure*}
    \vspace{-4ex}
    \subsection{T0 3B on ANLI R1}
    \centering
    \includegraphics[width=.95\linewidth]{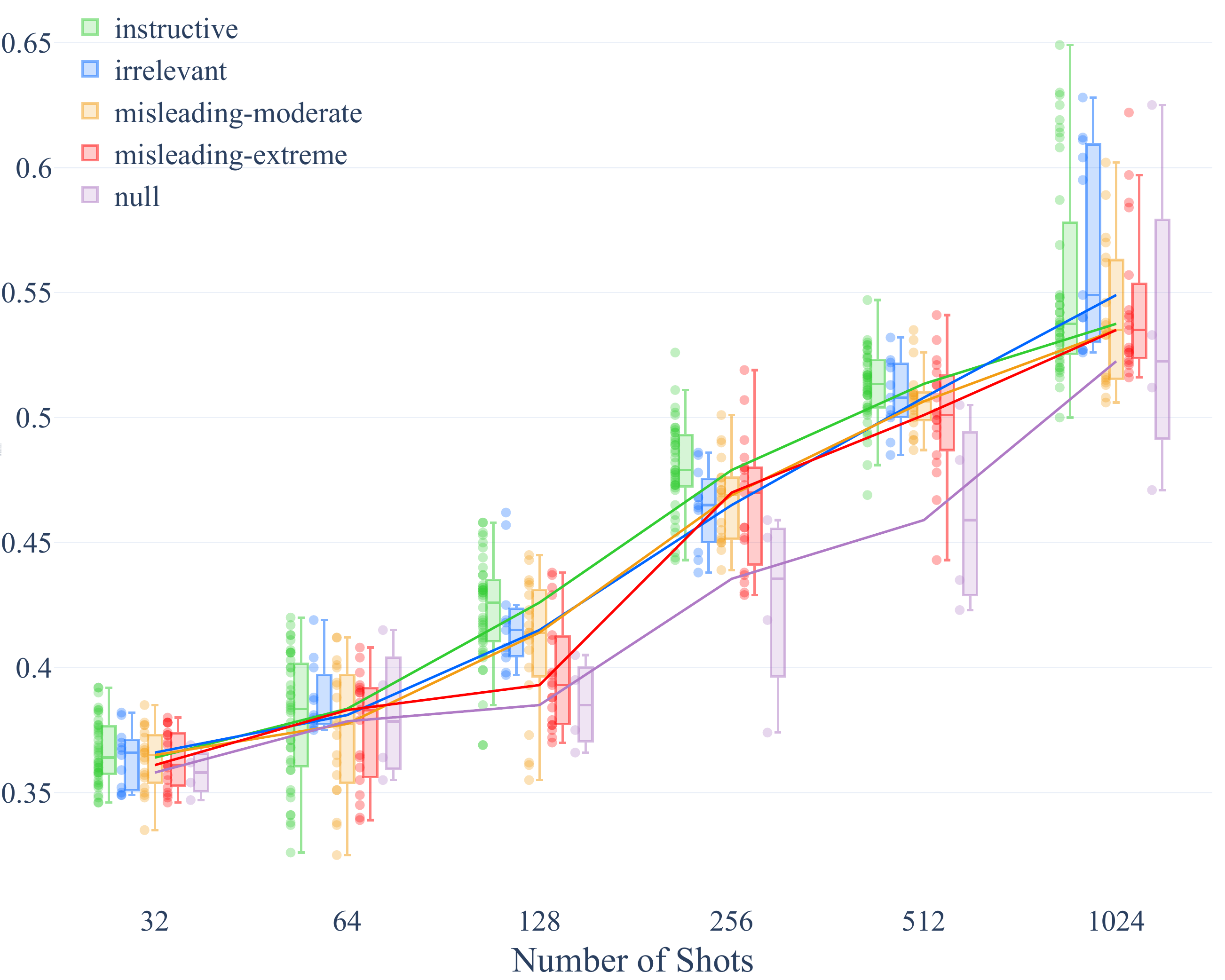}    
%    \vspace{-4ex}
\end{figure*}

\begin{table*}[h]
\centering
\resizebox{0.55\textwidth}{!}{%
\begin{tabular}{rlrrrr}
\toprule
 num. shots &   template category &  median &  q3 - q1 &   mean &  std. dev. \\
\midrule
         32 &         instructive &  0.3640 &   0.0185 & 0.3664 &     0.0129 \\
         32 &          irrelevant &  0.3660 &   0.0190 & 0.3637 &     0.0119 \\
         32 &  misleading-extreme &  0.3610 &   0.0200 & 0.3638 &     0.0117 \\
         32 & misleading-moderate &  0.3650 &   0.0175 & 0.3631 &     0.0125 \\
         32 &                null &  0.3580 &   0.0115 & 0.3580 &     0.0096 \\
         64 &         instructive &  0.3835 &   0.0395 & 0.3797 &     0.0255 \\
         64 &          irrelevant &  0.3810 &   0.0160 & 0.3878 &     0.0141 \\
         64 &  misleading-extreme &  0.3830 &   0.0340 & 0.3753 &     0.0223 \\
         64 & misleading-moderate &  0.3775 &   0.0400 & 0.3749 &     0.0259 \\
         64 &                null &  0.3785 &   0.0368 & 0.3817 &     0.0275 \\
        128 &         instructive &  0.4260 &   0.0233 & 0.4226 &     0.0214 \\
        128 &          irrelevant &  0.4150 &   0.0170 & 0.4190 &     0.0219 \\
        128 &  misleading-extreme &  0.3930 &   0.0340 & 0.3975 &     0.0227 \\
        128 & misleading-moderate &  0.4140 &   0.0318 & 0.4092 &     0.0274 \\
        128 &                null &  0.3850 &   0.0247 & 0.3852 &     0.0179 \\
        256 &         instructive &  0.4790 &   0.0197 & 0.4804 &     0.0181 \\
        256 &          irrelevant &  0.4650 &   0.0185 & 0.4640 &     0.0161 \\
        256 &  misleading-extreme &  0.4700 &   0.0355 & 0.4654 &     0.0259 \\
        256 & misleading-moderate &  0.4690 &   0.0242 & 0.4670 &     0.0167 \\
        256 &                null &  0.4355 &   0.0460 & 0.4260 &     0.0388 \\
        512 &         instructive &  0.5135 &   0.0185 & 0.5123 &     0.0147 \\
        512 &          irrelevant &  0.5080 &   0.0205 & 0.5088 &     0.0147 \\
        512 &  misleading-extreme &  0.5010 &   0.0265 & 0.5007 &     0.0233 \\
        512 & misleading-moderate &  0.5065 &   0.0105 & 0.5066 &     0.0127 \\
        512 &                null &  0.4590 &   0.0565 & 0.4615 &     0.0389 \\
       1024 &         instructive &  0.5375 &   0.0477 & 0.5539 &     0.0406 \\
       1024 &          irrelevant &  0.5490 &   0.0740 & 0.5690 &     0.0406 \\
       1024 &  misleading-extreme &  0.5350 &   0.0255 & 0.5447 &     0.0304 \\
       1024 & misleading-moderate &  0.5350 &   0.0467 & 0.5403 &     0.0279 \\
       1024 &                null &  0.5225 &   0.0543 & 0.5353 &     0.0651 \\
\bottomrule
\end{tabular}%
}
\end{table*}

\begin{table*}[h]
\subsection{T5 11B, T0 11B, and GPT-3 175B (\rf{fig:big-models})} \vspace{1ex}
\centering
\resizebox{0.7\textwidth}{!}{%
\begin{tabular}{llrrrr}
\toprule
       model &   template category &  median &  q3 - q1 &   mean &  std. dev. \\
\midrule
GPT-3 (175B) &         instructive &  0.6534 &   0.0722 & 0.6472 &     0.0429 \\
GPT-3 (175B) &          irrelevant &  0.6101 &   0.0361 & 0.6260 &     0.0326 \\
GPT-3 (175B) &  misleading-extreme &  0.6173 &   0.0072 & 0.6217 &     0.0143 \\
GPT-3 (175B) & misleading-moderate &  0.6498 &   0.0578 & 0.6318 &     0.0480 \\
T5 LMA (11B) &         instructive &  0.6679 &   0.1462 & 0.6797 &     0.0823 \\
T5 LMA (11B) &          irrelevant &  0.6426 &   0.0776 & 0.6368 &     0.0488 \\
T5 LMA (11B) &  misleading-extreme &  0.5993 &   0.0794 & 0.6070 &     0.0619 \\
T5 LMA (11B) & misleading-moderate &  0.5957 &   0.1137 & 0.6072 &     0.0653 \\
T5 LMA (11B) &                null &  0.5560 &   0.0442 & 0.5578 &     0.0332 \\
    T0 (11B) &         instructive &  0.7942 &   0.0623 & 0.7959 &     0.0392 \\
    T0 (11B) &          irrelevant &  0.7906 &   0.0632 & 0.7942 &     0.0384 \\
    T0 (11B) &  misleading-extreme &  0.7401 &   0.0650 & 0.7338 &     0.0496 \\
    T0 (11B) & misleading-moderate &  0.7942 &   0.0397 & 0.7858 &     0.0356 \\
    T0 (11B) &                null &  0.6986 &   0.0695 & 0.6847 &     0.0484 \\
  T0++ (11B) &         instructive &  0.8321 &   0.0316 & 0.8319 &     0.0282 \\
  T0++ (11B) &          irrelevant &  0.8267 &   0.0433 & 0.8207 &     0.0323 \\
  T0++ (11B) &  misleading-extreme &  0.8051 &   0.0614 & 0.8029 &     0.0593 \\
  T0++ (11B) & misleading-moderate &  0.8159 &   0.0487 & 0.8039 &     0.0333 \\
  T0++ (11B) &                null &  0.7509 &   0.0505 & 0.7379 &     0.0362 \\
\bottomrule
\end{tabular}
}
%\caption{Full results of}
\end{table*}

%\section{All ANLI results, in aggregate} \label{sec:all-ANLI}
%\subsection{ALBERT}
%\subsection{T5 750M}
%\subsection{T5 3B}
%\subsection{T0 3B}
%

\clearpage

\begin{figure*}
    \section{Results of Individual Templates} \label{sec:ind-temp}
    \subsection{ALBERT}
    \centering
    \includegraphics[width=\linewidth]{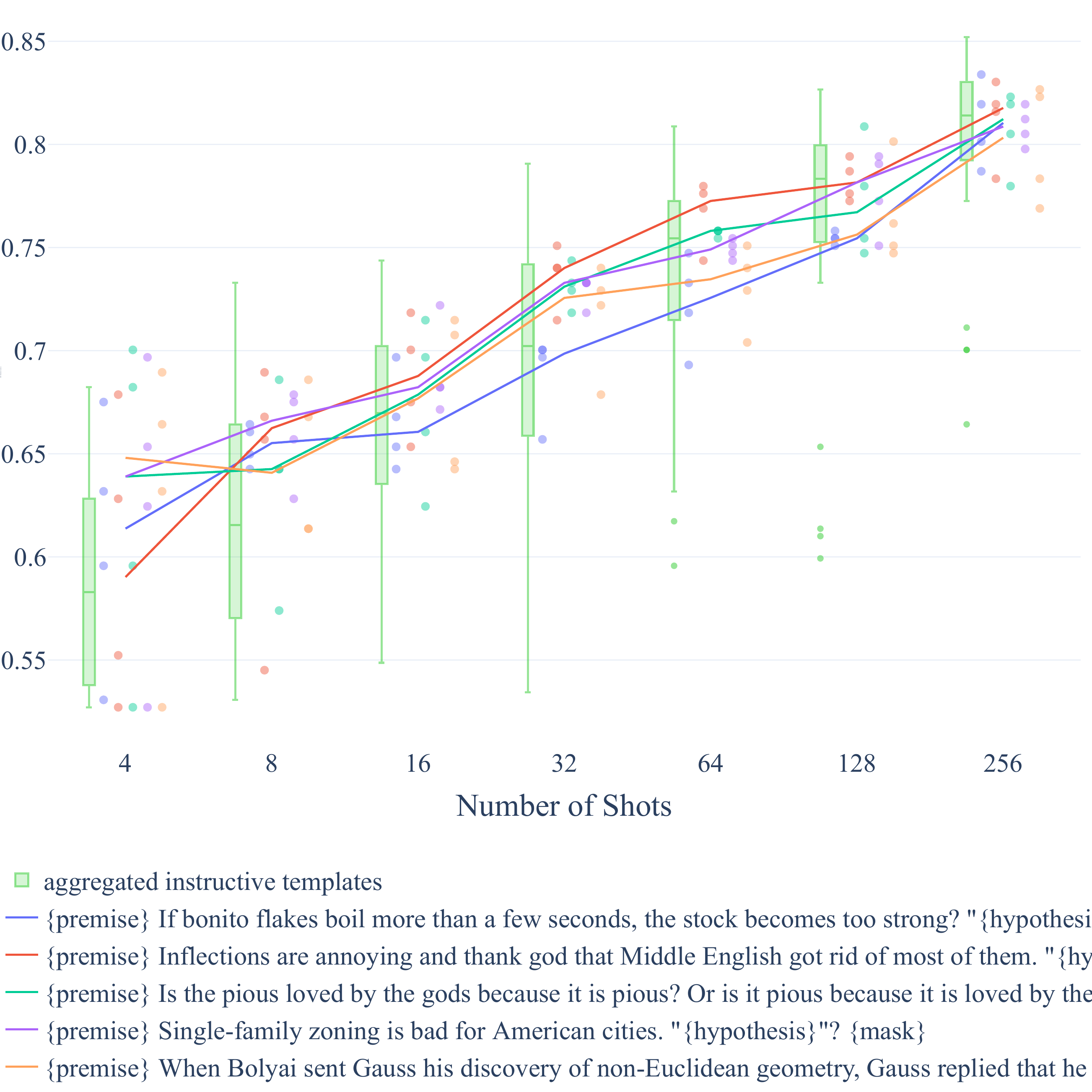}
    \caption{ALBERT with all irrelevant templates and the aggregated instructive for reference.}    
\end{figure*}

\begin{figure*}
    \centering
    \includegraphics[width=\linewidth]{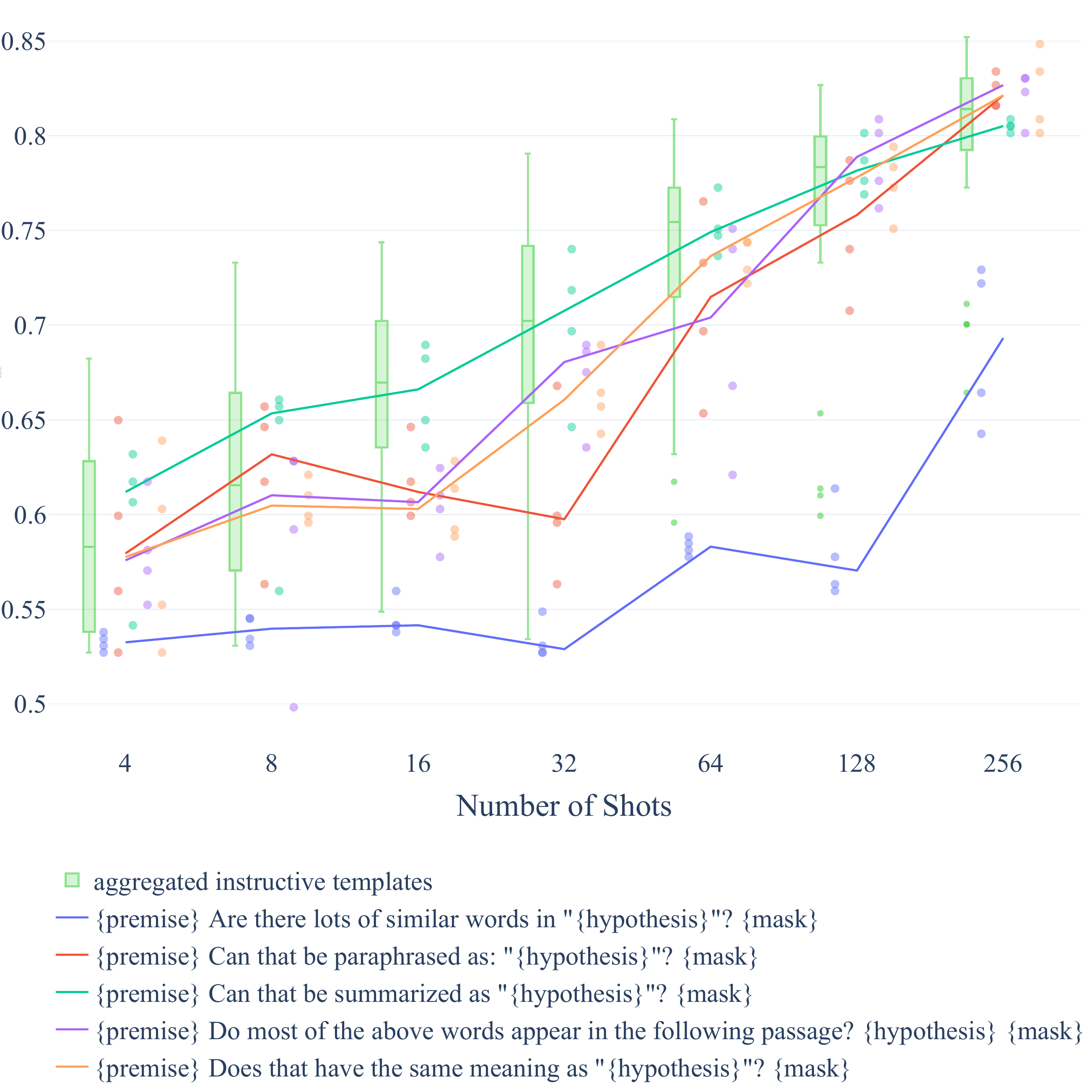}
    \caption{ALBERT with all misleading-moderate templates and the aggregated instructive for reference.}    
\end{figure*}

\begin{figure*}
    \centering
    \includegraphics[width=\linewidth]{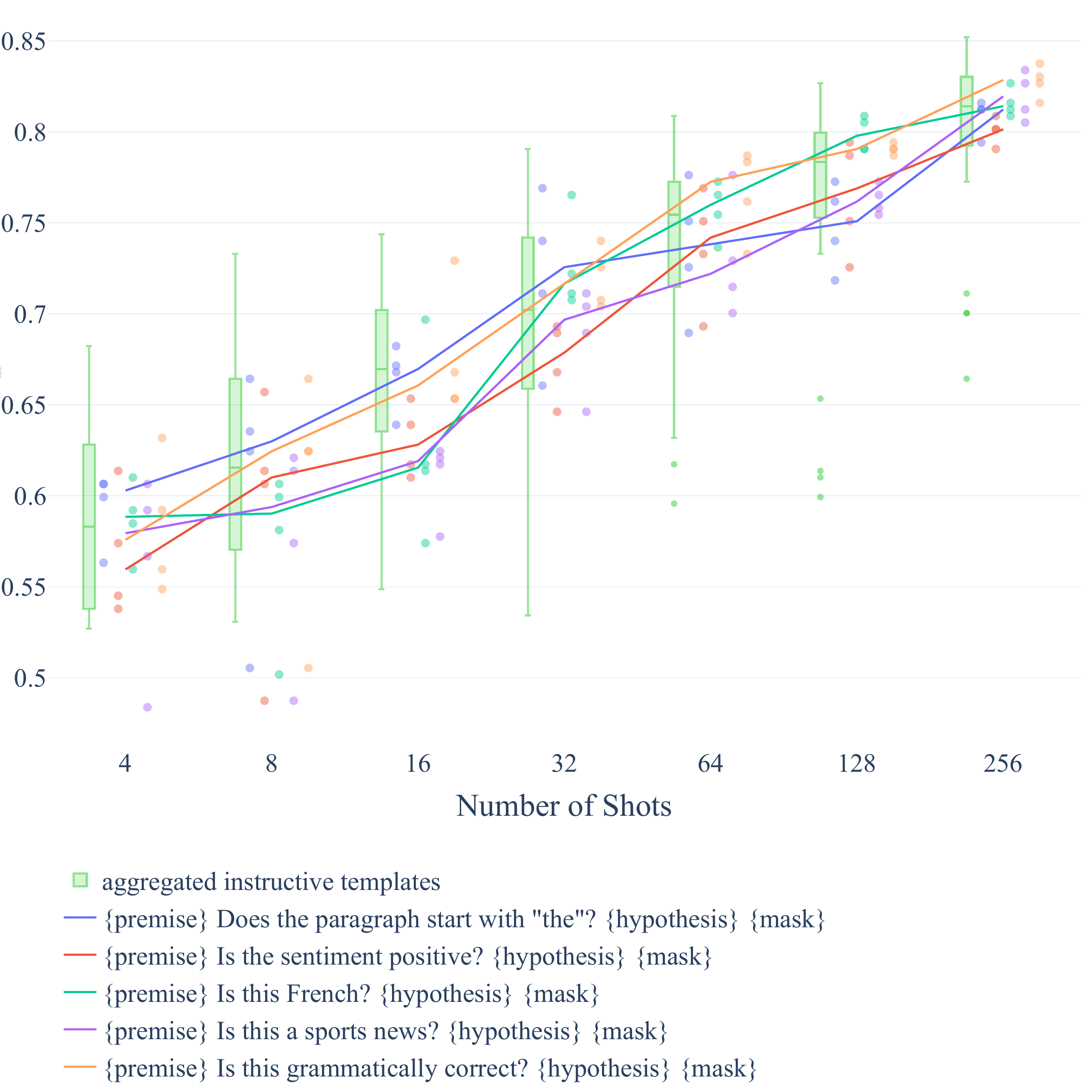}
    \caption{ALBERT with all misleading-extreme templates and the aggregated instructive for reference.}    
\end{figure*}

\begin{figure*}
    \centering
    \includegraphics[width=\linewidth]{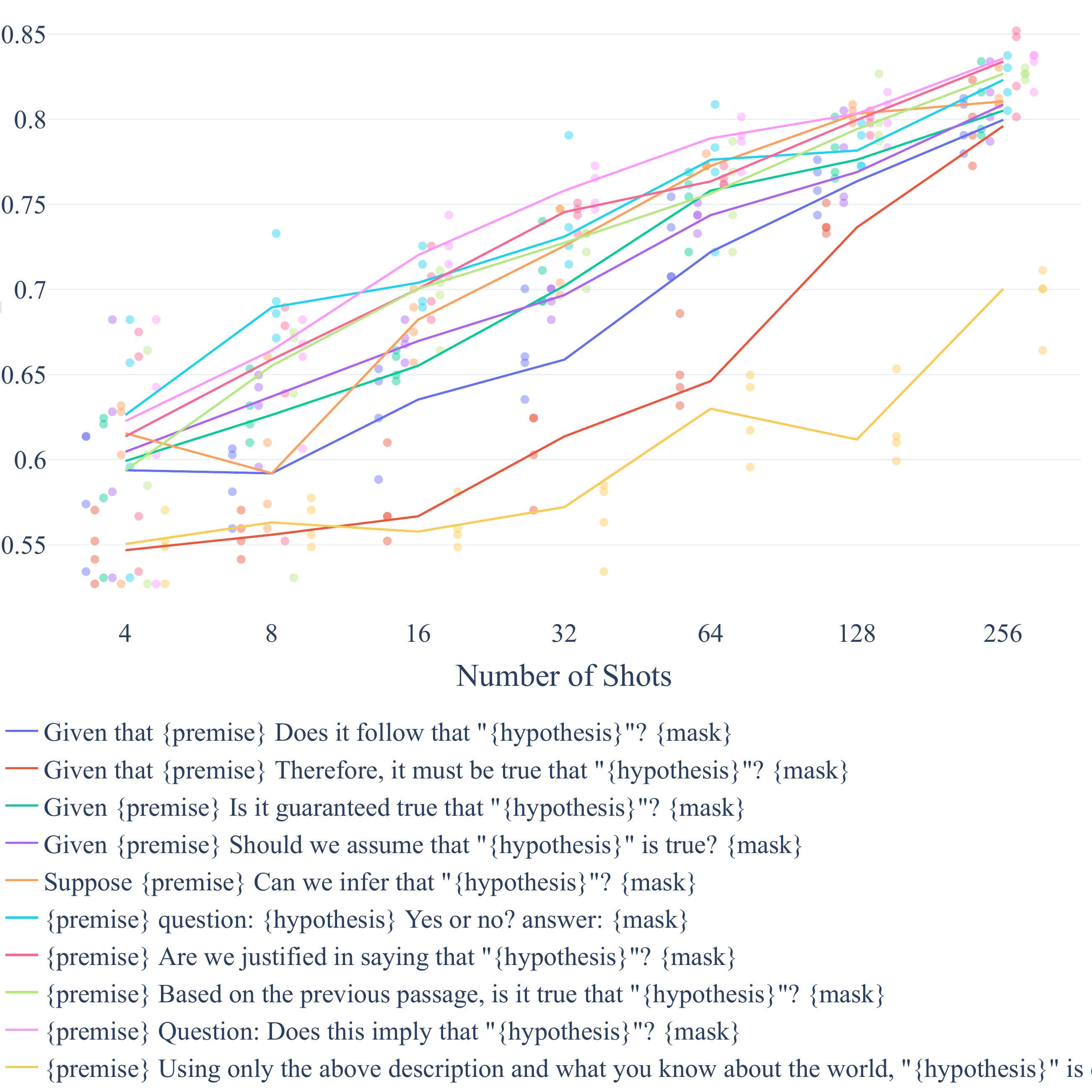}
    \caption{ALBERT with all instructive templates.}    
\end{figure*}

\begin{figure*}
    \subsection{T0 (3B)}
    \centering
    \includegraphics[width=\linewidth]{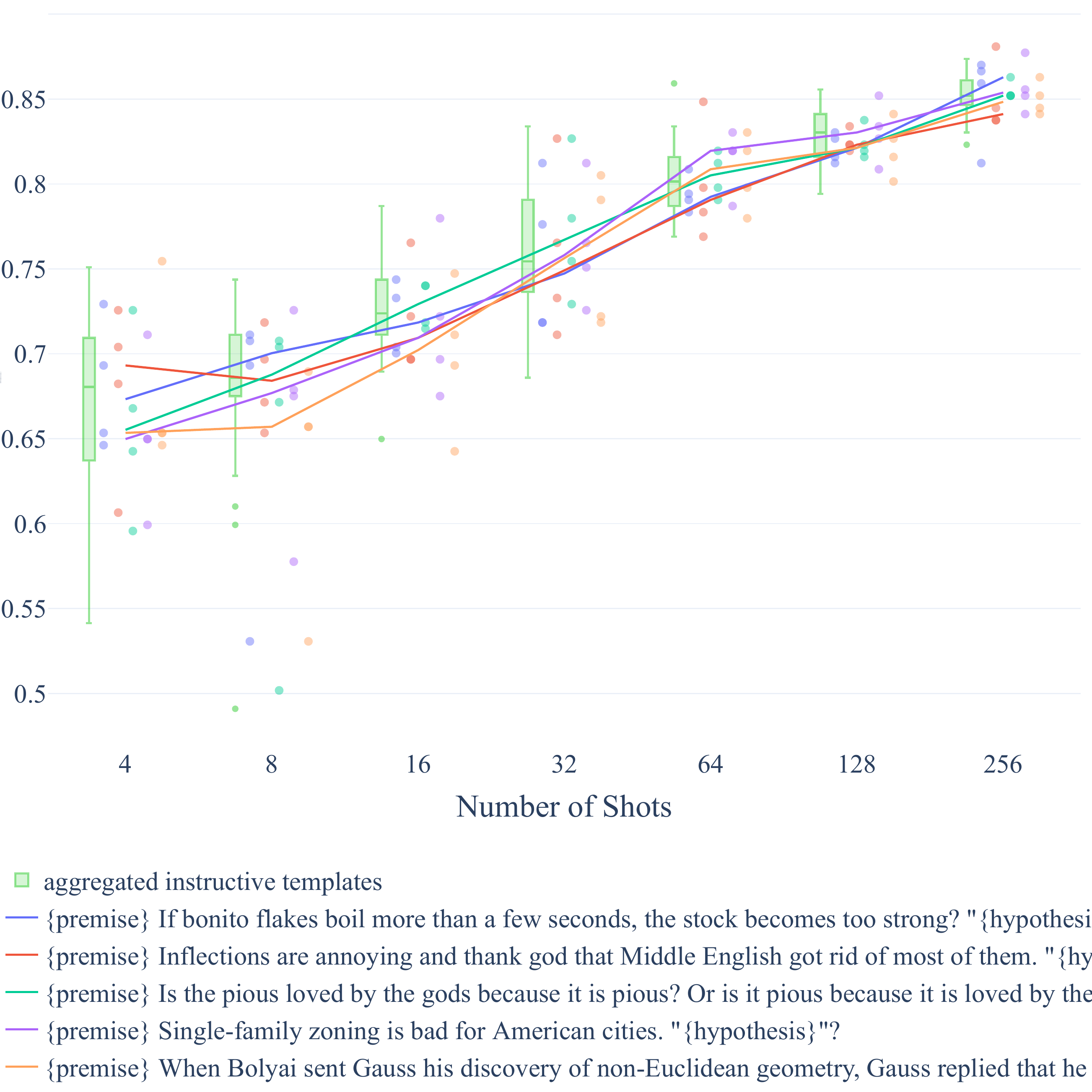}
    \caption{T0 (3B) with all irrelevant templates and the aggregated instructive for reference.}    
\end{figure*}

\begin{figure*}
    \centering
    \includegraphics[width=\linewidth]{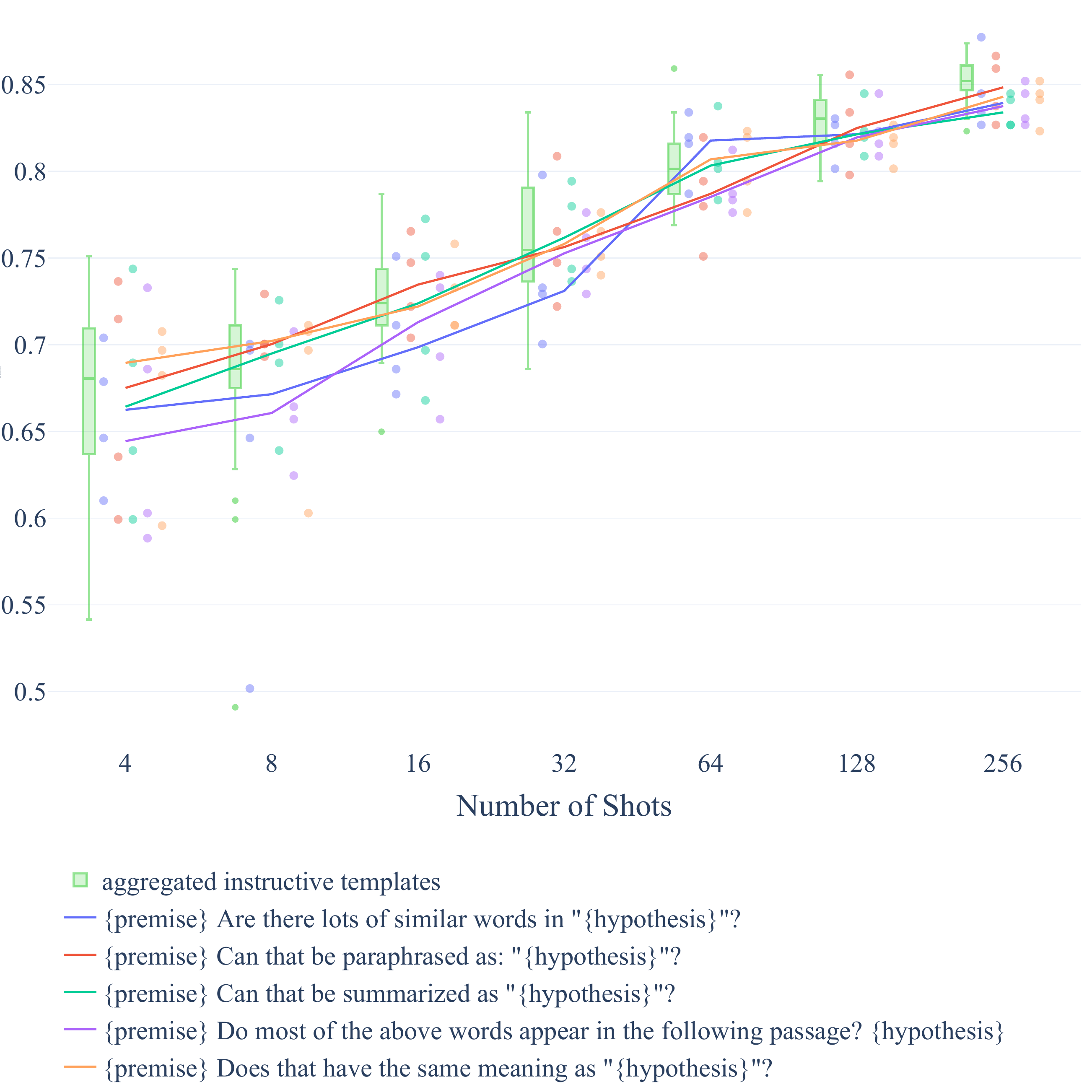}
    \caption{T0 (3B) with all misleading-moderate templates and the aggregated instructive for reference.}    
\end{figure*}

\begin{figure*}
    \centering
    \includegraphics[width=\linewidth]{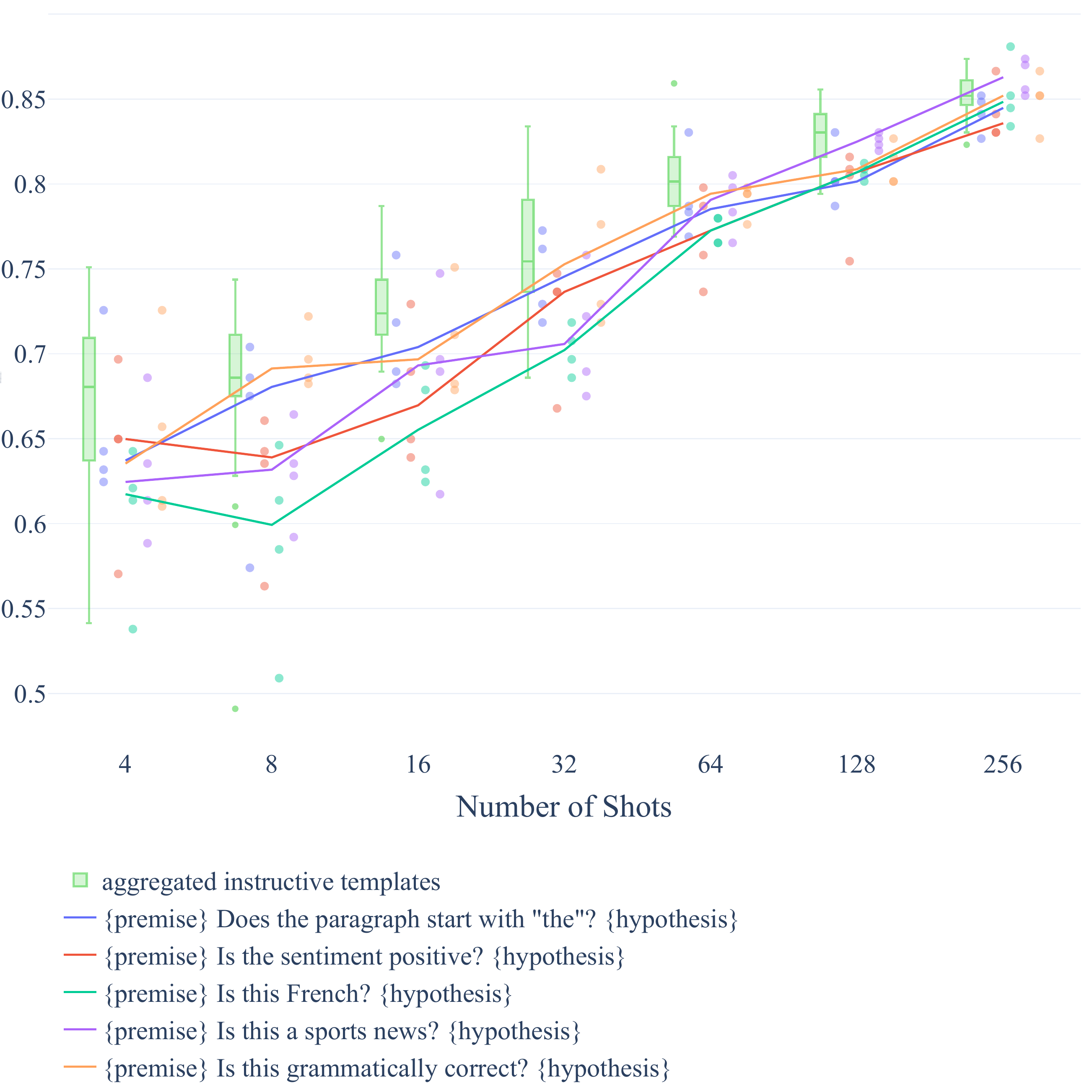}
    \caption{T0 (3B) with all misleading-extreme templates and the aggregated instructive for reference.}    
\end{figure*}

\begin{figure*}
    \centering
    \includegraphics[width=\linewidth]{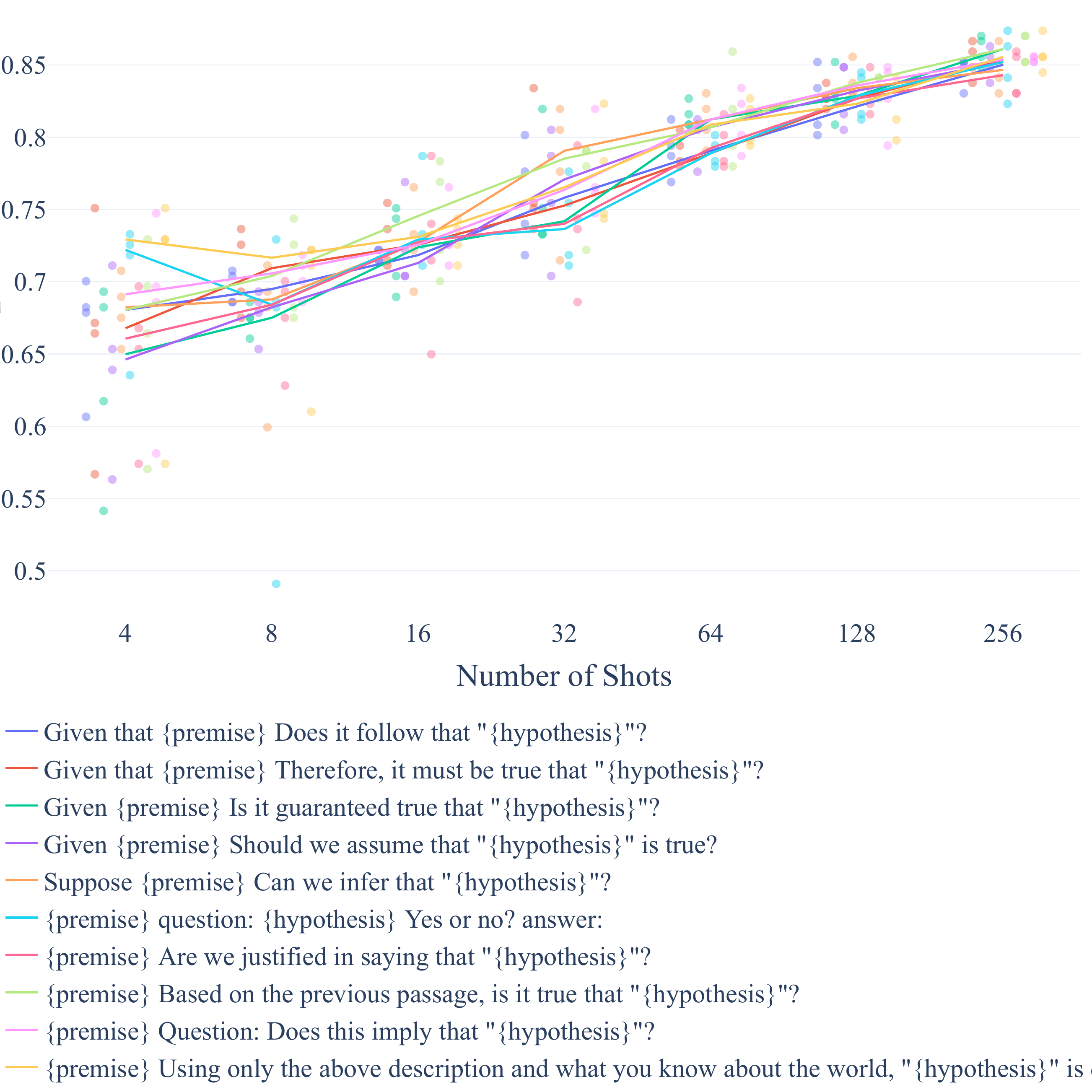}
    \caption{T0 (3B) with all instructive templates.}    
\end{figure*}

\begin{figure*}
    \subsection{T5 LM-Adapted (3B)}
    \centering
    \includegraphics[width=\linewidth]{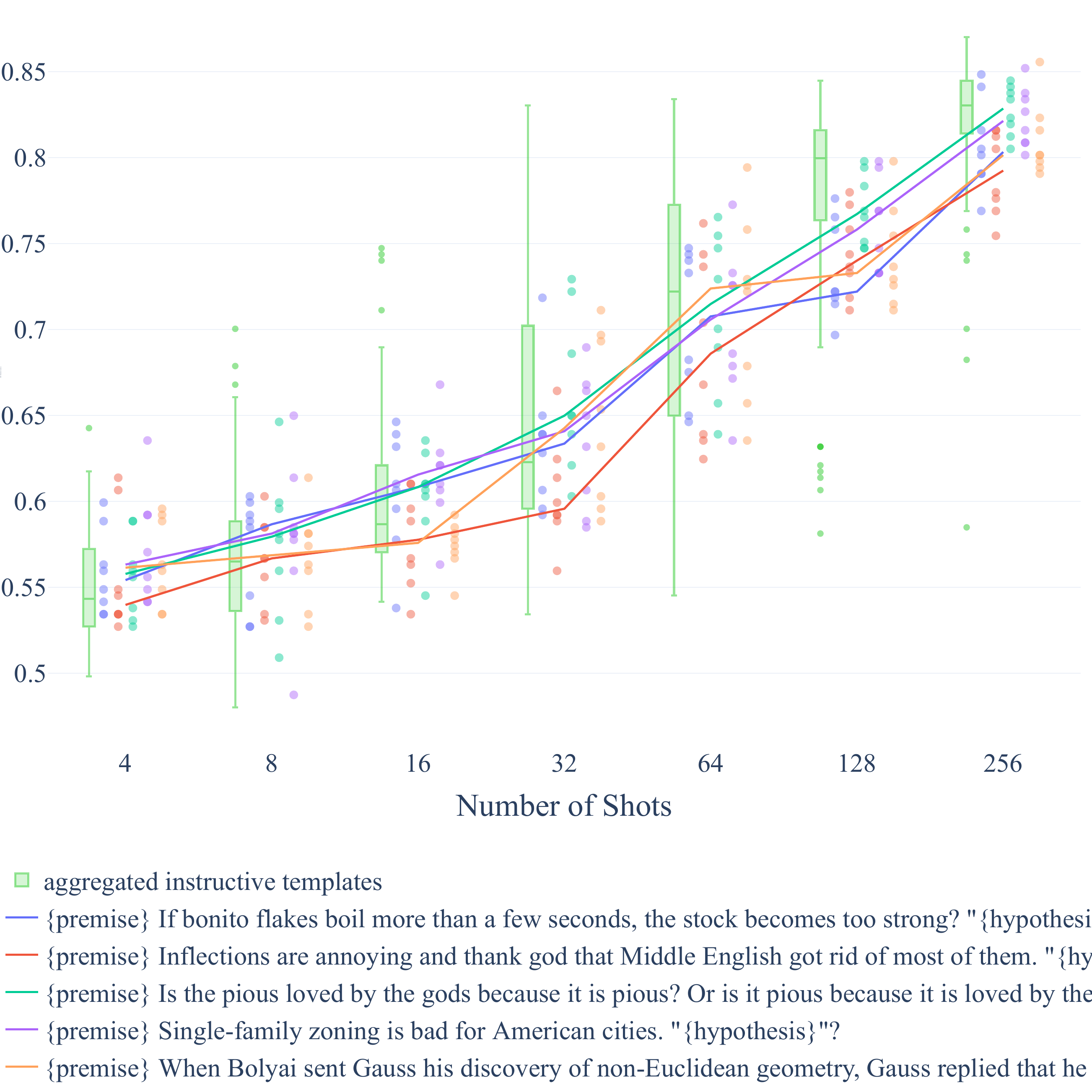}
    \caption{T5 LM-Adapted (3B) with all irrelevant templates and the aggregated instructive for reference.}    
\end{figure*}

\begin{figure*}
    \centering
    \includegraphics[width=\linewidth]{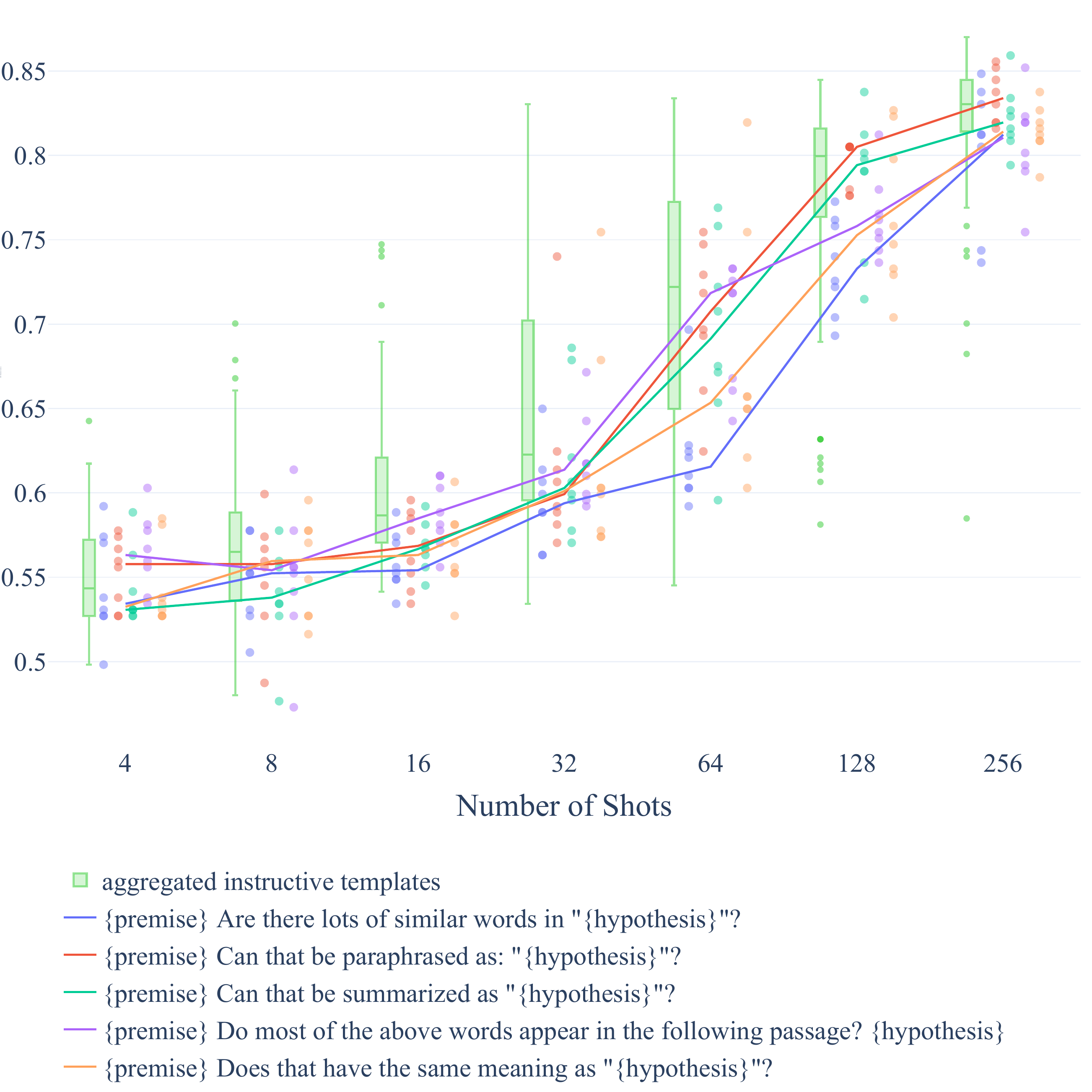}
    \caption{T5 LM-Adapted (3B) with all misleading-moderate templates and the aggregated instructive for reference.}    
\end{figure*}

\begin{figure*}
    \centering
    \includegraphics[width=\linewidth]{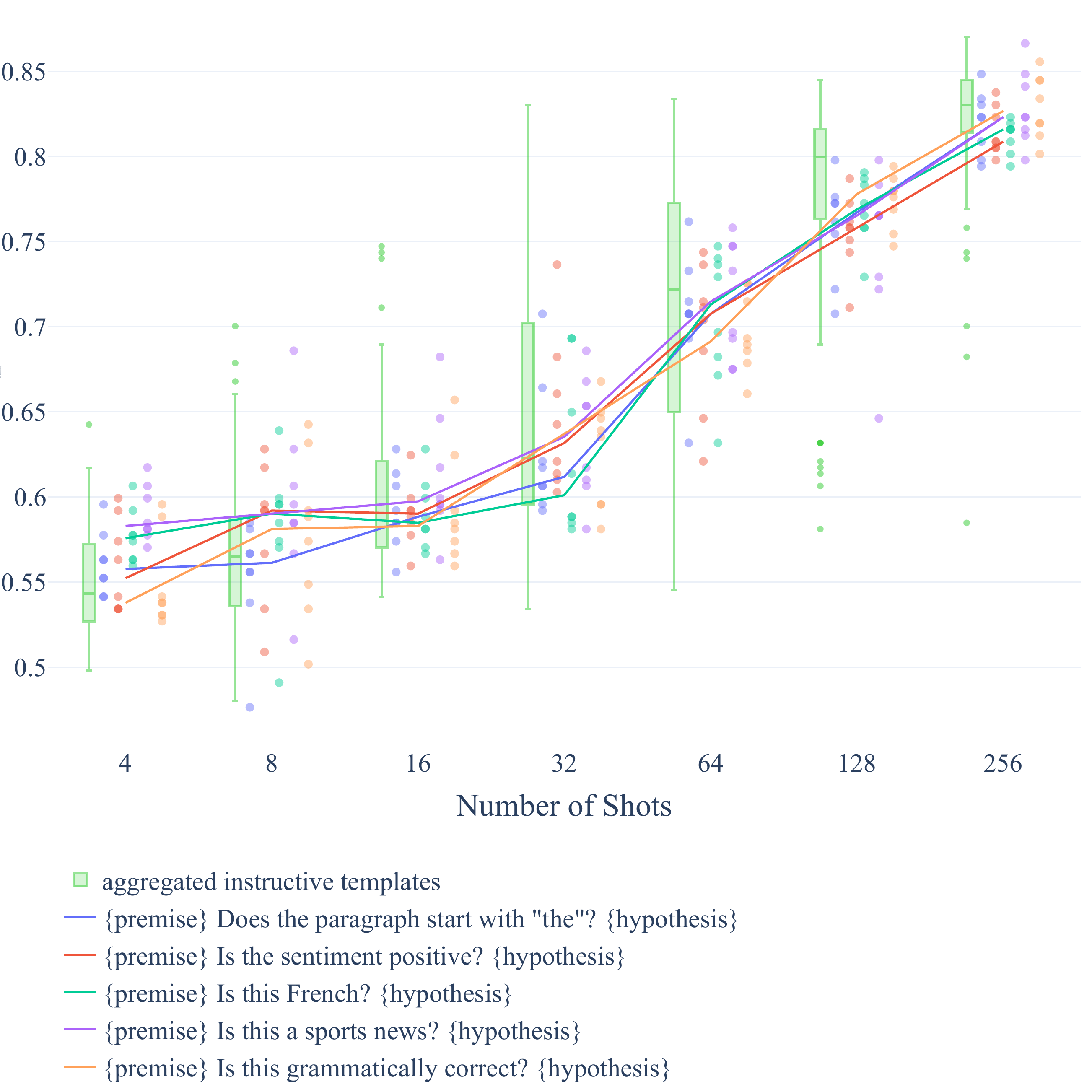}
    \caption{T5 LM-Adapted (3B) with all misleading-extreme templates and the aggregated instructive for reference.}    
\end{figure*}

\begin{figure*}
    \centering
    \includegraphics[width=\linewidth]{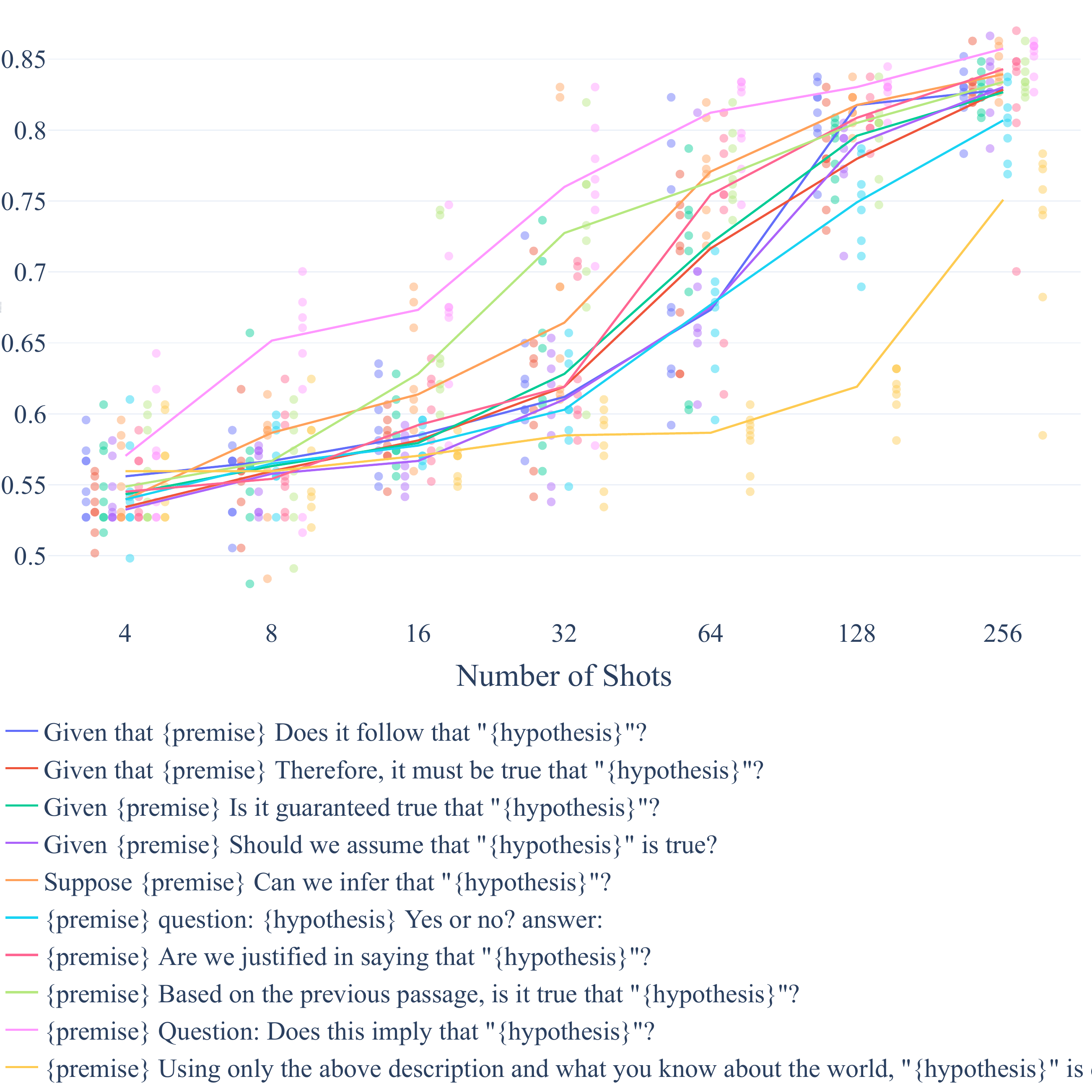}
    \caption{T5 LM-Adapted (3B) with all instructive templates.}    
\end{figure*}

\clearpage
\begin{table}[t!]
\vspace{-28ex}
\section{Zero-Shot Results (\rf{fig:zero-shot})} \label{sec:all-zero-shot} \vspace{1ex}
\centering
\resizebox{.46\textwidth}{!}{%
\begin{tabular}{lllr}
\toprule
     model &            category &       template name &  accuracy \\
\midrule
   T0 (3B) &         instructive &             MNLI\_YN &    0.7148 \\
   T0 (3B) &         instructive &              GPT\_YN &    0.6823 \\
   T0 (3B) &         instructive & justified\_in\_saying &    0.6426 \\
   T0 (3B) &         instructive &       should\_assume &    0.6498 \\
   T0 (3B) &         instructive &          is\_it\_true &    0.6462 \\
   T0 (3B) &         instructive &     guaranteed\_true &    0.6209 \\
   T0 (3B) &         instructive &        can\_we\_infer &    0.6354 \\
   T0 (3B) &         instructive &      does\_it\_follow &    0.6715 \\
   T0 (3B) &         instructive &     does\_this\_imply &    0.6679 \\
   T0 (3B) &         instructive &       modal\_be\_true &    0.6354 \\
   T0 (3B) & misleading-moderate &        words\_appear &    0.6462 \\
   T0 (3B) & misleading-moderate &       similar\_words &    0.6354 \\
   T0 (3B) & misleading-moderate &        same\_meaning &    0.6968 \\
   T0 (3B) & misleading-moderate &          paraphrase &    0.6390 \\
   T0 (3B) & misleading-moderate &           summarize &    0.6462 \\
   T0 (3B) &  misleading-extreme &      start\_with\_the &    0.6968 \\
   T0 (3B) &  misleading-extreme &         grammatical &    0.6859 \\
   T0 (3B) &  misleading-extreme &           sentiment &    0.6462 \\
   T0 (3B) &  misleading-extreme &          sportsball &    0.6426 \\
   T0 (3B) &  misleading-extreme &              french &    0.5668 \\
   T0 (3B) &          irrelevant &              zoning &    0.5704 \\
   T0 (3B) &          irrelevant &               gauss &    0.5523 \\
   T0 (3B) &          irrelevant &         katsuobushi &    0.5668 \\
   T0 (3B) &          irrelevant &          inflection &    0.6751 \\
   T0 (3B) &          irrelevant &           euthyphro &    0.6606 \\
   T0 (3B) &                null &          concat\_PHM &    0.6426 \\
   T0 (3B) &                null &          concat\_HPM &    0.6029 \\
\bottomrule
\end{tabular}
}
\label{tab:all-zero-shot}
\vspace{30ex}
\end{table}

\begin{table}[t!]
%\vspace{-2ex}
%\section{Zero-Shot Results (\rf{fig:zero-shot})} \vspace{1ex}
%\centering
\resizebox{0.49\textwidth}{!}{%
\begin{tabular}{lllr}
\toprule
     model &            category &       template name &  accuracy \\
\midrule
  T0 (11B) &         instructive &             MNLI\_YN &    0.8051 \\
  T0 (11B) &         instructive &              GPT\_YN &    0.8014 \\
  T0 (11B) &         instructive & justified\_in\_saying &    0.7112 \\
  T0 (11B) &         instructive &       should\_assume &    0.7437 \\
  T0 (11B) &         instructive &          is\_it\_true &    0.8051 \\
  T0 (11B) &         instructive &     guaranteed\_true &    0.6968 \\
  T0 (11B) &         instructive &        can\_we\_infer &    0.7690 \\
  T0 (11B) &         instructive &      does\_it\_follow &    0.7509 \\
  T0 (11B) &         instructive &     does\_this\_imply &    0.8014 \\
  T0 (11B) &         instructive &       modal\_be\_true &    0.6895 \\
  T0 (11B) & misleading-moderate &        words\_appear &    0.7184 \\
  T0 (11B) & misleading-moderate &       similar\_words &    0.7148 \\
  T0 (11B) & misleading-moderate &        same\_meaning &    0.7256 \\
  T0 (11B) & misleading-moderate &          paraphrase &    0.7256 \\
  T0 (11B) & misleading-moderate &           summarize &    0.6679 \\
  T0 (11B) &  misleading-extreme &      start\_with\_the &    0.6823 \\
  T0 (11B) &  misleading-extreme &         grammatical &    0.6390 \\
  T0 (11B) &  misleading-extreme &           sentiment &    0.6318 \\
  T0 (11B) &  misleading-extreme &          sportsball &    0.5921 \\
  T0 (11B) &  misleading-extreme &              french &    0.5271 \\
  T0 (11B) &          irrelevant &              zoning &    0.6318 \\
  T0 (11B) &          irrelevant &               gauss &    0.5560 \\
  T0 (11B) &          irrelevant &         katsuobushi &    0.5740 \\
  T0 (11B) &          irrelevant &          inflection &    0.7004 \\
  T0 (11B) &          irrelevant &           euthyphro &    0.6931 \\
  T0 (11B) &                null &          concat\_PHM &    0.6570 \\
  T0 (11B) &                null &          concat\_HPM &    0.6209 \\
T0++ (11B) &         instructive &             MNLI\_YN &    0.8592 \\
T0++ (11B) &         instructive &              GPT\_YN &    0.8231 \\
T0++ (11B) &         instructive & justified\_in\_saying &    0.7726 \\
T0++ (11B) &         instructive &       should\_assume &    0.8231 \\
T0++ (11B) &         instructive &          is\_it\_true &    0.8556 \\
T0++ (11B) &         instructive &     guaranteed\_true &    0.8231 \\
T0++ (11B) &         instructive &        can\_we\_infer &    0.8303 \\
T0++ (11B) &         instructive &      does\_it\_follow &    0.7798 \\
T0++ (11B) &         instructive &     does\_this\_imply &    0.8664 \\
T0++ (11B) &         instructive &       modal\_be\_true &    0.8087 \\
T0++ (11B) & misleading-moderate &        words\_appear &    0.7076 \\
T0++ (11B) & misleading-moderate &       similar\_words &    0.7329 \\
T0++ (11B) & misleading-moderate &        same\_meaning &    0.7545 \\
T0++ (11B) & misleading-moderate &          paraphrase &    0.7617 \\
T0++ (11B) & misleading-moderate &           summarize &    0.6968 \\
T0++ (11B) &  misleading-extreme &      start\_with\_the &    0.6498 \\
T0++ (11B) &  misleading-extreme &         grammatical &    0.7762 \\
T0++ (11B) &  misleading-extreme &           sentiment &    0.7365 \\
T0++ (11B) &  misleading-extreme &          sportsball &    0.5307 \\
T0++ (11B) &  misleading-extreme &              french &    0.4838 \\
T0++ (11B) &          irrelevant &              zoning &    0.5018 \\
T0++ (11B) &          irrelevant &               gauss &    0.5090 \\
T0++ (11B) &          irrelevant &         katsuobushi &    0.4801 \\
T0++ (11B) &          irrelevant &          inflection &    0.7220 \\
T0++ (11B) &          irrelevant &           euthyphro &    0.6715 \\
T0++ (11B) &                null &          concat\_PHM &    0.6426 \\
T0++ (11B) &                null &          concat\_HPM &    0.6029 \\
\bottomrule
\end{tabular}
}
%\label{tab:all-zero-shot}
\end{table}

\begin{figure*}[h]
    \section{Comparison of LM targets, Controlling for the Template} \label{sec:ceteris-paribus}
    \includegraphics[width=\linewidth]{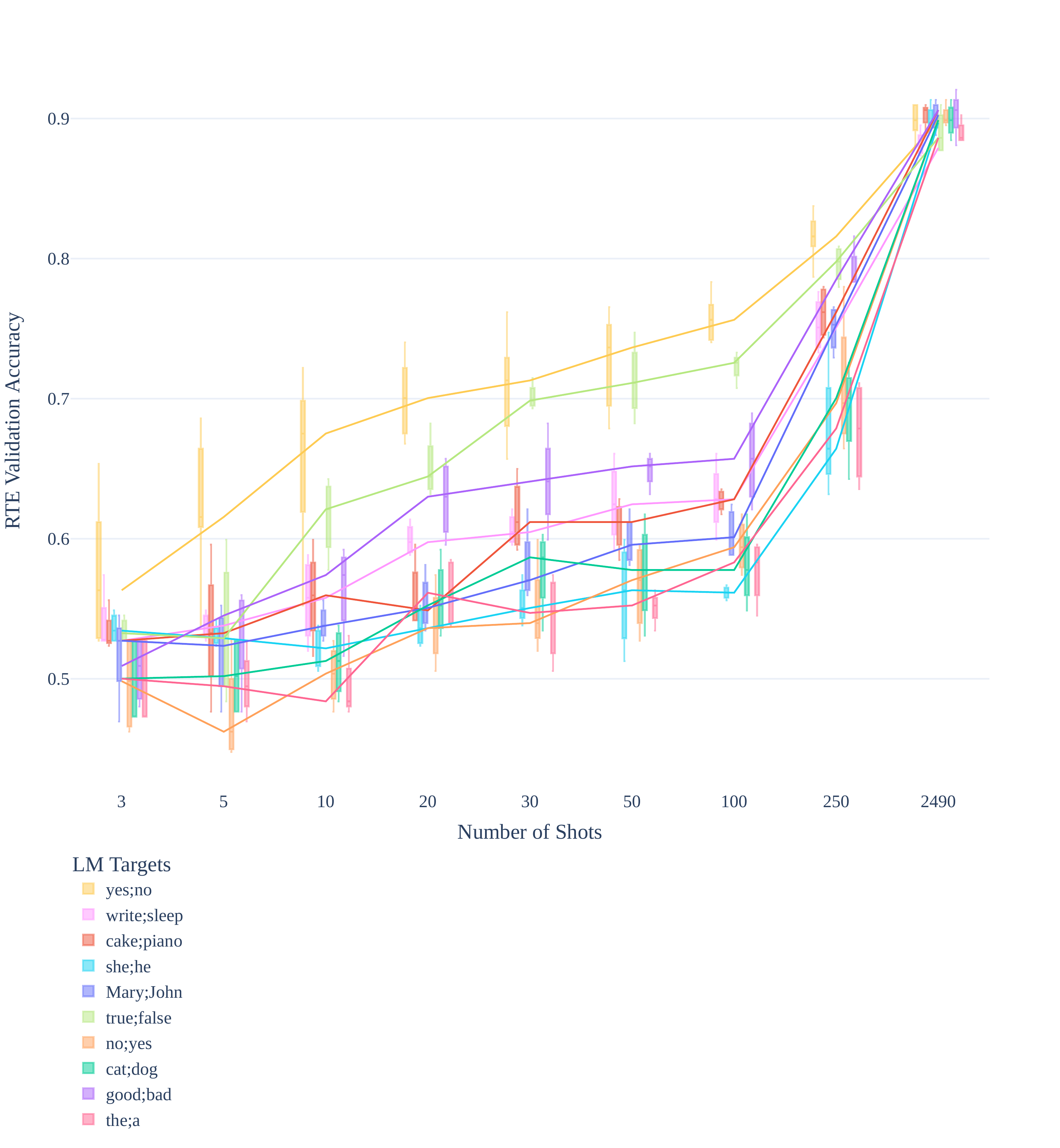}    
    \caption{The best performing irrelevant prompt for ALBERT, \prompt{\{premise\} Single-family zoning is bad for American cities. "\{hypothesis\}"? [mask]} with all LM targets.} %In general, yes-no > true-false > good-bad > arbitrary > reversed (e.g., no-yes). The arbitrary group includes antonyms, semantic similars (cat-dog), dissimilars (write-sleep), common first names (Mary-John), and highly frequent words (she-he). We found no substnatial difference whitin this arbitrary group. }
    \label{fig:zoning-verb}
\end{figure*}

\begin{figure*}[h]
    \includegraphics[width=\linewidth]{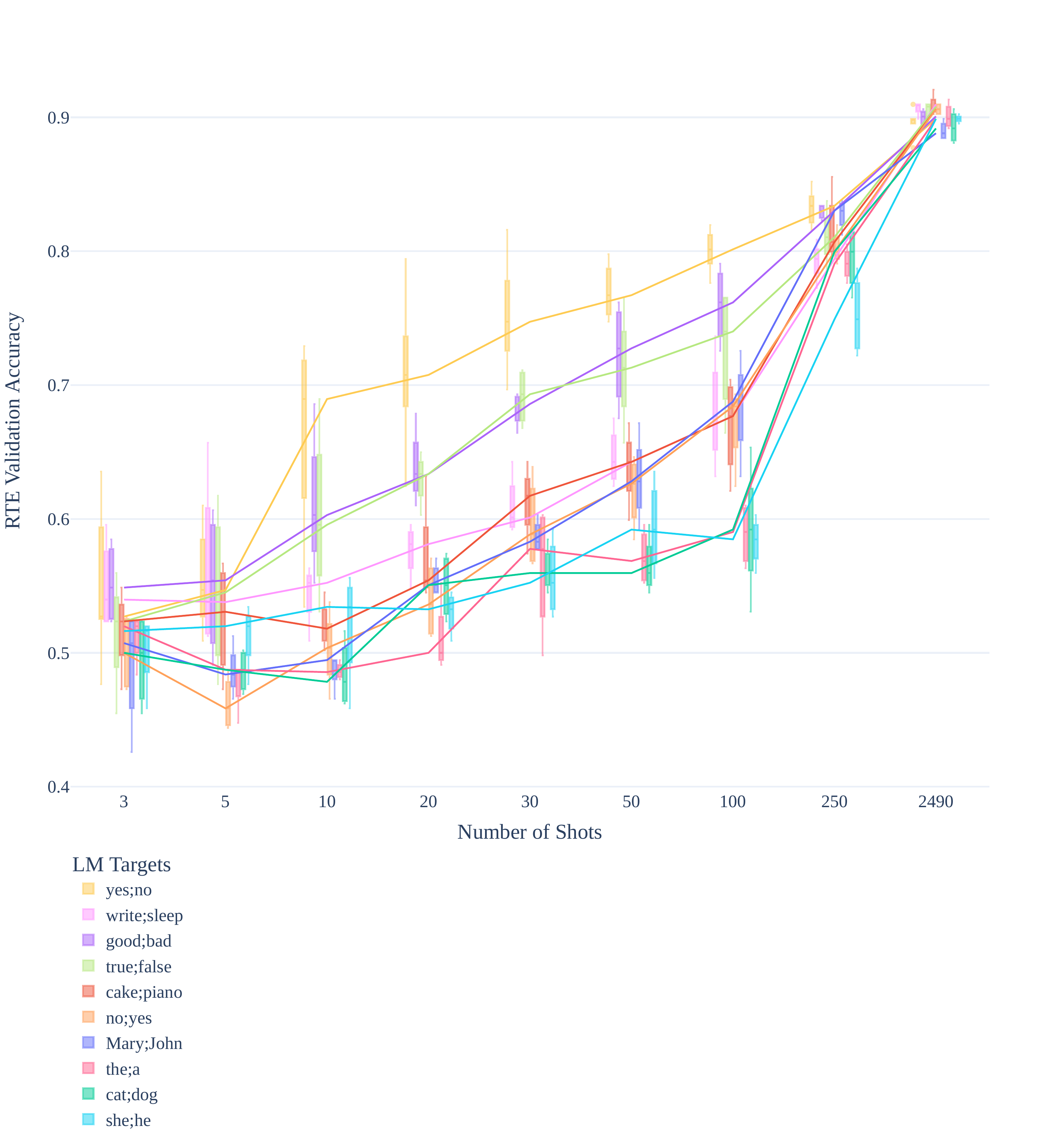}    
    \caption{The best-performing misleading prompt for ALBERT, \prompt{\{premise\} Does the paragraph start with "the"? [mask] "\{hypothesis\}"} with all LM targets.}
    \label{fig:misleading-verb}
\end{figure*}

\begin{figure*}[h]
    \includegraphics[width=\linewidth]{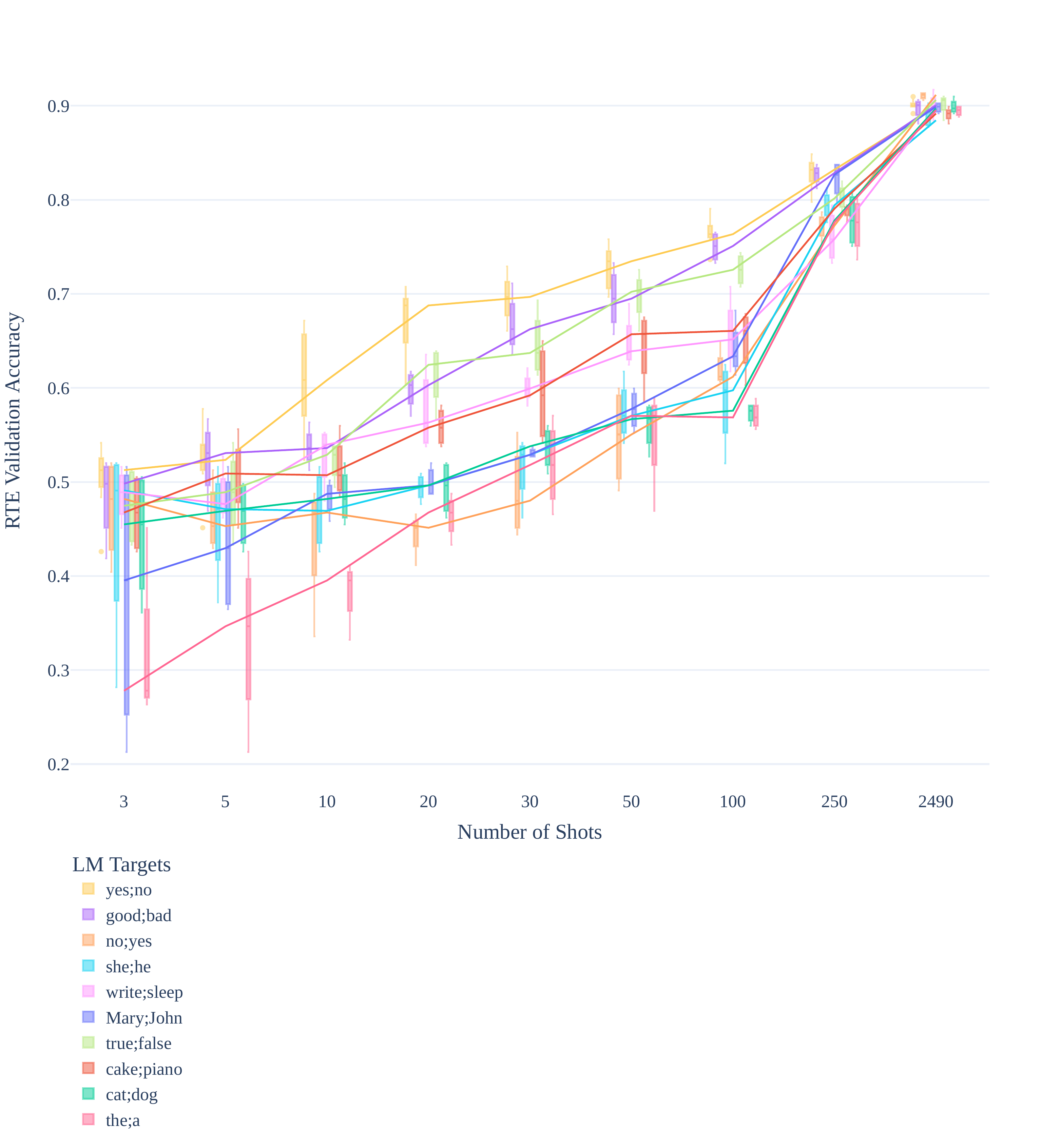}    
    \caption{The best-performing null prompt for ALBERT, \prompt{\{premise\} [mask] "\{hypothesis\}"} with all LM targets.}
    \label{fig:null-verb}
\end{figure*}

\clearpage
\begin{figure*}
    \section{Preliminary Results on HANS}
    \label{sec:hans-preliminary}
    \includegraphics[width=\linewidth]{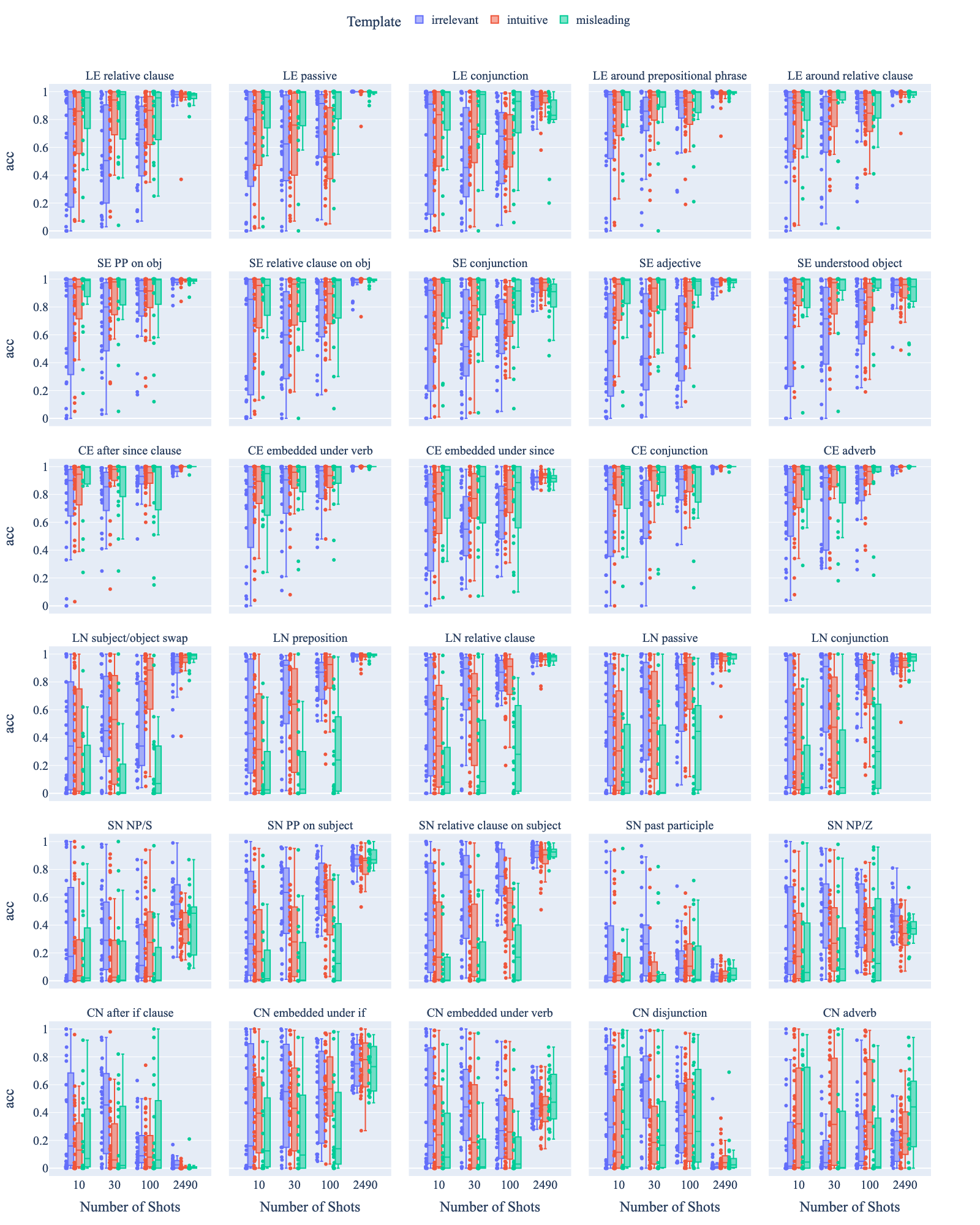}
    \caption{Few-shot RTE-trained ALBERT's zero-shot performance on HANS \citep{mccoy}. L = lexical, S = subsequence, C = constituency. E = true label is entailment. N = true label is non-entailment. Apologies but note the template category colors are different from those in the main text. “Intuitive” = instructive templates. In general, models perform similarly with instructive and irrelevant templates, but models with misleading templates fare worse, especially for lexical non-entailment cases (LN, fourth row). A full analysis will be furnished in a future version of this paper.}    
\end{figure*}

\clearpage
\begin{table*}[h]
\section{Preliminary Results on Winograd}
\label{sec:wsc-preliminary}
\resizebox{\textwidth}{!}{%
\begin{tabular}{lll} 
\toprule
category & template & accuracy\\ 
\midrule
instructive & Is ``\pron" the same as \refs? Yes or No? & 0.6538\\ 
instructive & Does ``\pron" refer to \refs? Yes or No? & 0.6731\\ 
instructive & Is ``\pron" \refs? Yes or No? & 0.5385\\ 
instructive & Should ``\pron" be \refs? Yes or No? & 0.5962\\ 
instructive & Does ``\pron" mean \refs? Yes or No? & 0.6442\\ 
instructive & Is``\pron" equivalent to \refs? Yes or No? & 0.6058\\ 
instructive & Does ``\pron" stand for \refs? Yes or No? & 0.6346\\ 
instructive & Can the pronoun ``\pron" be replaced with \refs? Yes or No? & 0.6250\\ 
 \\ 
misleading-extreme & Did ``\pron" eat cakes with \refs? Yes or No? & 0.6346\\ 
misleading-extreme & Is ``\pron" mother of \refs? Yes or No? & 0.6346\\ 
misleading-extreme & Was ``\pron" friend to \refs? Yes or No? & 0.6058\\ 
misleading-extreme & Did  ``\pron" marry \refs? Yes or No? & 0.6346\\ 
misleading-extreme & Can ``\pron" rent a car with \refs? Yes or No? & 0.6346\\ 
misleading-extreme & Should ``\pron" be brother of \refs? Yes or No? & 0.6346\\ 
misleading-extreme & Did ``\pron" speak to \refs? Yes or No? & 0.5673\\ 
misleading-extreme & Is ``\pron" cousins with \refs? Yes or No? & 0.6154\\ 
\bottomrule
\end{tabular}
}
\end{table*}
\begin{figure*}[h]
    \centering
    \includegraphics[width=0.8\linewidth]{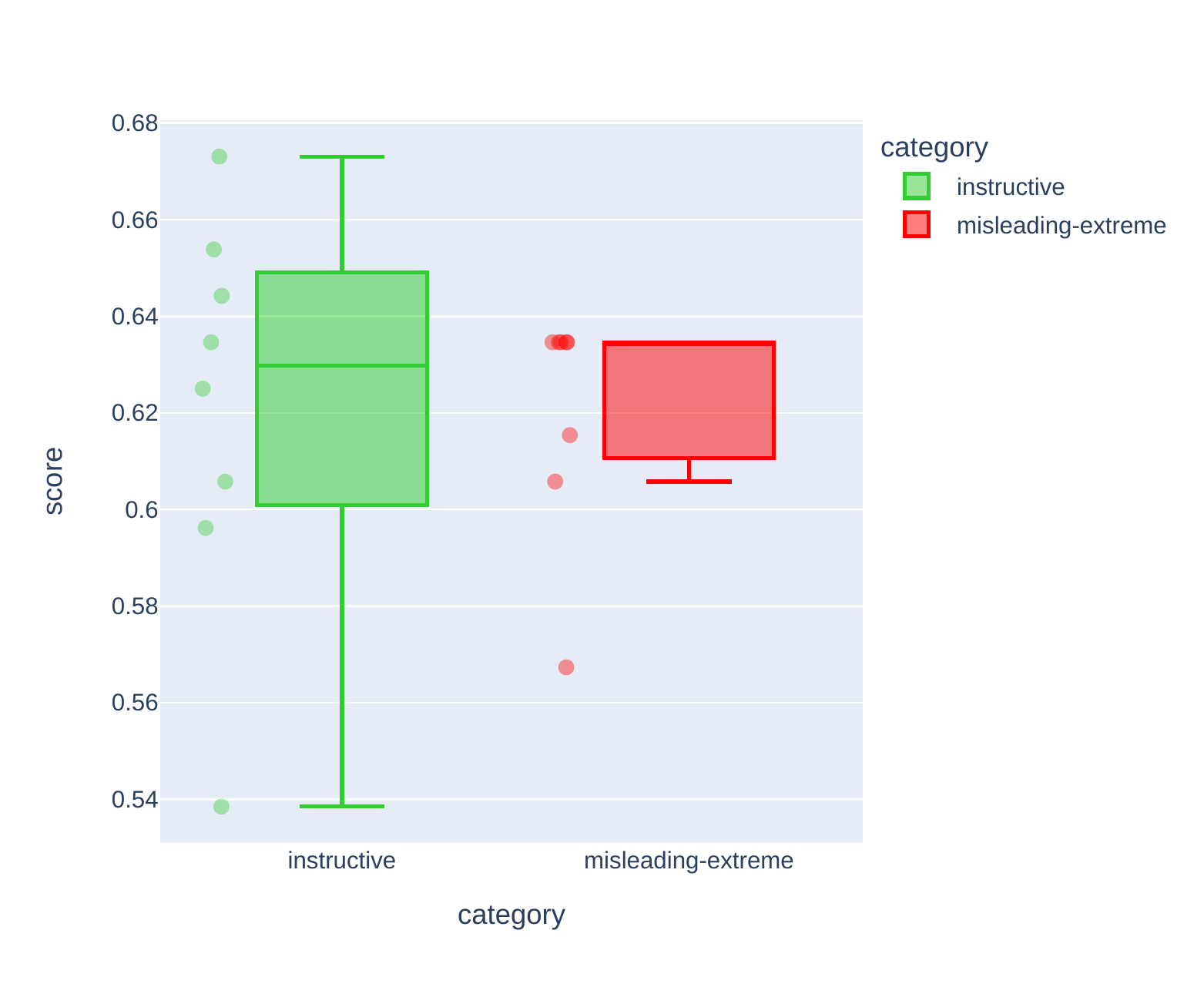}    
    \caption{Zero-shot accuracy of T0 on Winograd Schema Challenge (\citealp{winograd}; SuperGLUE version). We find no statistically significant difference between instructive and misleading-extreme templates.}
\end{figure*}

\end{document}